\documentclass[preprint,3p,times,twocolumn]{elsarticle}

\usepackage[table,xcdraw]{xcolor}

\usepackage{graphicx}
\usepackage{times}  
\usepackage{helvet}  
\usepackage{courier}  
\usepackage{url}  
\usepackage{subfigure}
\usepackage{algorithm}
\usepackage{algpseudocode}
\usepackage{comment}
\usepackage{multirow}
\usepackage[table,xcdraw]{xcolor}
\usepackage{amsmath} 
\usepackage{amssymb}  
\usepackage{todonotes}
\usepackage{hyperref}
\usepackage{zref-xr,zref-user}
\usepackage{balance}

\usepackage{float}
\usepackage{stfloats}

\usepackage{tikz}
\usetikzlibrary{decorations.pathreplacing} 

\usepackage{soul}

\algnewcommand\algorithmicforeach{\textbf{for each}}
\algdef{S}[FOR]{ForEach}[1]{\algorithmicforeach\ #1\ \algorithmicdo}

\journal{Robotics and Autonomous Systems}

\begin{document}

\begin{frontmatter}

\title{Traversability-aware path planning in dynamic environments}

\author{Yaroslav Marchukov and Luis Montano}

\address{Instituto de Investigaci\'on en Ingenier\'ia de Arag\'on (I3A),
University of Zaragoza, Spain \\ yamar@unizar.es, montano@unizar.es}

\begin{abstract}
Planning in environments with moving obstacles remains a significant challenge in robotics. While many works focus on navigation and path planning in obstacle-dense spaces, traversing such congested regions is often avoidable by selecting alternative routes. This paper presents Traversability-aware FMM (\emph{Tr-FMM}), a path planning method that computes paths in dynamic environments, avoiding crowded regions. The method operates in two steps: first, it discretizes the environment, identifying regions and their distribution; second, it evaluates the traversability of regions, aiming to minimize both obstacle risks and goal deviation. The path is then computed by propagating the wavefront through regions with higher traversability. Simulated and real-world experiments demonstrate that the approach ensures significant safety by keeping the robot away from obstacles while minimizing excessive goal deviations.

\end{abstract}


\end{frontmatter}

\section{Introduction}\label{sec:intro}

Robots operating without direct human supervision or intervention in everyday life are becoming increasingly common. Consequently, moving in spaces shared with humans emerged as a significant challenge in robotics \cite{complexity_social}. Typical examples of such environments include indoor settings like stores, warehouses, and airports \cite{mapf_warehouses}. In these crowded or busy environments, people often move unpredictably or without paying sufficient attention to robots, potentially leading to collisions or deadlock situations from which a robot cannot recover \cite{survey_social_navigation}. Therefore, it is crucial that robots are capable of avoiding such situations, where people are seen as dynamic obstacles needed to be avoided.

Classic and widely used navigation techniques, such as DWA \cite{dwa} and elastic bands \cite{elastic_bands}, struggle in the aforementioned situations. DWA is designed for static scenarios, while elastic bands are not suited for highly dynamic and crowded environments. Navigation methods that account for dynamic obstacles, such as VO \cite{shiller}, RVO \cite{rvo}, and ORCA \cite{orca}, are designed as local planners for maneuvering among people or moving obstacles, rather than as global planners for such scenarios. More recent approaches, often based on advanced learning techniques \cite{crowd_navigation}\cite{navigation_diego}\cite{navigation_object}, demonstrate higher success rates in avoiding collisions. 
All these techniques are most useful when the robot is already inside a crowd or has no choice but to pass through one, accepting the potential risk of collision. However, none of the above methods consider directly avoiding such regions, where robots are exposed to greater risk.

Since many of the scenarios we deal with contain multiple rooms, entrances, and corridors, robots could navigate through obstacle-free regions or areas with low dynamic obstacle density, thereby reducing the risk of collision. Recent navigation approaches as \cite{congested_navigation2} take into account obstacle-congested regions to avoid navigating through them. However, these navigation methods are primarily short-term planning approaches. Therefore, the most effective way to deal with the problem above is by planning paths that deliberately avoid such regions. This is the focus of the present work, a path planning method to avoid regions with a high presence of dynamic obstacles.

\begin{figure*}[tb!]
    \begin{center}
        \subfigure[]{\frame{\includegraphics[width=0.24\columnwidth]{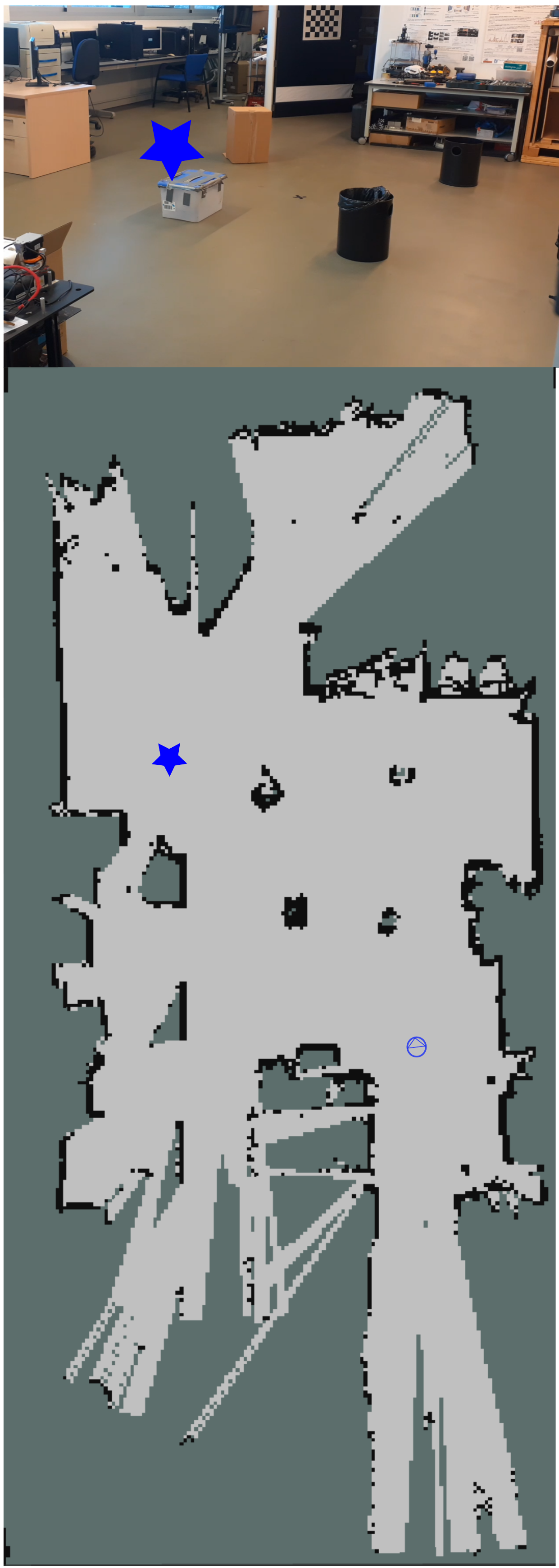}\label{fig:1}}} \hfill
        \subfigure[]{{\frame{\includegraphics[width=0.318\columnwidth]{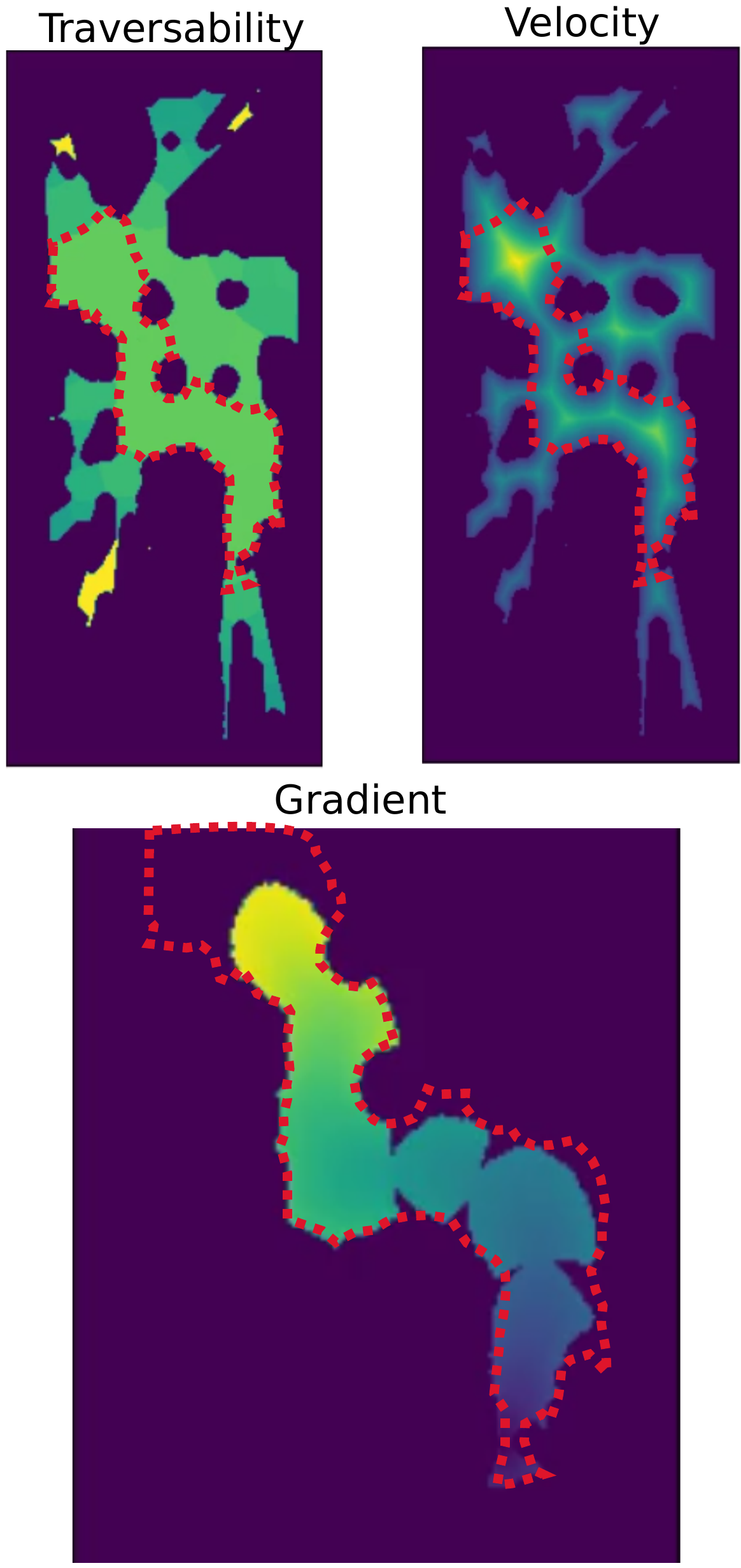}}\label{fig:2}}} \hfill
        \subfigure[]{\frame{\includegraphics[width=0.24\columnwidth]{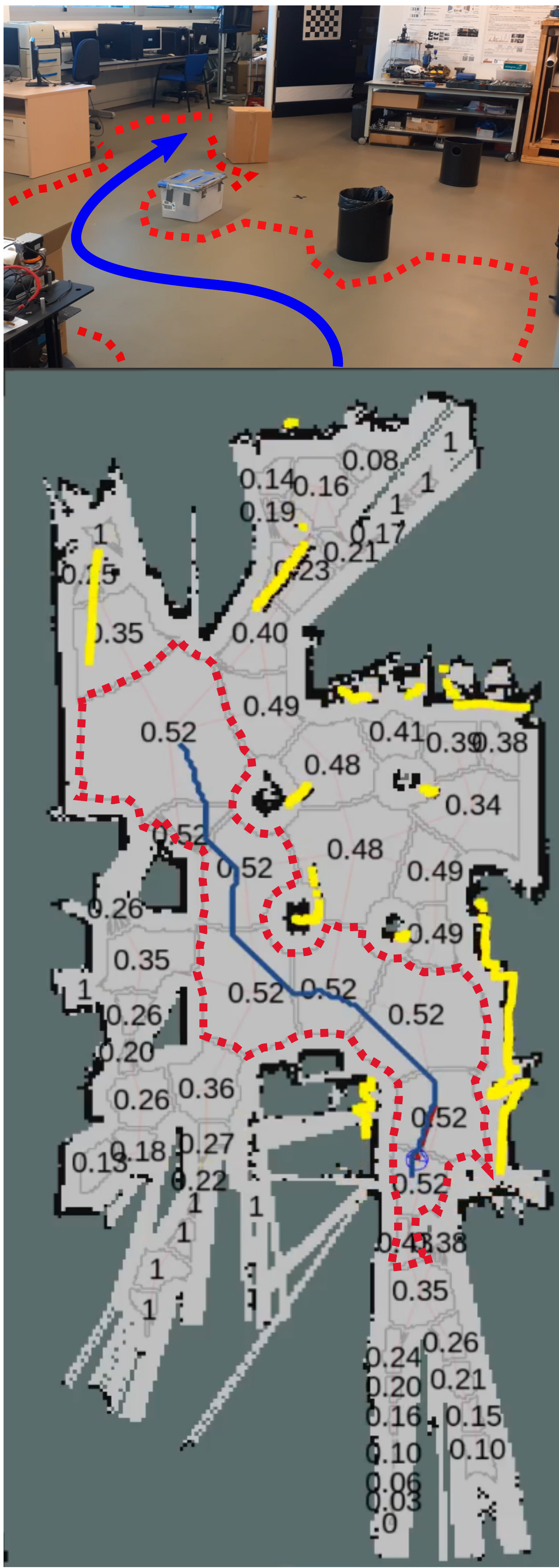}\label{fig:3}}}
        \hfill \vrule width 1pt \hfill
        \subfigure[]{\frame{\includegraphics[width=0.24\columnwidth]{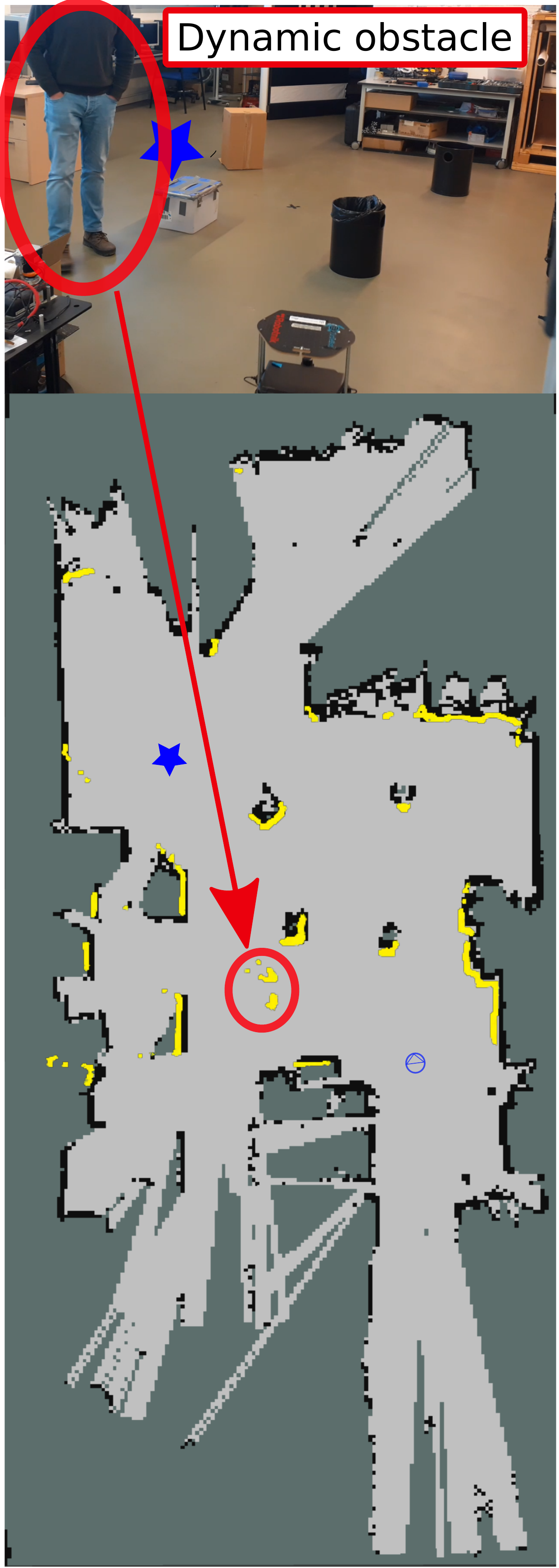}\label{fig:4}}} \hfill
        \subfigure[]{{\frame{\includegraphics[width=0.318\columnwidth]{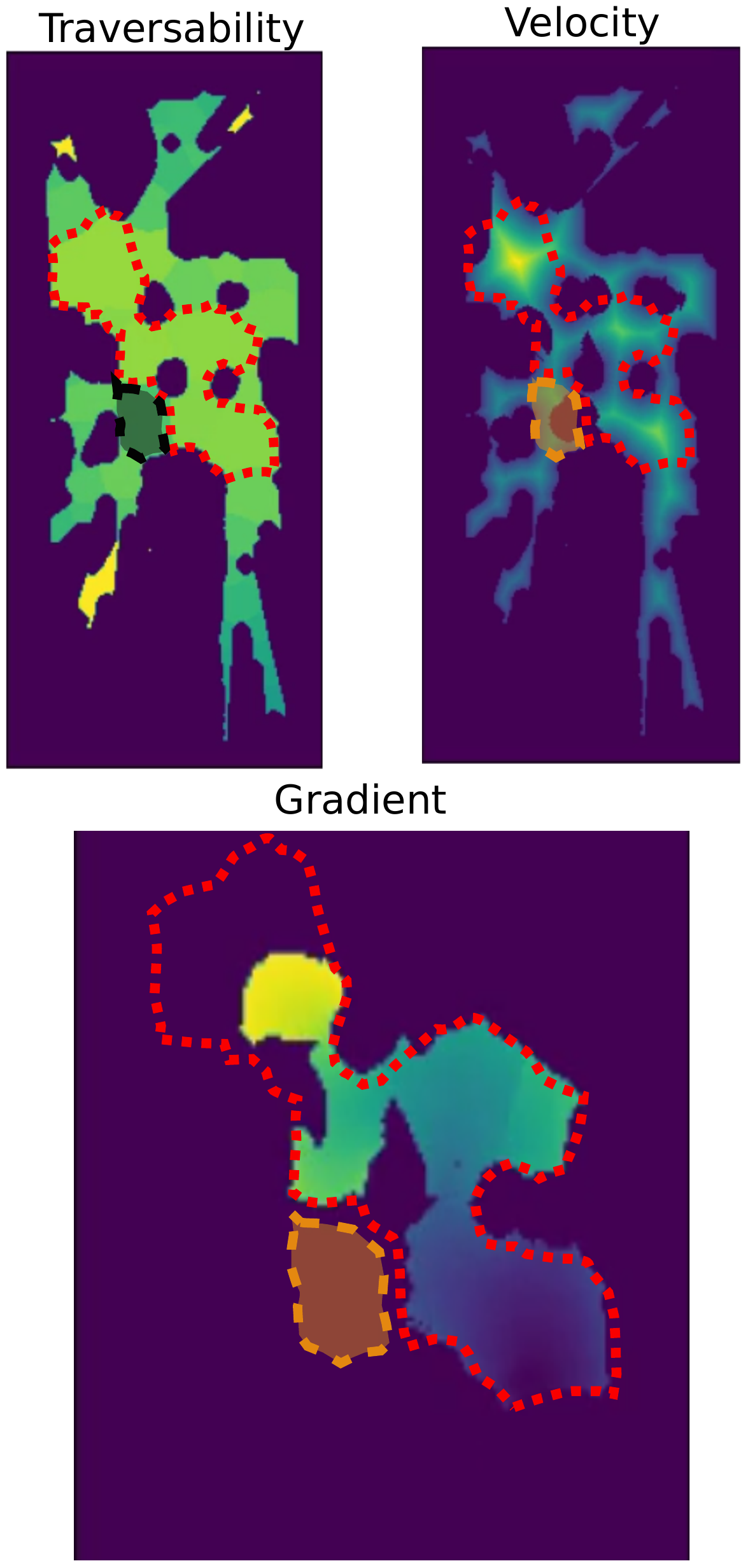}}\label{fig:5}}} \hfill
        \subfigure[]{\frame{\includegraphics[width=0.24\columnwidth]{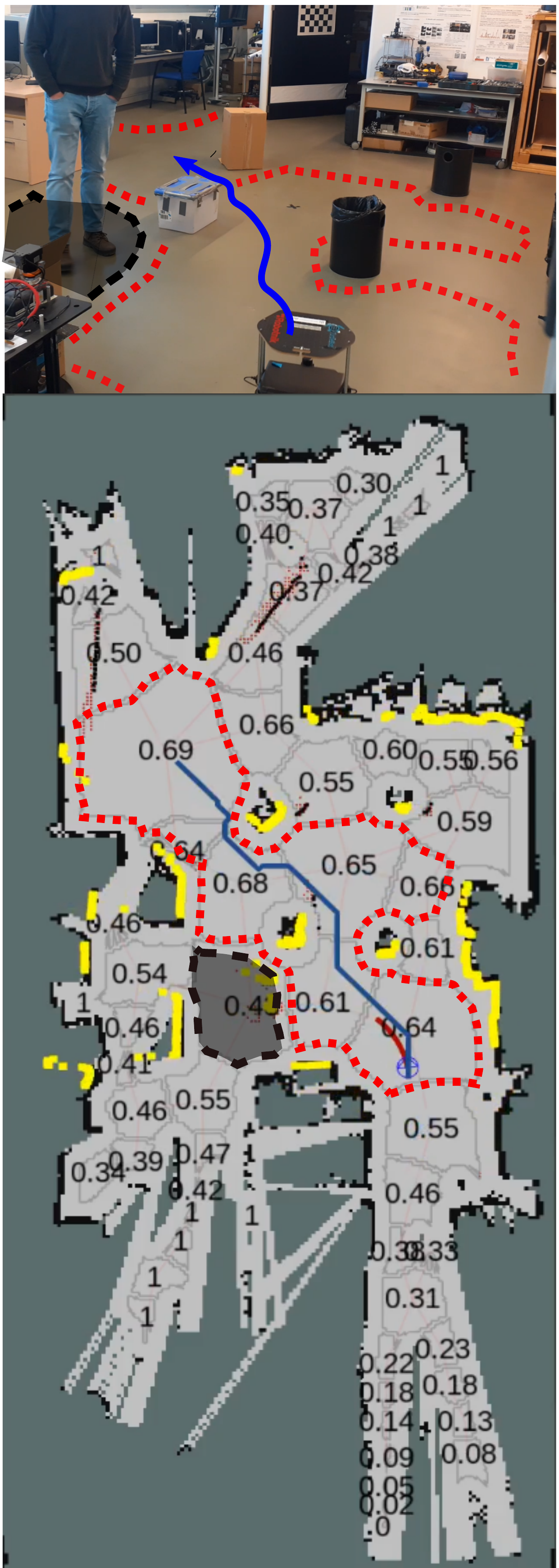}\label{fig:6}}}
        
    \end{center}
    \caption{
        Planning process of \emph{Tr-FMM} in a simple scenario. (a) illustrates the scenario with the static obstacles and the positions of the robot (blue circle) and goal (blue star). (b) displays the traversability and velocity maps with the gradient to lead the robot to the goal. The traversability map guides the wavefront through safer regions with minimal deviation (red dashed polygon), while the velocity map is used to avoid obstacles. (c) shows the obtained path through regions with higher traversability (direct to goal). When a person in movement or dynamic obstacle is detected in (d), \emph{Tr-FMM} takes into account this information. In (e), the region where the person is moving is depicted with black and orange polygons (two colors are used for the same region for better visualization in different maps). The traversability of this region is reduced due to a potential risk of collision with the person, so that, the gradient surrounds this region. The path through safer and direct regions to the goal is obtained in (f).
    }\label{fig:intro}
\end{figure*}

The proposed approach is called the Traversability-aware FMM (\emph{Tr-FMM}). It uses as a base tool the Fast Marching Method (FMM) \cite{fmm_sethian} due to its  advantages, which make it well-suited for this problem, as described in the following sections. \emph{Tr-FMM} discretizes the environment to extract information about different regions and their accessibility to the goal. The planner computes the traversability values of these regions based on observed dynamic obstacles, considering both their movements and the estimation of the probability of their dispersion into other areas. These two factors, along with the deviation from the direct path to the goal, are incorporated into the computation of the gradient that guides the robot toward the goal. This process is illustrated in Fig. \ref{fig:intro}.

\subsection{Related works}

\subsubsection{Why FMM?}

Among existing path planners, we selected FMM due to its advantages over other approaches. Sampling-based methods like RRT \cite{rrt} and PRM \cite{prm} are fast but produce suboptimal paths due to random sampling and poor sampling connection. Variants such as RRT* \cite{rrt_star} and FMT \cite{fast_marching_tree} improve path quality by refining connections but still require significant sampling time. IRRT \cite{informed_rrt} and BIT \cite{batch_informed_rrt} introduce heuristics to guide sampling, but their inherent randomness can still lead to imperfect paths, which is undesirable in dynamic environments.

Dijkstra's algorithm \cite{dijkstra}, is one of the most extended methods in path finding obtaining optimal paths. Dijkstra operates on graphs with discrete connections and fixed edge costs, while the Fast Marching Method (FMM) works on regular grids and solves the Eikonal equation to compute arrival times. Moreover, FMM models the propagation of a wavefront, whereas Dijkstra is limited to predefined graph steps. FMM produces more accurate and smoother paths by allowing interpolation, avoiding staircase-like trajectories. It also handles local variations in the cost field more effectively, resulting in more realistic and efficient paths.

A* algorithm is a heuristic extension of Dijkstra’s method, guiding the search using an estimate of the remaining cost to the goal. While A* can be more efficient in practice by reducing the number of explored nodes, it still operates on discrete graphs and its performance depends heavily on the quality of the heuristic. Compared to A*, FMM remains heuristic-free and provides smoother, more accurate paths by working in a continuous domain. This allows FMM to better approximate optimal trajectories in environments with spatially varying costs, where discrete methods like A* may introduce grid-induced artifacts.

Extensions like MH-A* \cite{mhastar} incorporate additional heuristics, such as obstacle avoidance, alongside the goal-directed heuristic. These heuristics are typically computed individually for each obstacle and guide the search away from undesirable regions. In our approach, we use FMM to repel the wavefront propagation from entire groups of moving obstacles, enabling global and smooth avoidance behavior during path computation.

While these methods are typically used to compute the shortest path to the goal, our approach adapts the basic FMM formulation to discretize the environment into regions and evaluate their traversability. This information is then used to generate paths that avoid high-risk areas, especially in densely populated environments.

Summarizing, FMM offers greater utility and flexibility over above-mentioned methods: (1) it enables environment discretization, (2) it captures information from entire sets of cells, such as obstacle groups, and (3) it integrates goal heuristics at a regional level, reducing the search space only to relevant areas.

\subsubsection{Path planning in dynamic environments}

Classic path-planning methods are ineffective in dynamic environments as they do not account for obstacle movement. To anticipate collisions, some approaches integrate social rules into traditional path-planning algorithms. In \cite{social_path_planning_eb}, the Elastic Band (EB) method is modified to incorporate human movement patterns for human-robot co-navigation. The work in \cite{social_path_planning_optimization} employs an optimization-based approach to generate paths that simulate potential human trajectories. In \cite{teachingRRT}, RRT* is adapted by learning cost function weights from human trajectories, enabling the planner to replicate demonstrated behaviors. Depending on the learned context, the computed path may or may not enter human personal space. Similarly, in \cite{social_path_planning_astar}, A* is used to model the spatial relationship between robots and humans over time, applying a Gaussian-based social cost model to predict human movements. FMM has also been applied to social navigation, as in \cite{social_path_planning_fmm}, where an actuation space around humans is used to anticipate their movements. However, their common limitation is considering only a few humans, as analyzing each one individually increases computation time, hindering real-time performance. And, similarly to the reactive navigation methods, mentioned in the previous section, these path planners are more useful when the robot is already moving among people. But, none of them considers explicit avoidance of the crowded regions or at least selecting to move in less-crowded areas when it is possible.

In \cite{fmm_social_navigation}, an FMM-based navigation system incorporates social costs to avoid small groups of people. However, like the previous planning methods, it assumes the robot must navigate within human-populated areas. In contrast, in many scenarios, bypassing densely occupied regions may be a more efficient strategy. \emph{Tr-FMM} avoids these types of regions by computing paths around them whenever possible. When finding a path through completely obstacle-free areas or without excessive detours is impossible, \emph{Tr-FMM} attempts to find a path through less risky regions, where obstacles are fewer or less disruptive.

The Crowd-Based Dynamic Blockages (CBDB) method \cite{cbdb} addresses the problem by identifying and evaluating entire obstacle-populated regions based on density. However, CBDB requires evaluating multiple paths to select the best one, increasing computational costs. \emph{Tr-FMM}, on the other hand, avoids crowded regions directly during the path search stage, eliminating the need for multiple path evaluations. Additionally, \emph{Tr-FMM} not only considers these regions but also accounts for adjacent ones at risk of becoming occupied.

\subsection{Contributions}

This work presents the development of the Traversability-aware FMM (\emph{Tr-FMM})\footnotemark, a path planning method for dynamic environments, and offers the following contributions:
\begin{itemize}
    \item Instead of modeling dynamic obstacles on an individual or small-group basis, our approach discretizes the environment into distinct regions and assigns a traversability value to each. This strategy efficiently captures the overall spatial distribution of obstacles without needing detailed models for every obstacle.
    \item We present a new traversability formulation of regions that quantifies: (1) the alignment of a region with the direct path to the goal; (2) the occupancy level, reflecting both the number of obstacles and their disruptive potential within the region; (3) the estimated probability of obstacles dispersing into adjacent regions.
    \item \emph{Tr-FMM} integrates a traversability map with a velocity map within the standard FMM framework. The traversability map guides the wavefront through regions with lower density and risk, while the velocity map steers the path away from both static and dynamic obstacles.
\end{itemize}

\footnotetext{The code will be available upon publication acceptance.}

The rest of the paper is organized as follows. Section \ref{sec:preliminaries} summarizes the variables used and some basic concepts of the FMM needed for this work. Section \ref{sec:trfmm} describes in detail the developed \emph{Tr-FMM}. Results and discussion are presented in Section \ref{sec:results}, and conclusions in Section \ref{sec:conclusions}.

\section{Preliminaries}\label{sec:preliminaries}

\subsection{Notation}\label{subsec:notation}

The planning surface is the static map of the scenario and is denoted as a grid map represented by $M$. A position in the grid is denoted by $x$ and a set of positions is $\mathbf{x}$. When a position is free of obstacles, it is denoted as $M(x)=1$ and when is occupied by an obstacle, static or dynamic, $M(x)=0$. Similarly $M(\mathbf{x})=0$ represents a set of cells occupied by obstacles, and $M(\mathbf{x})=1$ the positions are free cells. Variable $d(x_i, x_j)$ stands for the distance between the positions $x_i$ and $x_j$. The robot position is encoded as $x_r$ and the goal is $x_g$. We denote the positions of the static and dynamic obstacles as $\mathbf{x}_{so}$ and $\mathbf{x}_{do}$, respectively. Which, obviously accomplish $M(\mathbf{x}_{so})=0$ and $M(\mathbf{x}_{do})=0$. The trajectories traveled by the dynamic obstacles are expressed as $\tau_{do}$. A path is represented by $\pi$, and consists of a sequence of adjacent positions that go from the robot to the goal, $\mathbf{x}_\pi=[x_r,...,x_g]\in\pi$.

\begin{figure*}[tb!]
    \begin{center}
        \includegraphics[width=2.1\columnwidth]{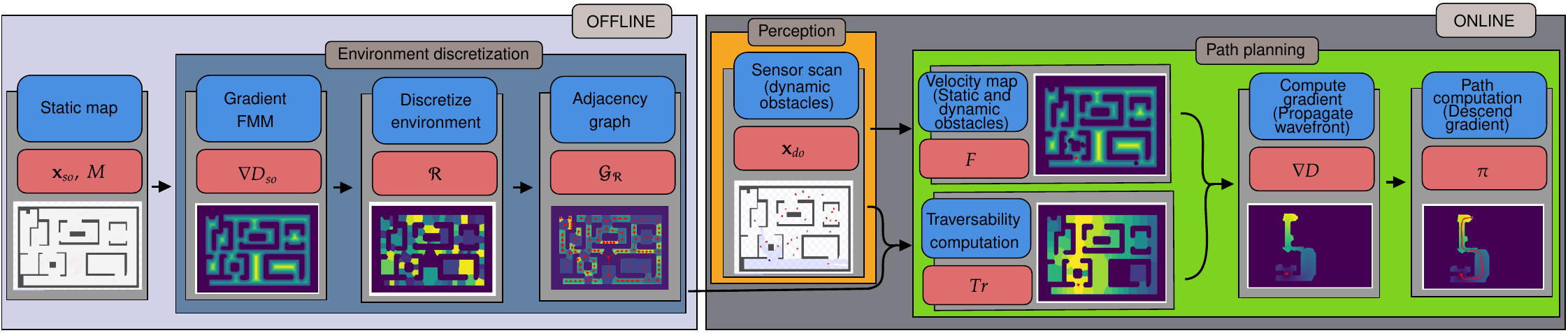}
    \end{center}
    \caption{
        Pipeline of the proposed \emph{Tr-FMM}.
    }
    \label{fig:diagram}
\end{figure*}

\subsection{FMM basics}\label{subsec:fmm}

The Fast Marching Method (FMM) propagates a wavefront $W$ from a source (typically the robot’s position), computing the distance gradient $\nabla D$ via interpolation from neighboring values. Wavefront propagation relies on the velocity map $F$, where $F(x)=1$ in free space and $F(x)=0$ in obstacles. The resulting gradient has no local minima, so descending it from any point (e.g. goal position) leads to the source (the robot's position).

With non-uniform $F$ values (i.e., $F(x)>0$ and $F(x)\neq 1$),  the wavefront spreads faster in areas with higher values of $F$, yielding lower gradient values. This favors to obtain paths through them. FMM also supports multiple source wavefronts, in which case the gradient contains local minima, guiding descent to the nearest source.

These features are exploited in \cite{fmm_planning}, where two gradients are used. First, the wavefront is propagated from the static obstacles, obtaining the distance gradient from these positions, $\nabla D_{so}$. Second, the values of this gradient are saturated, and the velocity is set to $F=\nabla D_{so}$ to maintain a safety distance from obstacles and generate smooth paths. We adopt this idea in our work for: (1) also enhancing safety by maintaining distance from obstacles, (2) increasing maneuverability and escape options near dynamic obstacles, and (3) improving field of view to detect moving obstacles earlier.

\section{Traversability-aware Fast Marching Method}\label{sec:trfmm}

\subsection{Method}\label{subsec:method}

The general overview of the developed method is shown in Fig.\ref{fig:diagram}, where two phases are distinguished. The offline phase preprocesses the environment, taking advantage of the fact that static obstacles remain unchanged. The online phase is responsible for computing the path based on observable changes in the environment.

\textbf{Offline.} First, the method computes the gradient $\nabla D_{so}$ by initializing the wavefront from the static obstacles $\mathbf{x}_{so}$ with uniform propagation, setting $F=M$. This gradient is then used to: (1) discretize the environment into a set of $N$ regions $\mathcal{R}=[\mathcal{R}_1, .. , \mathcal{R}_N]$ and (2) extract relevant information such as region dimensions and distribution, represented as a graph $\mathcal{G}_\mathcal{R}$. Both elements are key to estimate the traversability of these regions. $\mathcal{G}_\mathcal{R}$ provides all possible directions the robot can take to reach the goal. 
If a new static obstacle appears, it is treated as dynamic and avoided, but the predefined regions and their graph remain constant. Therefore, this phase is computed only once and is further described in Sect.\ref{subsec:environment_discretization}.

\textbf{Online.} In this phase the path is replanned in response to dynamic obstacle movement or newly appearing obstacles, registered from the sensor scans. First, the velocity map is set to $F=\nabla D_{so}$ to maintain a safe distance from static obstacles, as explained in Sect.\ref{subsec:fmm}. Then, the values of the gradient at the positions of dynamic obstacles, are subtracted to generate a path that avoids those areas. Doing this, the planner takes into account the exact posisionts of dynamic obstacles in order to avoid them. At the same time, the method computes the traversability of each region, incorporating three criteria, detailed in Sect.\ref{subsec:traversability_cost}: (1) deviation of the region from the direct path to the goal, (2) occupation of the region, assessing how disruptive the dynamic obstacles are within it; and (3) probability of obstacle dispersion into adjacent regions. Both the traversability and velocity maps are then used to propagate the wavefront and compute the distance gradient. Since the traversability map encodes region-level information, it directs the wavefront through regions with higher traversability. The velocity map, on the other hand, ensures avoidance of dynamic obstacles within the selected regions. In summary, the traversability map guides the wavefront through favorable regions, while the velocity map keeps the path clear of obstacles, as discussed in Sect. \ref{subsec:traversability_wavefront}. Finally, the path is obtained by descending the gradient provided by \emph{Tr-FMM}.

\subsection{Environment discretization (Offline)}\label{subsec:environment_discretization}

\begin{figure*}[tb!]
    \begin{center}
        \subfigure[Maxima $\mathbf{x}_{\mathcal{R}}$ of $\nabla D_{so}$]{\includegraphics[width=.25\textwidth]{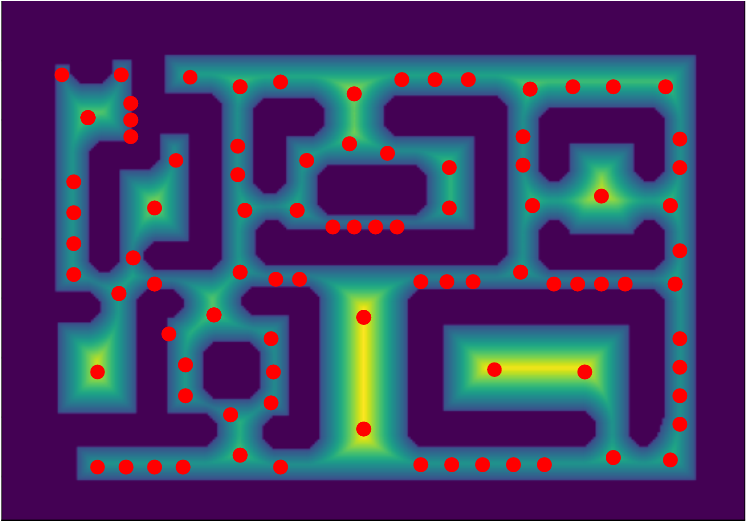}\label{fig:discr_max}} \hfill
        \subfigure[Gradient $\nabla D_R$ from $\mathbf{x}_{\mathcal{R}}$ ]{\includegraphics[width=.25\textwidth]{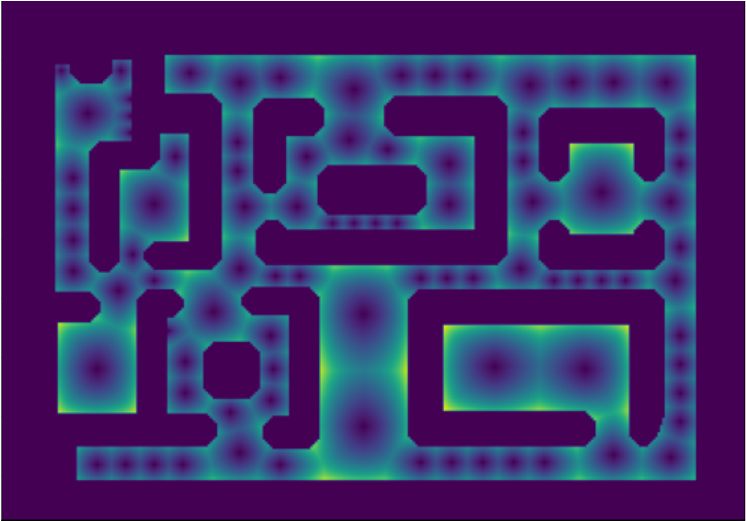}\label{fig:discr_grad}} \hfill
        \subfigure[Regions $\mathcal{R}$]{\includegraphics[width=.25\textwidth]{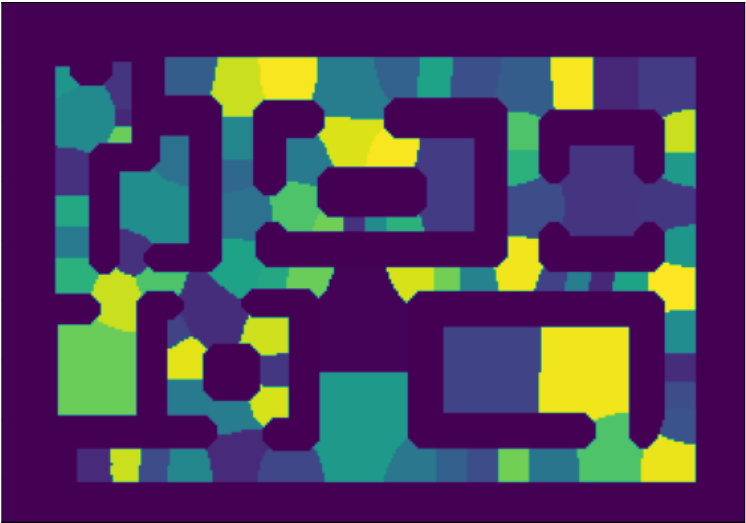}\label{fig:discr_regions}} \hfill
        \subfigure[Adjacency graph $\mathcal{G}_\mathcal{R}$]{\includegraphics[width=.231\textwidth]{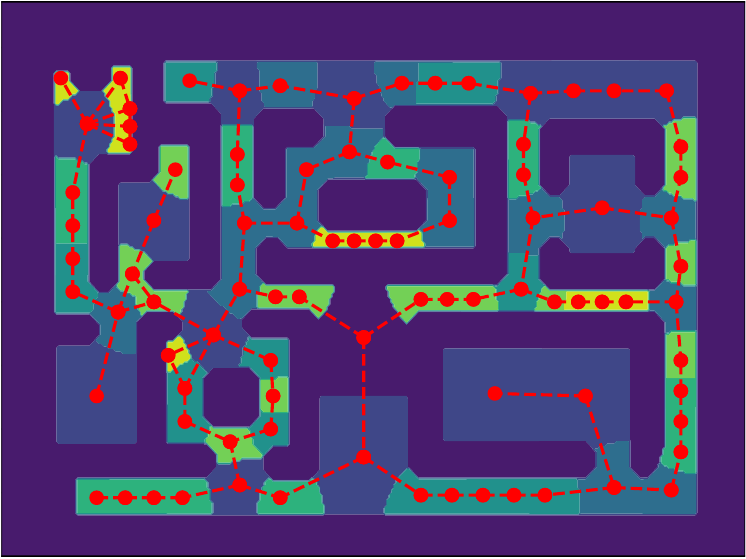}\label{fig:discr_graph}}
    \end{center}
    \caption{
        Environment discretization process. More yellow points represent higher values of the gradient in Fig.\ref{fig:discr_max} and \ref{fig:discr_grad}.
    }
    \label{fig:discr}
\end{figure*}

The first step of our approach is to partition the environment into a discrete set of regions. This discretization crucially reduces the number of possible paths that must be evaluated, thereby significantly lowering the computational cost of the planning process. Using just cells as positions, a value must be assigned to every position in the grid map, resulting in an enormous number of potential paths to search and evaluate. In contrast, our approach assigns a single traversability value to each region, uniformly applied to all cells within that region, effectively narrowing the search space.

The environment discretization process is performed using FMM and illustrated in Fig.\ref{fig:discr}. First, the algorithm iteratively identifies the maxima $\mathbf{x}_\mathcal{R}$ of the gradient $\nabla D_{so}$ (Fig.\ref{fig:discr_max}), which serve as the origins of the regions. When a new maximum $x_{\mathcal{R}_i} \in \mathbf{x}_{\mathcal{R}}$ is found, the algorithm selects its corresponding gradient value $\nabla D_{so}(x_{\mathcal{R}_i})$, which represents the minimum distance to the nearest static obstacle. To prevent selecting multiple positions within the same area, all gradient values $\nabla D_{so}$ within the range of $\nabla D_{so}(x_{\mathcal{R}_i})$ are subtracted.

Next, the wavefront propagates from $\mathbf{x}_{\mathcal{R}}$, computing the gradient (Fig.\ref{fig:discr_grad}) while simultaneously labeling the traversed positions to define the regions $\mathcal{R}$ (Fig.\ref{fig:discr_regions}). The boundaries between regions are established where wavefronts from different origins collide, as seen in Figs.\ref{fig:discr_grad} and \ref{fig:discr_regions}. Finally, the region graph $\mathcal{G}_\mathcal{R}$ is constructed by identifying adjacencies between regions, as seen in Fig.\ref{fig:discr_graph}.

Many works in the robotics literature focus on the topic of scenario partitioning. In \cite{topometric_segmentation}, a topological map is extracted to identify precise areas in both structured and unstructured environments for decision-making or navigation. More advanced hierarchical partitioning methods, which divide the scenario into well-defined spaces such as halls, rooms, or corridors, have been proposed in \cite{hierarchical_topometric_segmentation} and \cite{sgraphs}. However, these approaches aim to identify regions in a way that is interpretable for humans—e.g., as rooms or corridors. While such information may be useful for predicting areas where humans (or dynamic obstacles) are more likely to appear, it does not necessarily provide information about their current movement, particularly in terms of how disturbing or disruptive they are for the robot.

As previously described and shown in Fig.\ref{fig:discr_regions}, the regions obtained with our method are shaped according to the surrounding static obstacles. In this way, regions encode the actual free space available when the robot is within them. This characteristic is key when dynamic obstacles move through the environment, as it allows the system to extract information to quantify the occupation of each region—that is, how disruptive those dynamic obstacles are—as described in the following section.

\subsection{Traversability-based gradient computation (Online)}\label{subsec:trfmm_gradient}

First, the method computes and assigns traversability values to the regions to guide the wavefront through more favorable areas. Then, the wavefront propagates with a higher priority given to regions with greater traversability, ensuring the path avoids crowded areas while reducing deviation from the direct route to the goal.

\subsubsection{Regions traversability computation}\label{subsec:traversability_cost}
\begin{figure*}[tb!]
    \begin{center}
        \subfigure[]{{\includegraphics[width=.19\textwidth]{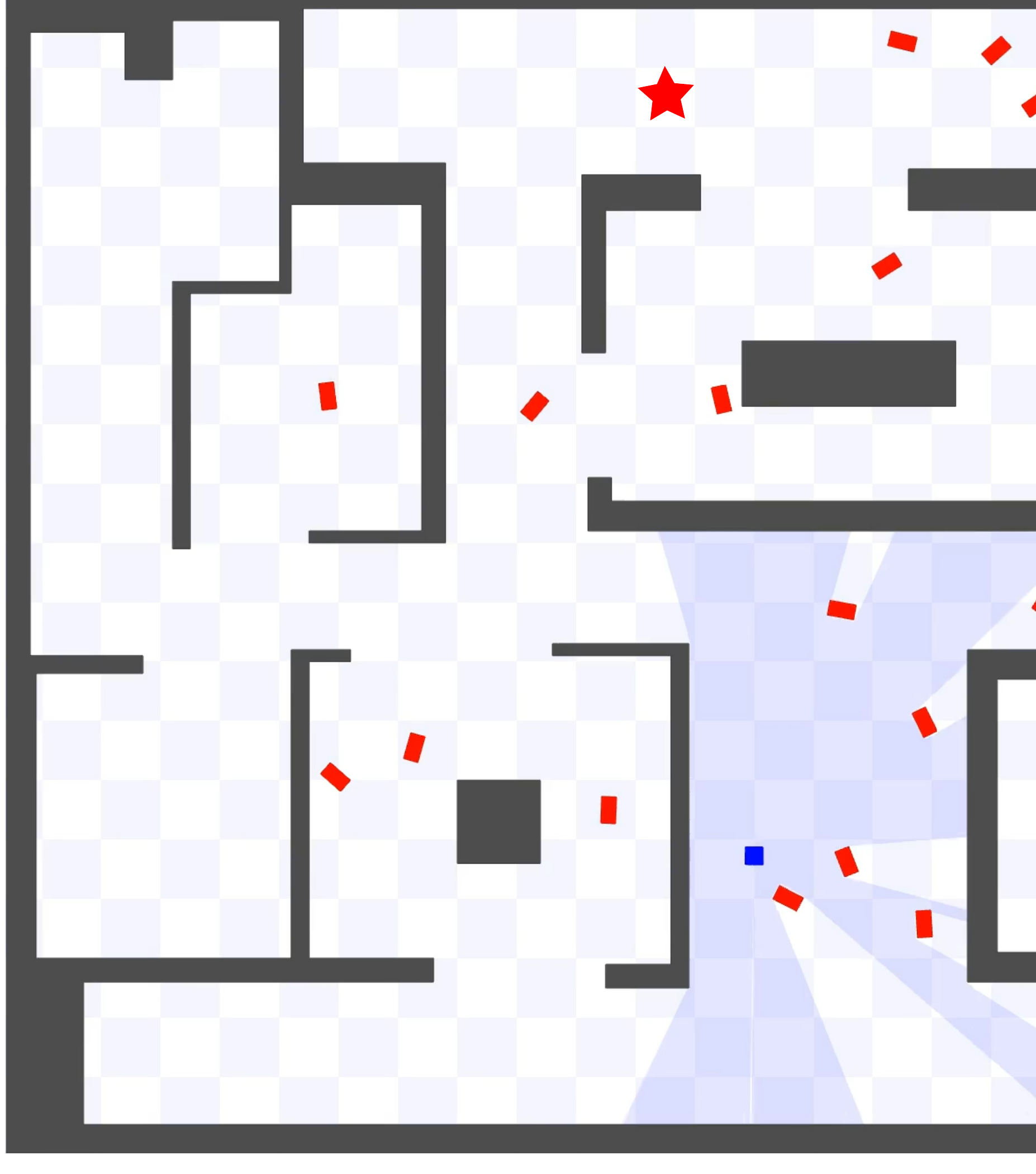}}\label{fig:costs_scen}} \hfill
        \subfigure[ ]{\includegraphics[width=.18\textwidth]{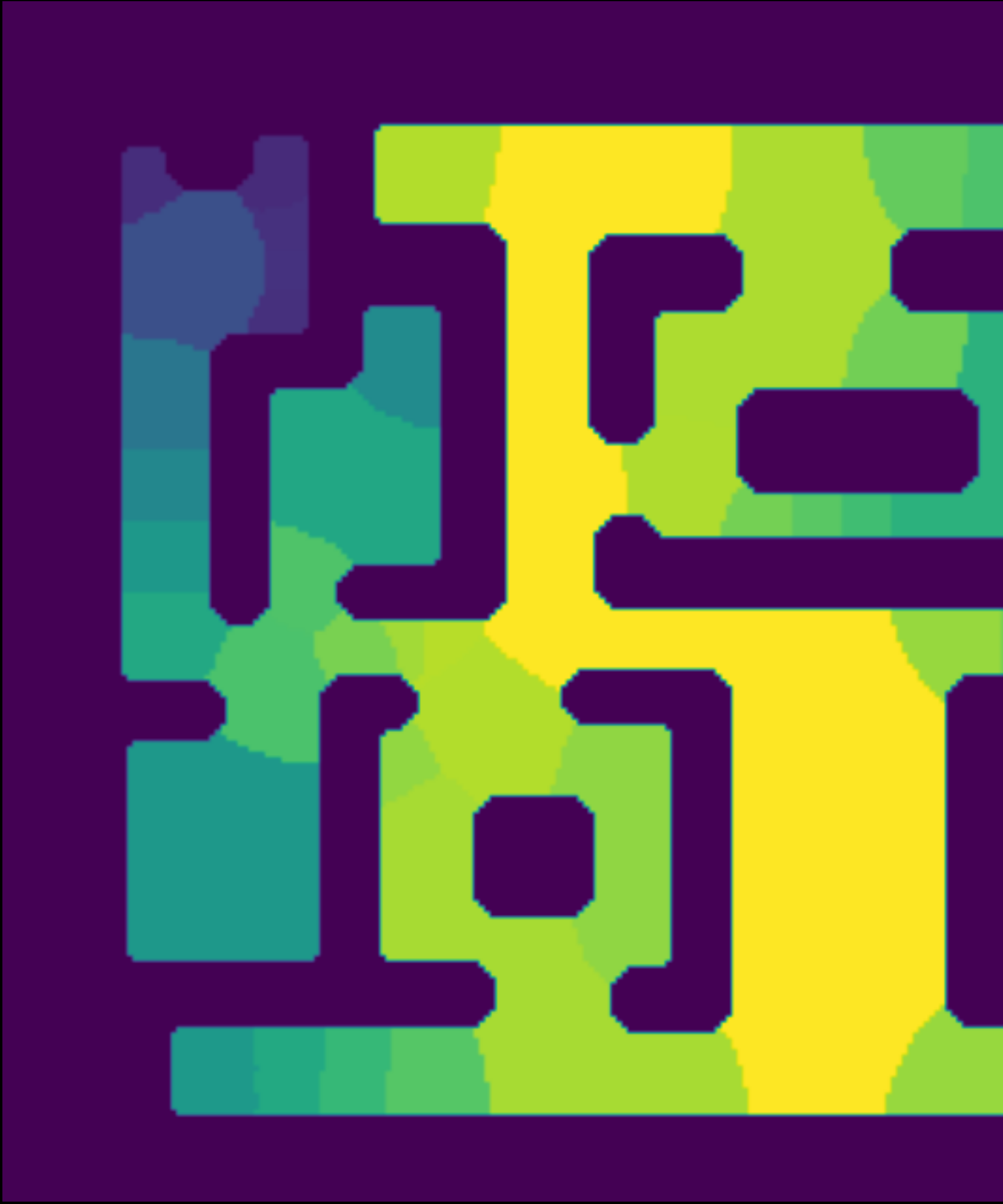}\label{fig:costs_deviation}} \hfill
        \subfigure[]{\includegraphics[width=.18\textwidth]{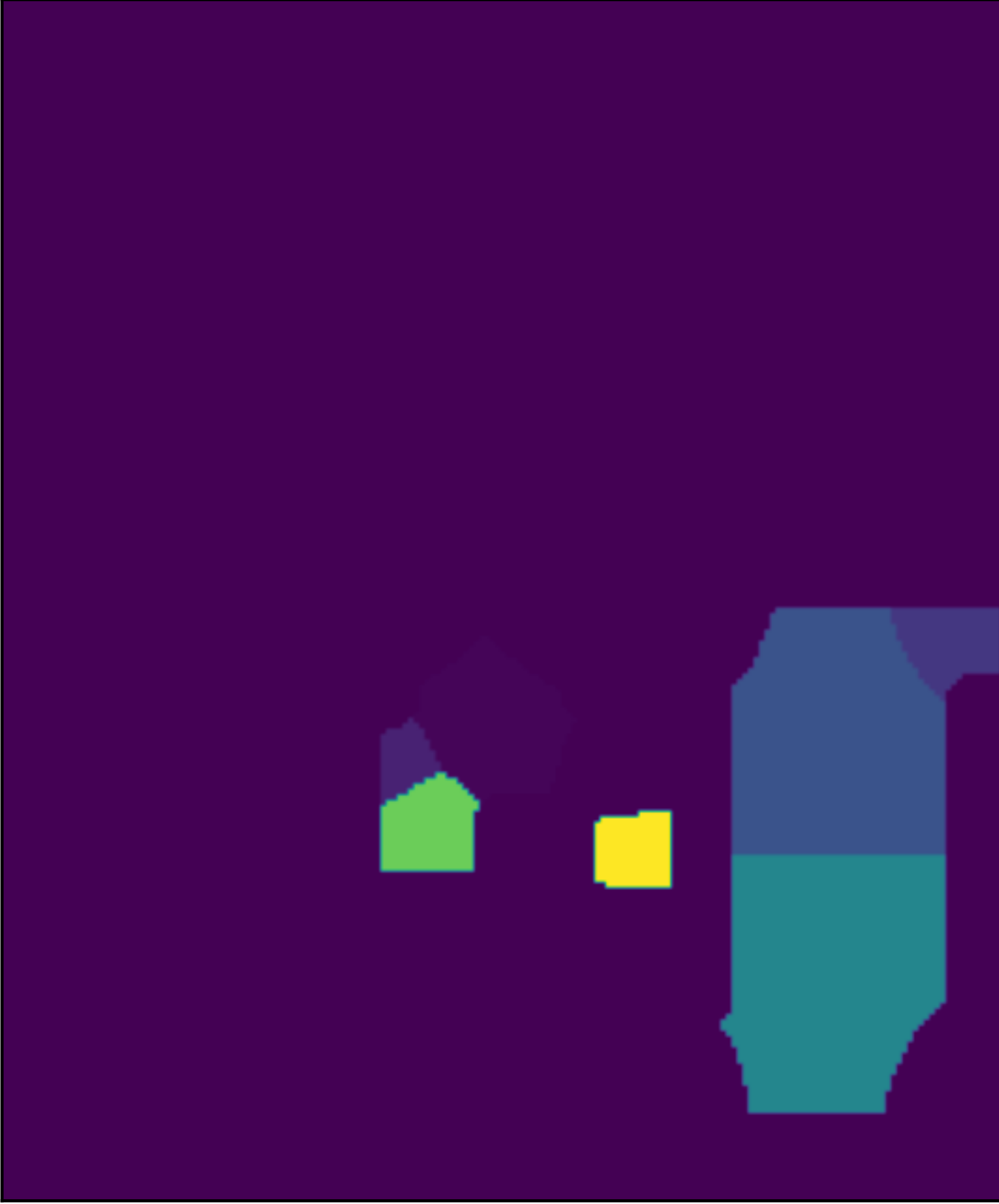}\label{fig:costs_occupation}} \hfill
        \subfigure[]{\includegraphics[width=.18\textwidth]{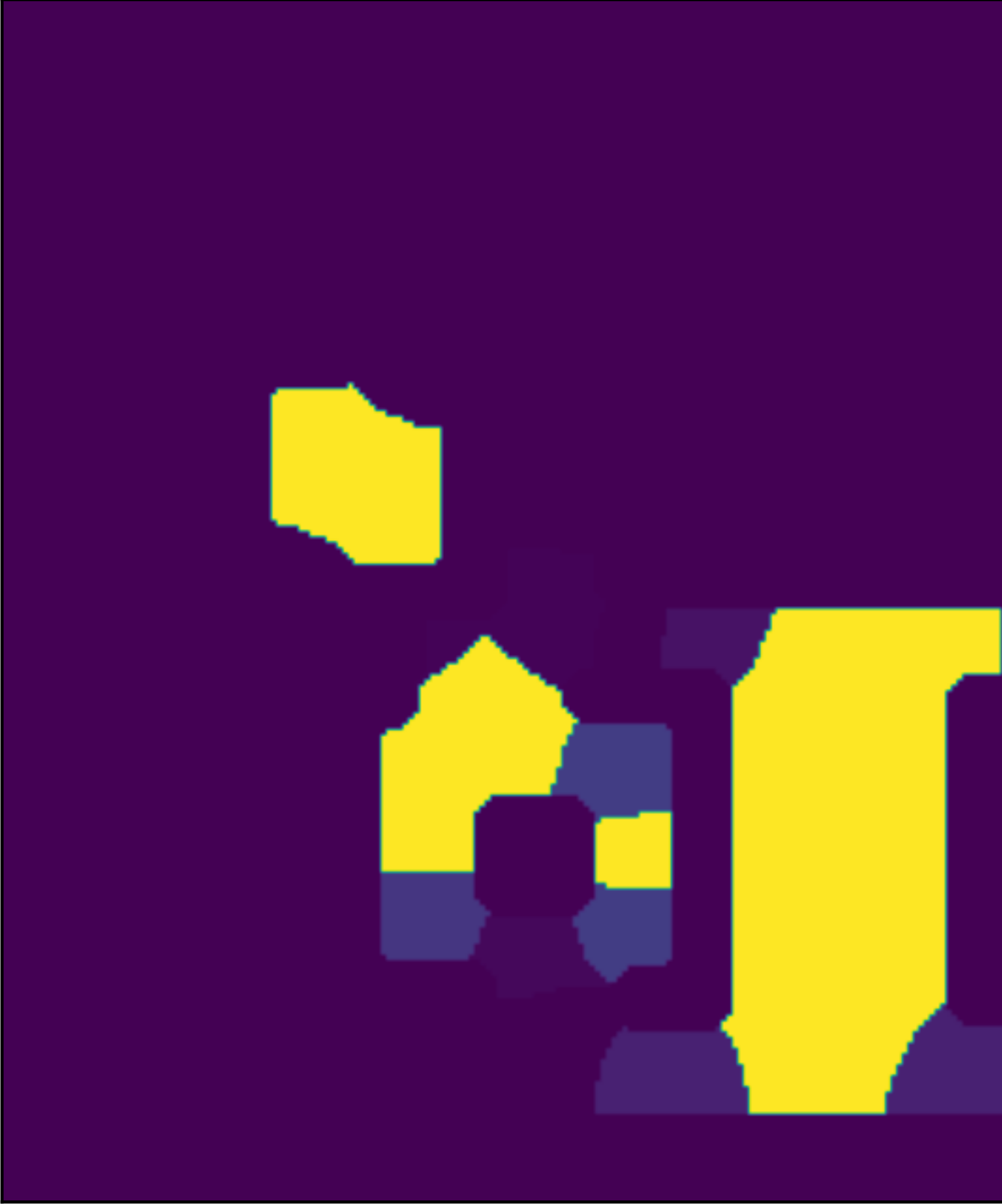}\label{fig:costs_dispersion}} \hfill
        \subfigure[]{\includegraphics[width=.18\textwidth]{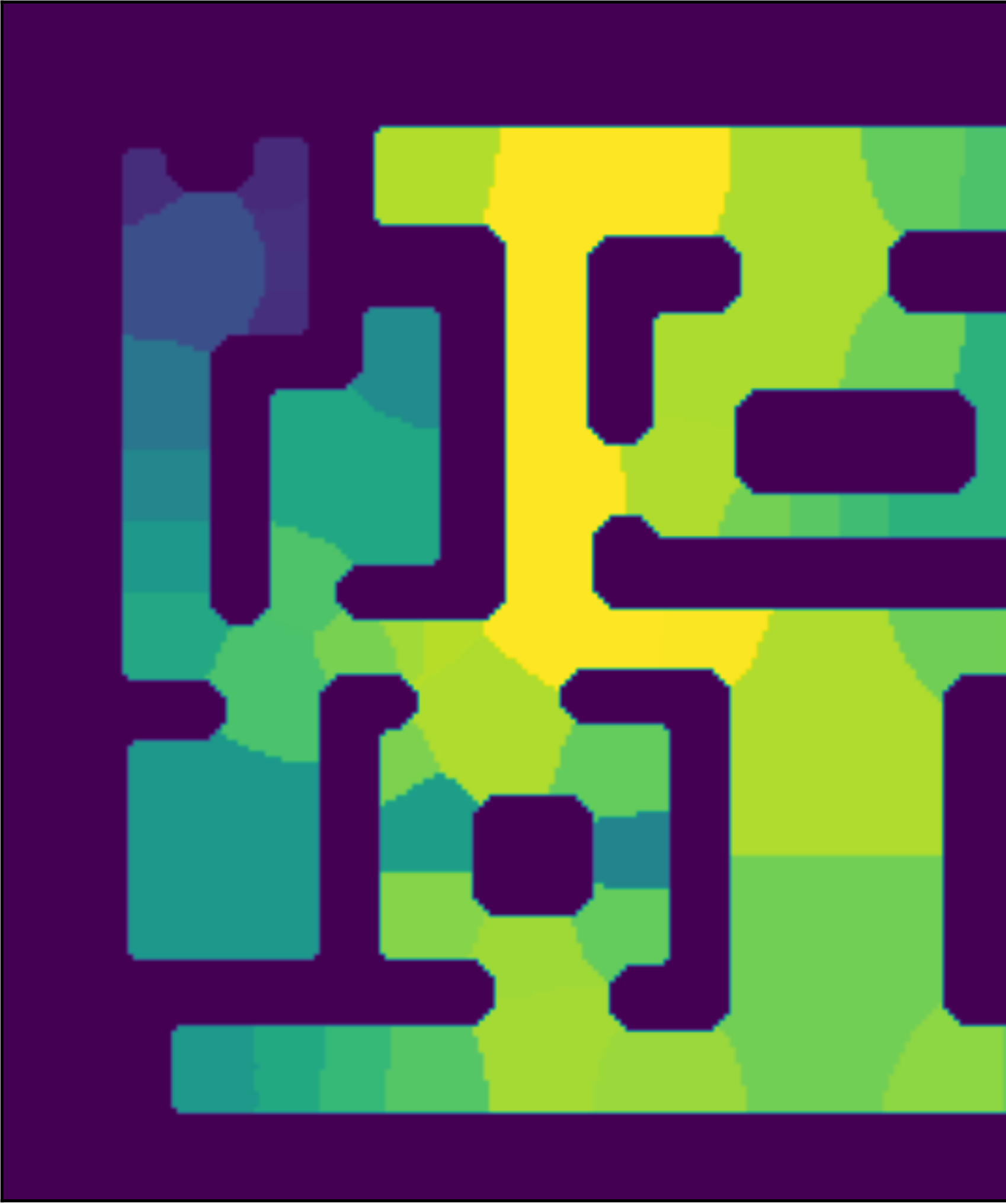}\label{fig:costs_traversability}}
    \end{center}
    \caption{
        The variables involved in traversability computation. (a) Scenario (goal illustrated with red star), (b) Deviation, (c) Occupation, (d) Dispersion, (e) Traversability. Red star in (a) depicts the goal. Yellow color in (b)-(e) correspond to maximum values. The traversability in (e) is obtained from the previous variables using eq.\ref{eq:traversability}.
    }
    \label{fig:traversability_costs}
\end{figure*}

\begin{figure*}[tb!]
    \begin{center}
        \subfigure[Deviation and dispersion]{\includegraphics[width=0.8\columnwidth]{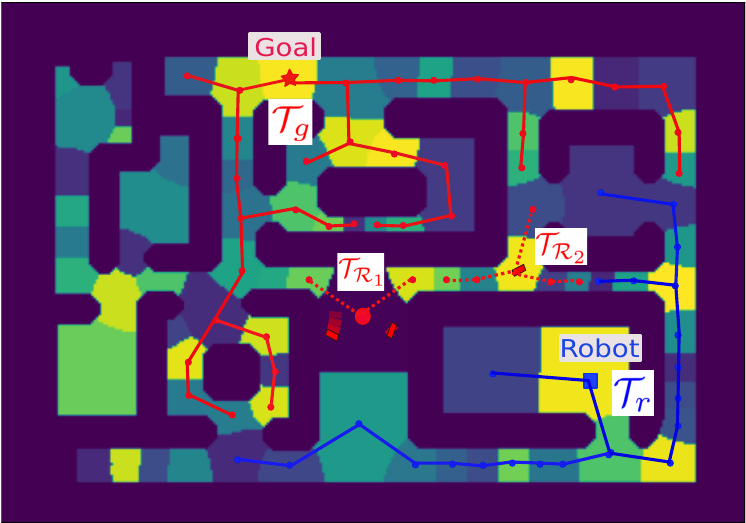}\label{fig:disp_expl}} \hspace{10mm}
        \subfigure[Occupation]{\raisebox{0.2cm}{\includegraphics[width=0.9\columnwidth]{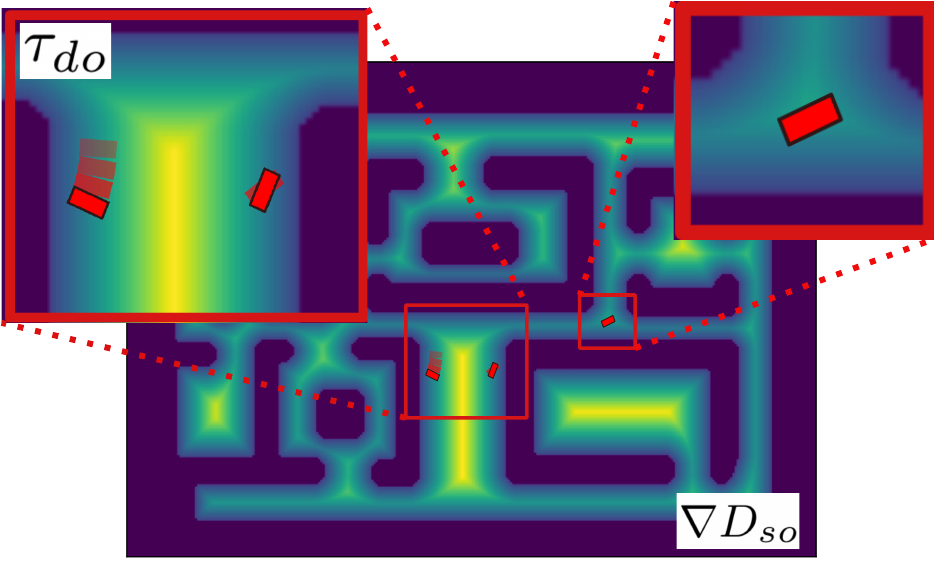}}\label{fig:occup_expl}}
    \end{center}
    \caption{
        Illustrative explanation of traversability variables: (a) Deviation and dispersion, with only representative edges of the trees $\mathcal{T}_r$ (solid blue), $\mathcal{T}_g$ (solid red), and $\mathcal{T}_\mathcal{R}$ (dashed red) shown to avoid overlap. The representative edges of $\mathcal{T}_r$ and $\mathcal{T}_g$ are the closest to their origin/root, while those of $\mathcal{T}_\mathcal{R}$ correspond to regions penalized enough by dispersion to be avoided. (b) Occupation, computed using $\nabla D_{so}$ and obstacle trajectories $\tau_{do}$ in eq.\ref{eq:occupation}. The rightmost region has higher occupation despite a single static obstacle, as its small size and central obstacle position make it more obstructive.
    }
    \label{fig:maps_expl}
\end{figure*}

involving three main factors, which are described in detail below and illustrated in Figs.\ref{fig:traversability_costs} and \ref{fig:maps_expl}.

\begin{itemize}
    \item[--] \emph{Deviation} (Fig.\ref{fig:costs_deviation})
\end{itemize}

Since the regions may have dynamic obstacles in movement and will be potentially surrounded, the first factor to consider is the admissible deviation from the direct path to the goal. To address this, we compute the distances from both the robot and the goal to each region. Using $\mathcal{G}_\mathcal{R}$, we generate tree graphs from the robot’s region ($\mathcal{T}_r$) and the goal’s region ($\mathcal{T}_g$), respectively, see, Fig.\ref{fig:disp_expl}. These trees are constructed using Dijkstra's algorithm \cite{dijkstra}, with costs based on distances between regions, approximated by edge lengths in $\mathcal{G}_\mathcal{R}$. Consequently, the distance from the robot to a given region $i$ is the branch length in $\mathcal{T}_r$, denoted as $D_{ri}$, while the distance from the goal’s region to region $i$ is $D_{gi}$. Thus, the total distance from the robot to the goal through region $i$ is given by $D_{rig}=(D_{ri} + D_{gi})$. As shown in Fig. \ref{fig:costs_deviation}, regions that provide a more direct route to the goal have higher values, ensuring they are prioritized during wavefront propagation. Note that all distances between regions, approximated by the lengths of the edges, are used solely to estimate the possible travel distances of the robot and do not represent the actual distance traveled by it.

\begin{itemize}
    \item[--] \emph{Occupation} (Fig.\ref{fig:costs_occupation})
\end{itemize}

It measures the presence and movement patterns of dynamic obstacles within the region, assessing how much they could hinder the robot's motion. To quantify this variable, we use the gradient of static obstacles, $\nabla D_{so}$, as defined in eq.\ref{eq:occupation}:

\begin{equation}\label{eq:occupation}
    O_{i} = \frac{\overline{\nabla D_{so}(\mathbf{x}_{\tau_i})}}{\overline{\nabla D_{so}(\mathbf{x}_{i})}} \ , \ \mathbf{x}_{\tau_i} \in \tau_{do_i} \in \mathcal{R}_i \ , \ \mathbf{x}_i \in \mathcal{R}_i
\end{equation}

In this expression, we analyze the movement of obstacles by extracting the normalized values of the gradient along their trajectories, $\overline{\nabla D_{so}(\mathbf{x}_{\tau_i}})$, as seen in Fig.\ref{fig:occup_expl}. By normalizing these values with respect to the entire region, $\overline{\nabla D_{so}(\mathbf{x}_{i}})$, we capture not only the proportion of the region occupied by obstacles but also the specific locations where they are moving. This distinction is crucial: an obstacle near a static obstacles is less disruptive than one moving freely in the middle of a region, where it poses a greater risk to the robot’s navigation. As illustrated in Fig.\ref{fig:costs_occupation}, smaller regions with an obstacle positioned in the center have a higher occupation cost and therefore present a greater risk when traversed. Conversely, larger regions, even if they contain more obstacles, are less risky for the robot to navigate through.

\begin{itemize}
    \item[--] \emph{Dispersion} (Fig.\ref{fig:costs_dispersion})
\end{itemize}

Since this work deals with dynamic obstacles, not only occupied regions pose a risk for the robot, but also adjacent ones, as obstacles may disperse into nearby regions by the time the robot reaches them. We  estimate the probability of obstacles spreading to other regions over time. This way, dispersion is used to repel the path to the goal not only from the currently occupied regions, but also from adjacent ones, thereby reducing the risk of encountering and colliding with a dynamic obstacle.

First, in eq.\ref{eq:dynamism}, we compute the dynamism of the obstacles in region $i$ by measuring the proportion of the region covered by moving obstacles:

\begin{equation}\label{eq:dynamism}
    P_{do_i} =  \frac{A_{\tau_{do_i}}}{A_i}
\end{equation}

where $A_{\tau_{do_i}}$ denotes the area covered by  the trajectories of the obstacles in region $i$ measured in number of grid positions, and $A_i$ is the area of region $i$. This value represents the probability of obstacles in region $i$ leaving that region. To determine their potential movement, we compute trees for all obstacle-containing regions to all the regions as described above, $\mathcal{T}_\mathcal{R} = [\mathcal{T}_{\mathcal{R}_1}, .. , \mathcal{T}_{\mathcal{R}_k}]$, with $k$ representing the number of regions with obstacles. The probability of the obstacles of region $j$ to be dispersed reaching region $i$ is:

\begin{equation}\label{eq:dispersion}
    P^j_{do_i} =  \frac{A_{\tau_{do_i}}}{A_j}P_{do_i}
\end{equation}

To assess whether a region is at risk of being occupied by obstacles upon the robot's arrival, we compute the distance from obstacle-containing region $j$ to some region $i$, denoted as $D^j_{do_i}$. As with deviation, these distances are approximated by the lengths of the consecutive edges along the branches of the tree starting from region $j$. Since the precise movement patterns of obstacles are unknown, we assume they may linger within intermediate regions before reaching $i$. Thus, we introduce a penalization factor to adjust the distance, as formulated in eq.\ref{eq:dispersion_dist}, representing the estimated distance $D^+_{do_{i}}$ from the nearest obstacle-containing region to $i$:\begin{equation}\label{eq:dispersion_dist}
    D^+_{do_{i}} =  D^j_{do_i}*(1+P^j_{do_i}) \ | \ j \in \{1..N\}\ 
\end{equation}

\begin{itemize}
    \item[--] \emph{Traversability} (Fig.\ref{fig:costs_traversability})
\end{itemize}

\begin{equation}\label{eq:traversability}
    Tr_i = \left\{
    \begin{array}{rr}
        D_{rig}  , & D_{ri} < D^+_{do_{i}} \\
        D_{rig}*(1-O_i*P^+_{do_i})  , & D_{ri} \geq D^+_{do_{i}}
    \end{array}
    \right.
\end{equation}

The traversability value of a region $i$ is summarized in eq.\ref{eq:traversability}, incorporating the previously explained variables. If the estimated distance of the robot to a region $i$ is shorter than the estimated distance of the obstacles to $i$, the robot is expected to arrive first, making $i$ a non-risky region, so only the distance component, $D_{rig}$, contributes to the traversability value. The planner currently assumes obstacles move at a similar velocity to the robot, using distances, although an alternative approach could involve using time and tracking obstacle speeds, which is beyond the scope of this work.

Conversely, if the distance from the robot to region $i$ is greater, obstacles may arrive first, making $i$ a risky region. The distance to the goal through $i$ is then penalized based on the probability of obstacles reaching $i$, as computed in eqs.\ref{eq:occupation} and \ref{eq:dispersion}. Here $P^+_{do_i}$, represents the probability of obstacles from the closest region to $i$ reaching $i$. Eq.\ref{eq:traversability} also introduces an adaptive decision margin: when the robot is far from the goal, multiple paths of similar length exist, allowing the planner to favour obstacle-free or low-risk regions. When the robot is close to the goal, avoiding occupied regions requires larger detours, so paths through such regions are only penalized if they are highly occupied.

In this way, the traversability defined in eq.\ref{eq:traversability} integrates the three variables—deviation to the goal, occupation, and potential dispersion of obstacles—into a single measure, working in synergy. These components are interdependent and must be considered jointly to ensure a meaningful and effective solution. If the occupation component is removed, the planner will compute the direct path to the goal, essentially reverting to the basic FMM solution. If the dispersion is ignored, \emph{Tr-FMM} will only avoid currently occupied regions, potentially increasing the collision risk. And the deviation from the goal is required in the computation process to not excessively repel the path from the regions occupied by dynamic obstacles with the deviation component, avoiding this way excessive detours.

The selection between circumventing dense obstacle regions or traversing through them is not treated as two distinct strategies but emerges naturally from the traversability expression. Regions with higher traversability values indicate safer or more efficient routes, leading the planner to prefer paths through those regions. Conversely, regions with lower traversability values are avoided unless detouring would cause excessive deviation.


\subsubsection{Traversability-based wavefront propagation}\label{subsec:traversability_wavefront}

\begin{figure*}[tb!]
    \begin{center}
        \subfigure[Traversability $Tr$]{\includegraphics[width=.3\textwidth]{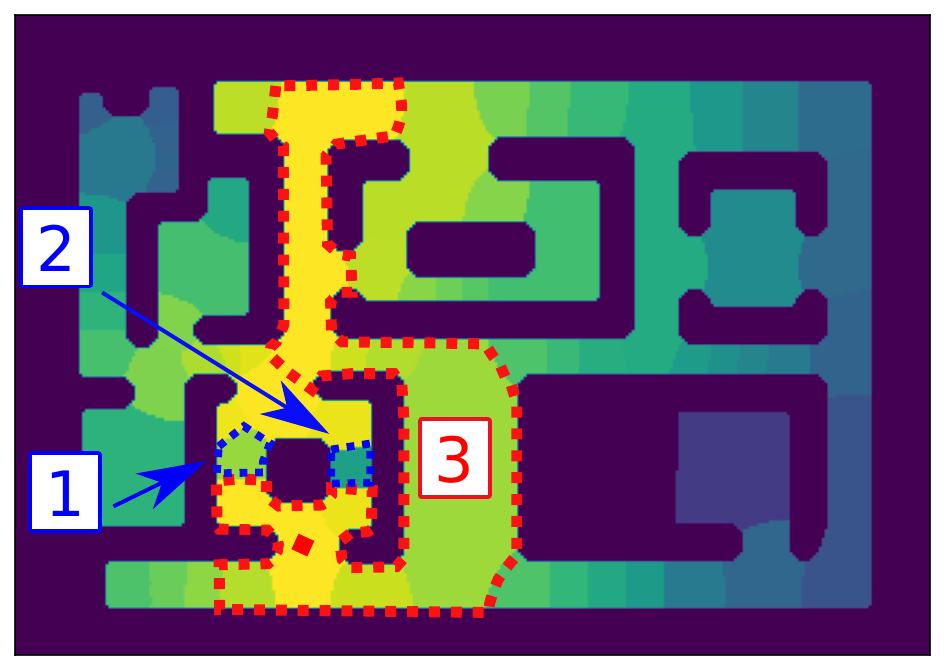}\label{fig:maps_traversability}} \hspace{5mm}
        \subfigure[Velocity $F$]{\includegraphics[width=.3\textwidth]{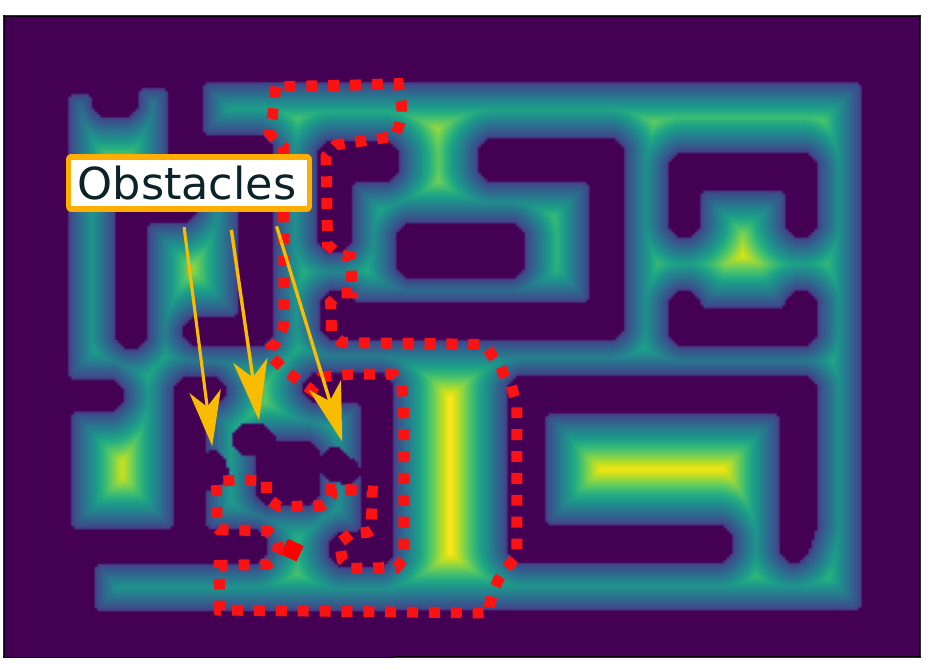}\label{fig:maps_velocity}} \hspace{5mm}
        \subfigure[Gradient]{\includegraphics[width=.3\textwidth]{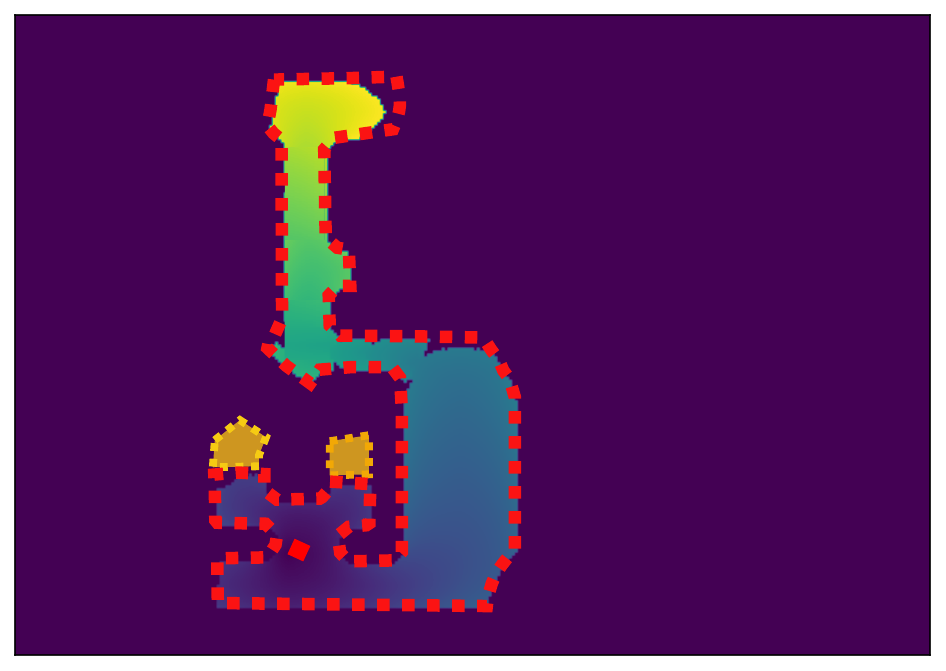}\label{fig:maps_gradient}}
    \end{center}
    \caption{
        Two maps employed by \emph{Tr-FMM} to propagate the wavefront and compute the gradient. The position of the robot is depicted with red square. In \ref{fig:maps_traversability}, after detecting obstacles, in regions 1 and 2, the traversability map guides the wavefront through regions (3), with higher traversability. The velocity map of \ref{fig:maps_velocity} directs the wavefront away from static obstacles and around dynamic ones to avoid collisions inside the regions. The resulting gradient to obtain the path is depicted in \ref{fig:maps_gradient}.
    }
    \label{fig:maps}
\end{figure*}

\emph{Tr-FMM} propagates the wavefront from the robot's position to the goal through regions with higher traversability, as computed in Sect.\ref{subsec:traversability_cost}. As described earlier, the standard FMM uses the velocity map $F$ to accelerate or slow down wavefront propagation, sorting the wavefront positions $W$ in ascending order based on the gradient value $\nabla D$ from the source.
\begin{equation}\label{eq:wavefront}
    W'=
    \begin{cases} 
        \{ x_1, .., x_{i-1}, x^+, x_i, .., x_n \}, & C(x^+, x_i) = 1 \\
        W \cup \{ x^+ \}, & C(x^+, x_i) = 0
    \end{cases}
\end{equation}\begin{equation}\label{eq:wavefront_condition}
    \begin{array}{lr}
        C(x^+, x_i) = \left[ Tr(x^+) > Tr(x_i) \right] \lor \\
        \lor \left[ \ [ Tr(x^+) = Tr(x_i) ] \land [ \nabla D(x^+) < \nabla D(x_i) ] \ \right]
    \end{array}
\end{equation}


In contrast, \emph{Tr-FMM} modifies this sorting process and the insertion of new nodes $x^+$ into $W$, eq.\ref{eq:wavefront}, by incorporating a condition that considers both velocity and traversability, as defined in eq.\ref{eq:wavefront_condition} and illustrated in Fig.\ref{fig:maps}. If eq.\ref{eq:wavefront_condition} is accomplished, new nodes are inserted with priority given first to regions with higher traversability and, second, to positions with the same priority but a lower gradient value. This ensures that region traversability is always prioritized over distance. If eq.\ref{eq:wavefront_condition} is not accomplished, the new nodes append at the end of the ordered set $W$.

Since traversability already accounts for the distance to the goal (eq.\ref{eq:traversability}), the method naturally discourages deviations from the direct path. However, when occupation and dispersion risks heavily penalize a region’s traversability—as seen with the two obstacle-filled regions above the robot in Fig.\ref{fig:maps_traversability}—the wavefront is redirected through an alternative, obstacle-free route to compute the gradient of Fig.\ref{fig:maps_gradient}, as discussed in Sect.\ref{subsec:traversability_cost}.

\subsection{Complexity}\label{sec:complexity}

The complexity of the proposed method consists of two parts: discretization, performed once \textit{offline}, and wavefront propagation, which occurs every time the path is replanned \textit{online}. Discretization (Sect.\ref{subsec:environment_discretization}) requires two executions of FMM: one to obtain  $\nabla D_{so}$ and another to compute $\nabla D_\mathcal{R}$. The complexity of FMM is $O(N_f log N_f)$ in the worst case \cite{fmms}, where $N_f$ is the number of free cells in the grid obtained as $N_f=|\mathbf{x}| \ | \ M(\mathbf{x})=0$. Searching all the maxima requires $N_f(N_r+1)$ iterations, where $N_r$ is the number of regions. But, since $N_r$ is constant and much lower than $N_f$, the complexity simplifies to $O(N_f)$. Thus, the complexity of discretization is: $O(N_f log N_f) + O(N_f) = O(N_f log N_f)$.

Traversability computation (Sect.\ref{subsec:traversability_cost}), requires $N_k+2$ executions of Dijkstra's algorithm, with complexity $O(N_r log N_r)$, where $N_k$ is the number of tree computations from regions with obstacles and other two tree calculations are from the regions of the robot and goal. As $N_k \leq N_r$, the number of executions is proportional to $N_r$, but $N_r$ is constant. So, the complexity of this part is: $O(N_r log N_r) = O(1)$, because $log N_r$ of a small number is constant. Wavefront propagation (Sect.\ref{subsec:traversability_wavefront}), has the complexity of FMM: $O(N_f log N_f)$. Therefore, the overall complexity of \emph{Tr-FMM} is $O(N_f log N_f)$  for both the preprocessing and gradient computation steps. The gradient descent step is negligible.

\section{Results}\label{sec:results}

\subsection{Simulations}

\begin{figure*}[tb!]
    \begin{center}
        \subfigure[Scenario]{\includegraphics[width=0.3\textwidth]{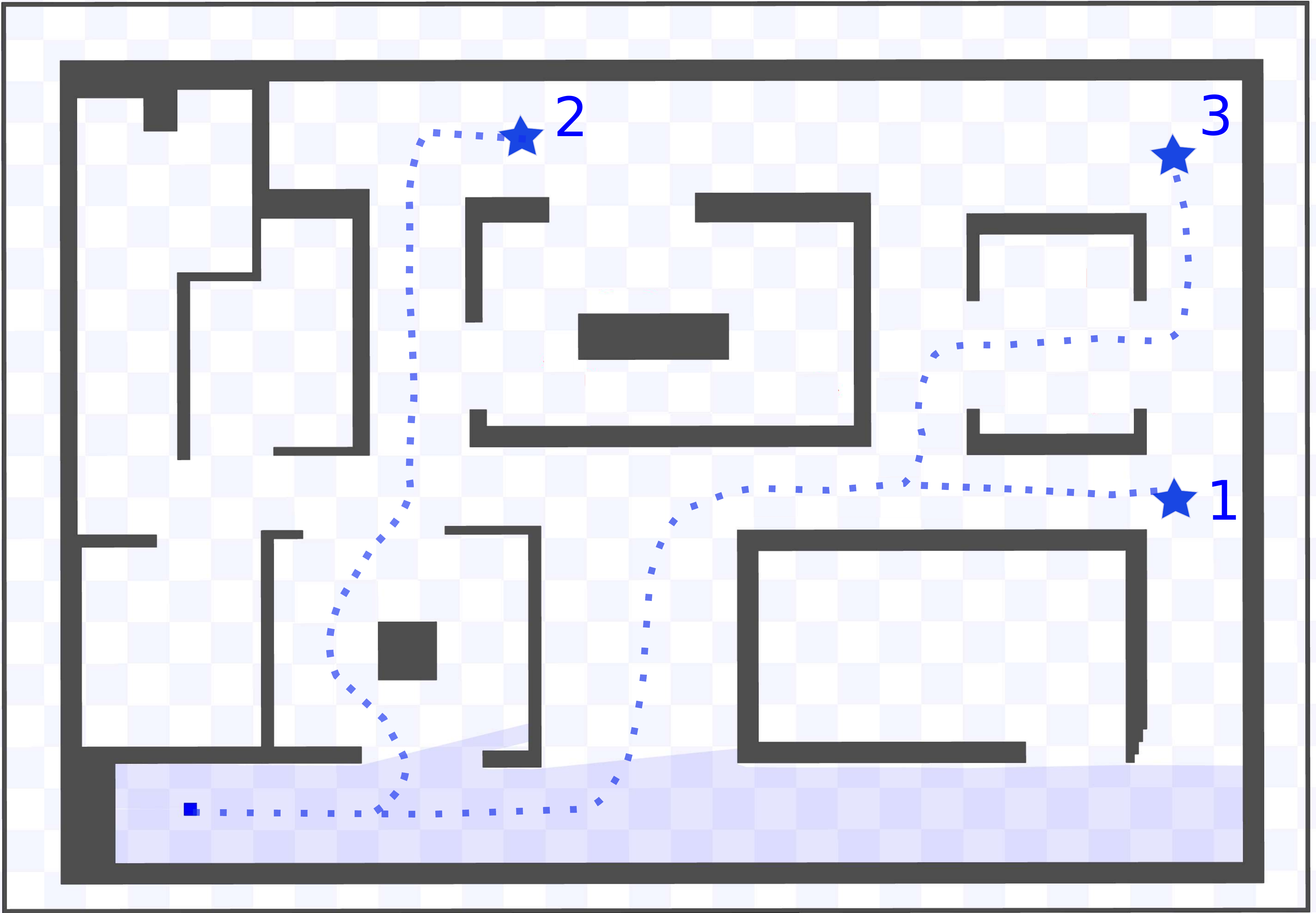}\label{fig:scen_map}} \hspace{5mm}
        \subfigure[Dense]{\includegraphics[width=0.3\textwidth]{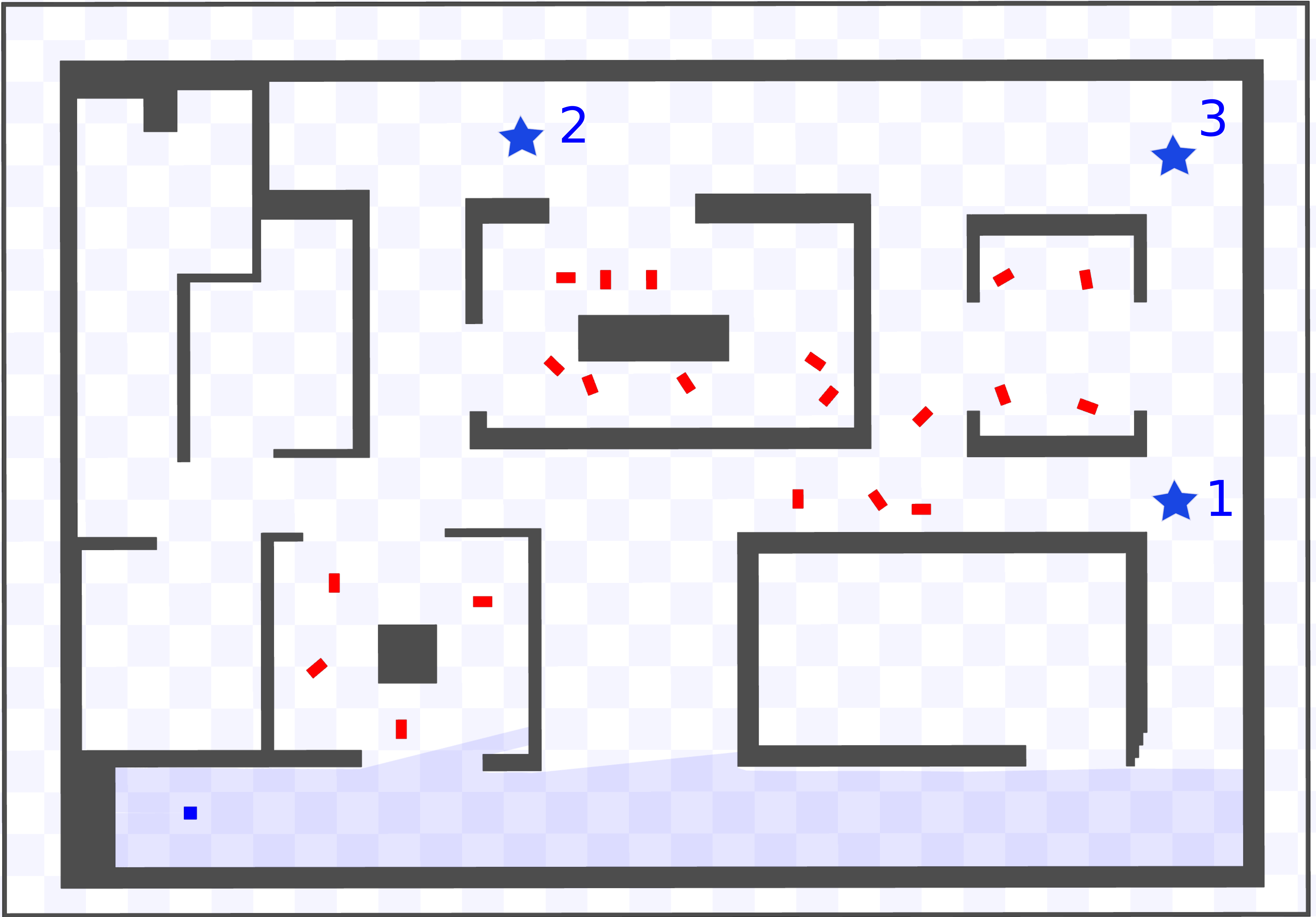}\label{fig:scen_dense}} \hspace{5mm}
        \subfigure[Disperse]{\includegraphics[width=0.3\textwidth]{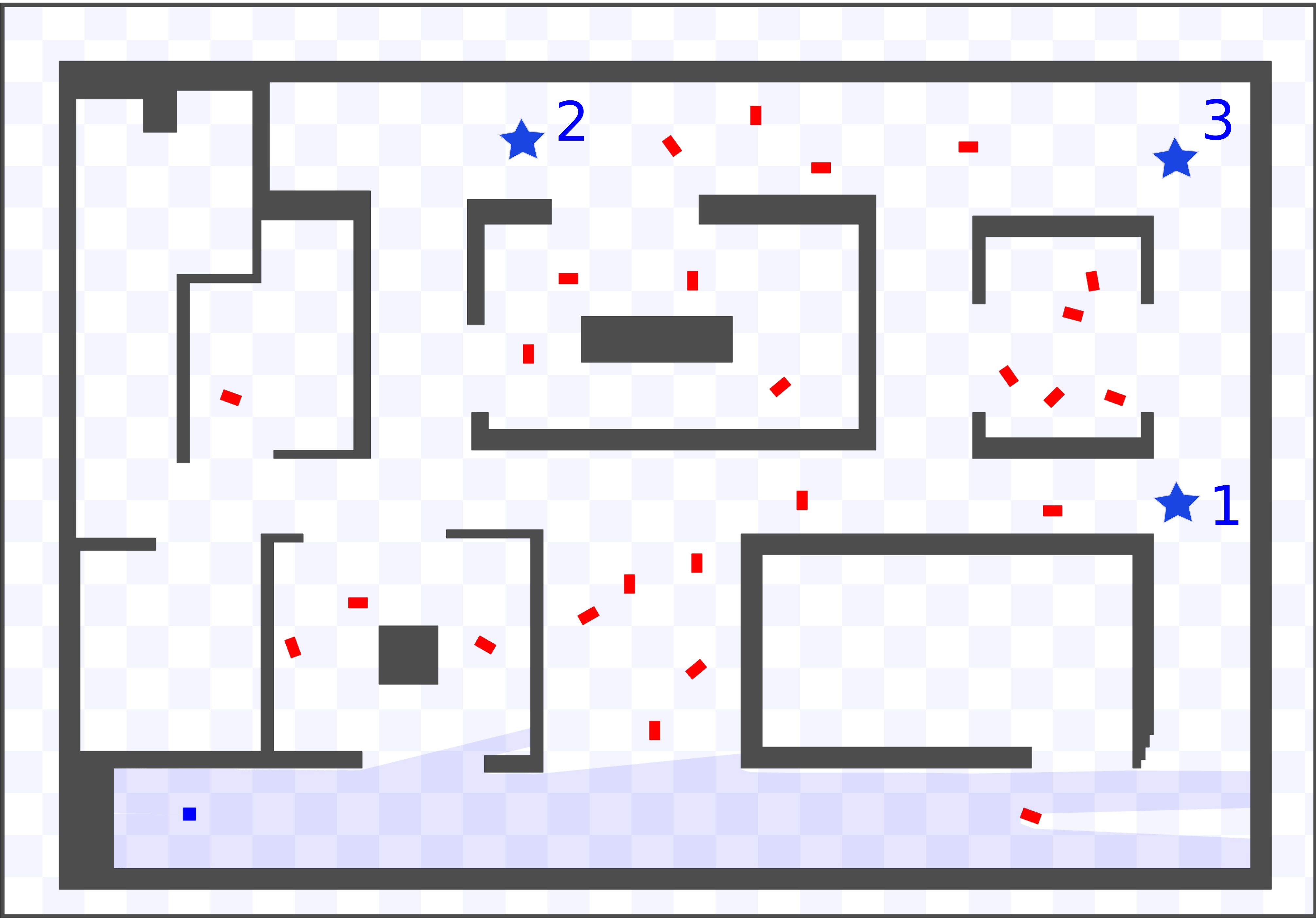}\label{fig:scen_disperse}}
    \end{center}
    \caption{
        Scenario used for simulations in two different situations. (a) shows the environment, where the blue rectangle represents the robot’s initial position, the blue stars indicate the goals to be reached, and the blue lines depict direct paths to each goal in the absence of dynamic obstacles. (b) and (c) show two configurations of the initial positions of dynamic obstacles. In (b), the obstacles are densely concentrated in specific areas, allowing paths through obstacle-free regions. In (c), the obstacles are dispersed throughout the scenario from the beginning, so the robot must find a path through less dense or lower-risk areas, if possible.
    }
    \label{fig:res_scenario}
\end{figure*}

The proposed method was implemented in C++ on the Robot Operating System (ROS) and tested using the Stage simulator. The video of an illustrative explanation of the proposed approach and the simulations is available in the link\footnotemark. 

\footnotetext{\url{https://bitly.cx/k1RS}}

\textbf{Scenario.}
The experiments were conducted in a typical indoor static environment (Fig.\ref{fig:scen_map}) that provides a diverse spatial structure, including both large and small rooms, as well as multiple alternative routes between the robot’s starting position and the goals. This topology offers meaningful route selection opportunities, allowing the planner to choose between different narrow and wide corridors or small and big room sequences depending on the evolving conditions of the environment. It reflects common public indoor spaces where human presence tends to concentrate, such as stores, university buildings, or offices.

On top of this static layout, two distinct configurations of dynamic obstacles were introduced to elicit different planning behaviors. In the dense configuration (Fig.\ref{fig:scen_dense}), obstacles are concentrated in specific regions, forming high-congestion zones that can be avoided under certain circumstances. In the dispersed configuration (Fig.\ref{fig:scen_disperse}), obstacles are spread throughout the environment in a non-uniform way, presenting routes with different levels of risk. These configurations allow for the evaluation of the planner’s ability to adapt its trajectory based on obstacle distribution and movement patterns, balancing efficiency and safety under varying dynamic conditions.

\begin{figure*}[tb!]
    \begin{center}
        \subfigure[t=5s]{\frame{\includegraphics[width=0.325\textwidth]{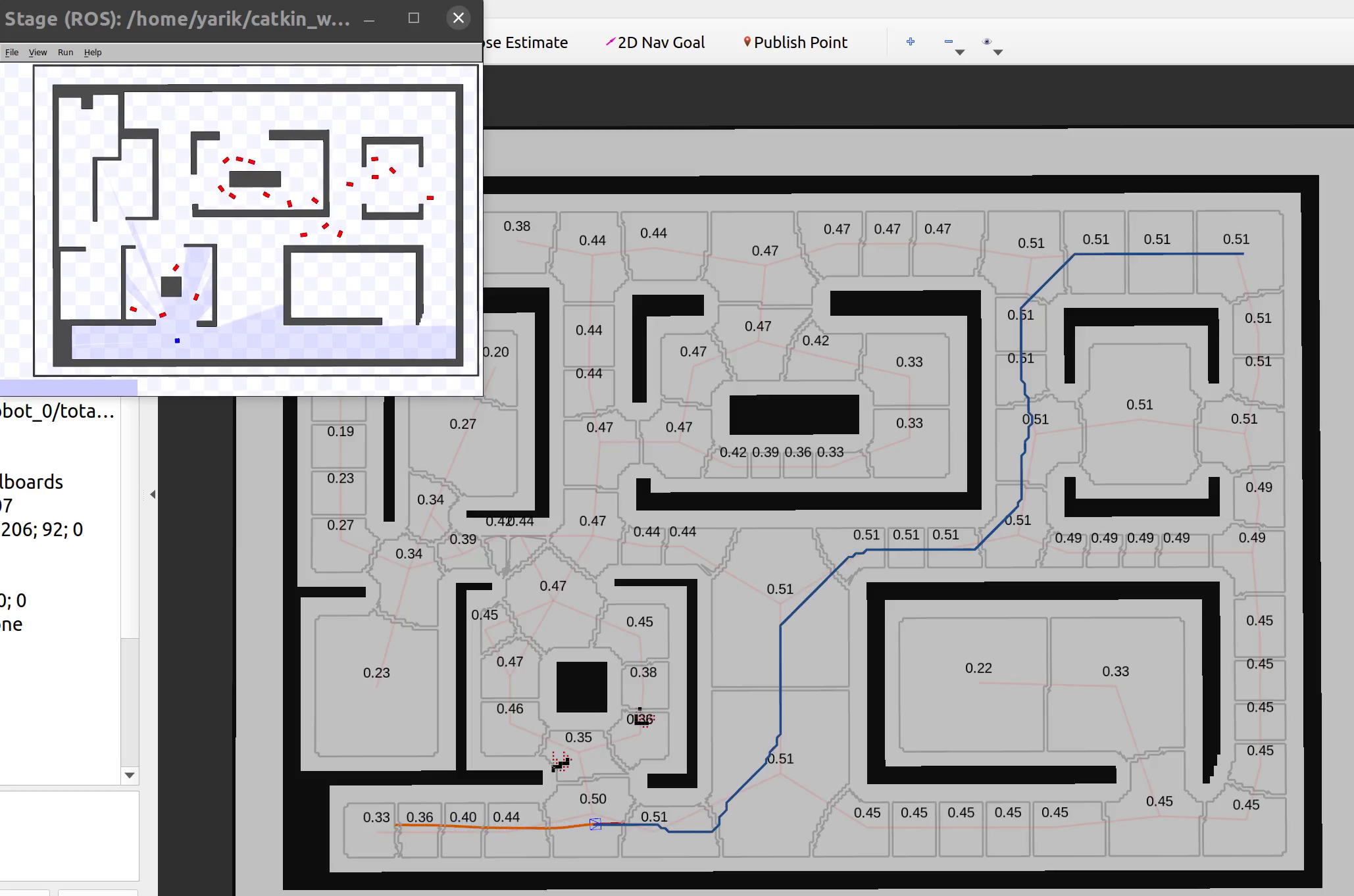}}} \hfill
        \subfigure[t=15s]{\frame{\includegraphics[width=0.325\textwidth]{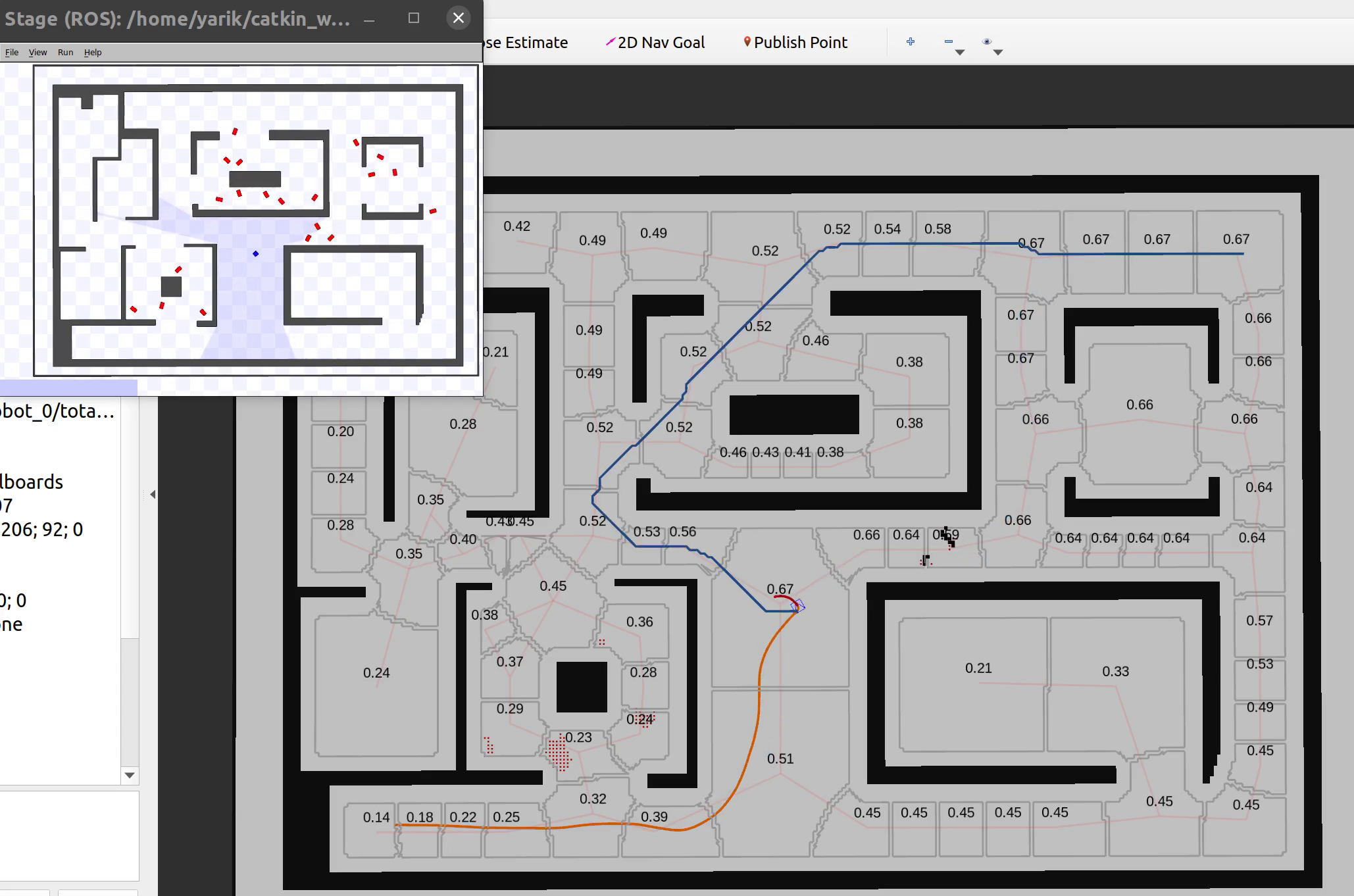}}} \hfill
        \subfigure[t=25s]{\frame{\includegraphics[width=0.325\textwidth]{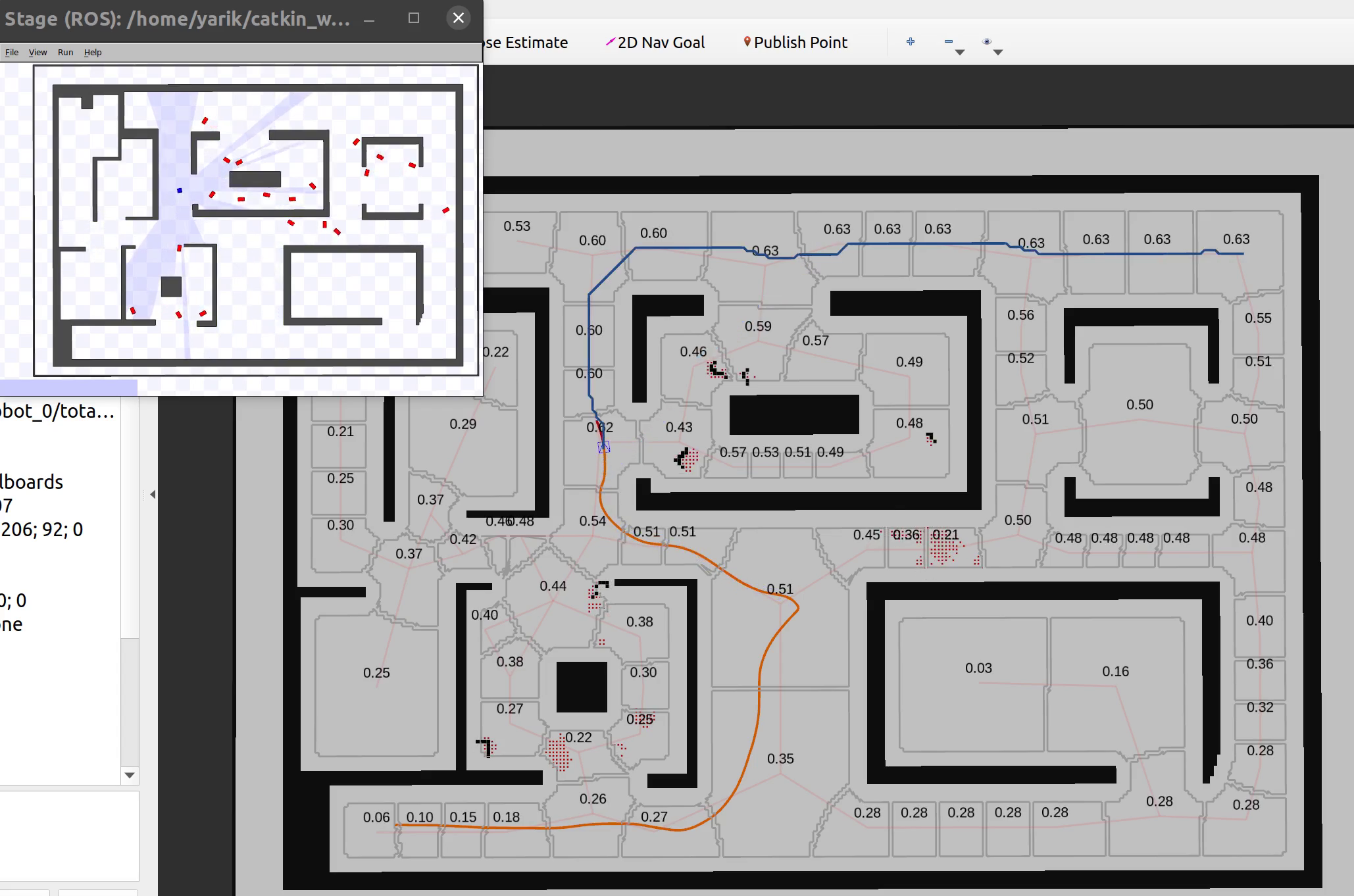}}} \hfill
    \end{center}
    \caption{
        Snapshots of a simulation example. The robot moves toward goal 3 in the densely concentrated scenario from Fig.\ref{fig:scen_dense}. (a) Initial planned path; dynamic obstacles detected by the robot are shown as black dots, and their trajectories as red dots. The planned and already traveled paths are represented by blue and orange lines, respectively. (b) After detecting new dynamic obstacles in the corridor, \emph{Tr-FMM} recomputes a larger path through obstacle-free regions, as the detour is not significant. (c) A new, safer path is computed after observing an additional set of dynamic obstacles.
    }
    \label{fig:res_execution}
\end{figure*}

Furthermore, the placement of both the goals and the dynamic obstacles was designed jointly to increase the probability of interactions between the robot and moving agents. This coordination ensures that reaching the goals often involves traversing or navigating around congested areas, thereby requiring the planner to re-evaluate its trajectory and select safer alternatives when appropriate. As such, the setup supports a thorough evaluation of the planner’s ability to balance efficiency with risk avoidance in dynamic environments. This behavior is illustrated in Fig.\ref{fig:res_execution}, which shows a sequence of snapshots where the robot adjusts its planned trajectory in response to emerging dynamic obstacles along the path to the goal.

\textbf{Dynamic obstacles motion.}
The dynamic obstacles move independently, mimicking the uncoordinated behavior of pedestrians in indoor public spaces. Their movement patterns consist of alternating phases of being stationary, walking forward, stopping again, and occasionally turning before continuing in a new direction. This model reflects typical human motion observed in environments such as stores or university buildings, where people move intermittently rather than continuously. Importantly, obstacles do not collide with each other, preserving realism in the simulation.

The movement of each obstacle is governed by a purely random process, with a fixed seed used to ensure reproducibility of the experiments. This randomness introduces significant variability and unpredictability in obstacle trajectories, posing a dynamic challenge to the planner. Consequently, the planner must continuously re-evaluate and adapt its trajectory in response to the evolving positions of the obstacles, balancing the trade-off between path efficiency and collision avoidance. We have launched 25 different random movements of dynamic obstacles in each trial evaluating each configuration: obstacle distribution - goal - evaluated planner - type of perception employed.

\textbf{Metrics.}
To evaluate the planner’s performance in dynamic environments, multiple quantitative metrics were considered. These include the total path length and the time taken to reach the goal, which reflect the efficiency and responsiveness of the planner, respectively. Additionally, the time the robot remains stationary was measured, capturing delays caused by re-planning or hesitation due to initially suboptimal path choices. Safety was assessed through distance metrics to dynamic obstacles, specifically the minimum distance during the trajectory, indicating critical proximity, and the average distance, representing the overall safety margin. The success rate was defined based on four outcome categories summarizing collision-related performance: success without collision, success with non-critical collision, failure due to critical collision, and failure without collision, although the latter was not observed in any of the experiments. Together, these metrics provide a comprehensive understanding of the planner’s ability to balance efficiency, safety, and robustness in dynamic, obstacle-rich environments.

The path is computed with \emph{Tr-FMM} with a frequency of 100Hz.
Once the path is computed, the Dynamic Window Approach (DWA) generates velocity commands for the robot. Dynamic obstacles move at linear speeds of 0.2 m/s and angular speeds of 0.5 rad/s, while the robot’s maximum linear and angular speeds are 0.5 m/s and 1.0 rad/s, respectively.

To ensure a fair and isolated evaluation of the proposed \emph{Tr-FMM} path planner, DWA—a classical and widely adopted local navigation method—was deliberately selected as the low-level controller. The use of more sophisticated navigation strategies that incorporate dynamic obstacle avoidance could mask the origin of failures, making it unclear whether collisions result from the path planner or the navigator. Since this work focuses exclusively on the design and performance of the path planner, using a standard, reactive controller allows clearer attribution of results. This also justifies the use of success rate, including collision occurrences, as a meaningful metric to directly assess the quality of the computed paths.

Since no ground truth exists due to the dynamic nature of the environment, we compare with a baseline solution (BS), obtained using the basic FMM without dynamic obstacles (dashed lines in Fig.\ref{fig:scen_map}). This serves as a reference for how far the \emph{Tr-FMM} solution is from the optimal one. Experiments were conducted on a machine with an Intel Core i7-13700 processor (2.1 GHz) and 16 GB of RAM. The average time for \emph{Tr-FMM} to compute the path across all experiments was 30-40 ms.

\subsubsection{Dispersion of obstacles and goals}

\begin{figure*}[tb!]
    \begin{center}
        
        \subfigure[Traveled distance (m)]{\includegraphics[width=.45\textwidth]{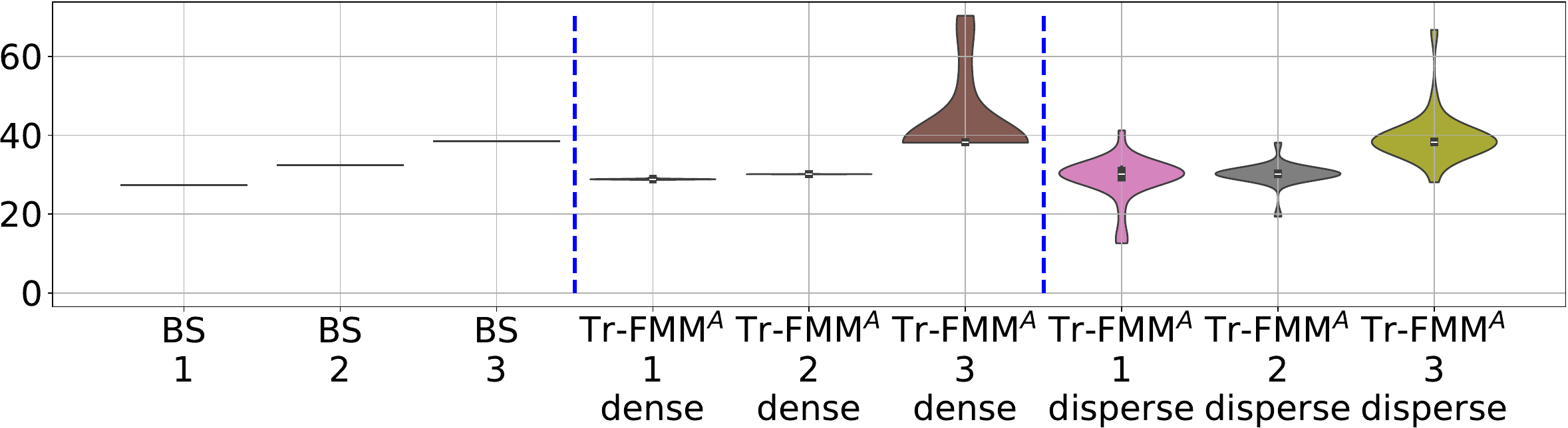}\label{fig:res_goals_dist}} \hspace{5mm}
        \subfigure[Times (s)]{\includegraphics[width=.45\textwidth]{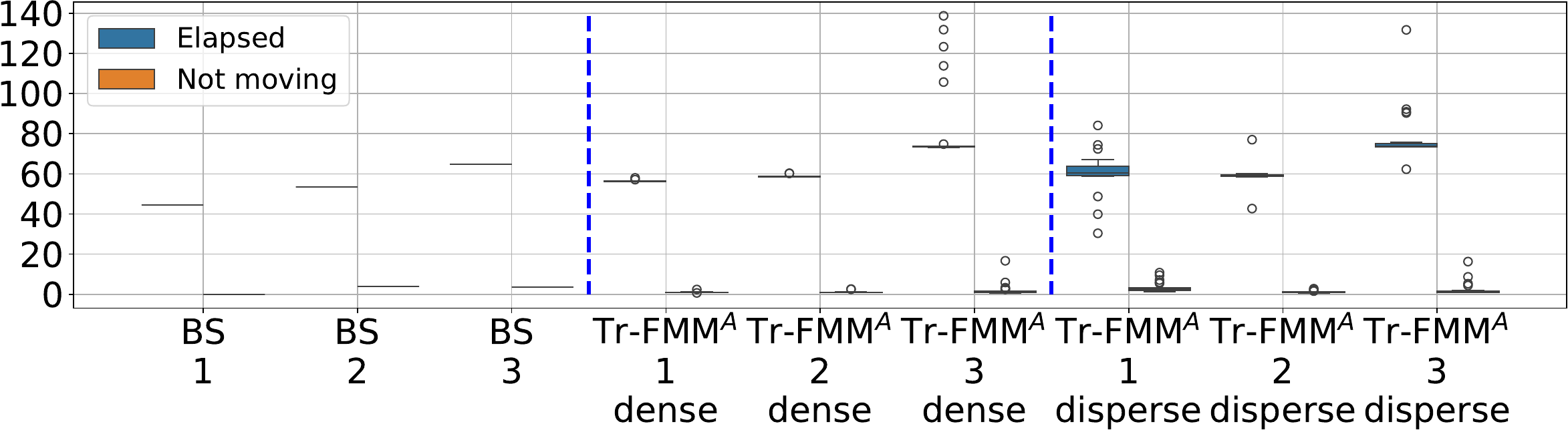}\label{fig:res_goals_time}} \\
        \subfigure[Distance to obstacles (m)]{\includegraphics[width=.45\textwidth]{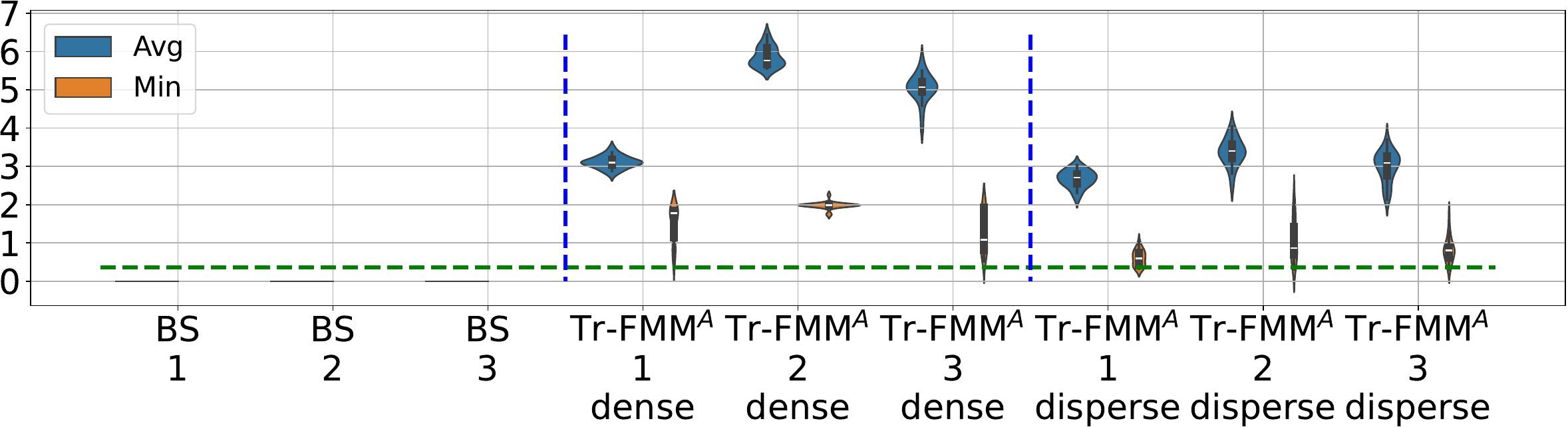}\label{fig:res_goals_d2o}} \hspace{5mm}
        \subfigure[Success rates]{\includegraphics[width=.45\textwidth]{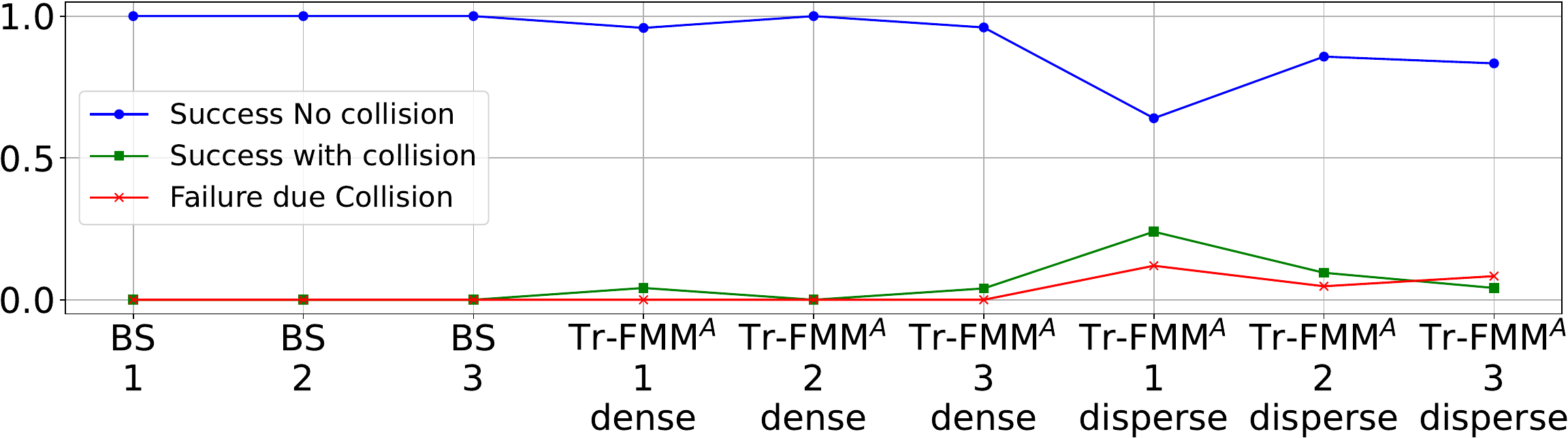}\label{fig:res_goals_succ}}
    \end{center}
    \caption{
        Results based on the dispersion of the obstacles and goals. The numbers indicate the index of the goal and the green line in (c) illustrates the radius of the robot.
    }
    \label{fig:res_goals}
\end{figure*}

We begin our evaluation of \emph{Tr-FMM} by analyzing its performance under different configurations of dynamic obstacle dispersion and goal placements. To minimize the inherent uncertainty characteristic of dynamic environments, this initial experiment assumes that the robot has complete and continuous knowledge of the positions of all dynamic obstacles throughout the mission. At every timestep, the planner recomputes the optimal path based on the updated obstacle positions, effectively simulating an overhead view of the scenario. This type of perception serves as a basis for further comparison with the more common case in robotics, in which the robot perceives the environment only through its onboard sensors.

In the dense scenario, where dynamic obstacles are initially clustered in specific regions as shown in Fig.\ref{fig:scen_dense}, obstacle-free corridors are often available at the beginning of the mission. Under these conditions, \emph{Tr-FMM} takes advantage of these clear regions to compute safe and efficient paths. In contrast, the dispersed scenario, illustrated in Fig.\ref{fig:scen_disperse}, presents a more challenging situation: dynamic obstacles are more uniformly spread across the environment, significantly reducing the availability of obstacle-free regions. Here, the planner must evaluate the relative risk of each region and prioritize less risky paths to minimize potential collisions.

For the first two goals, which are relatively close to the robot's starting position, the dynamic obstacles have limited time to spread. This results in the planner favoring alternative paths through adjacent or parallel corridors, avoiding shared regions with moving obstacles and increasing the safety margin in case of dense distribution of obstacles—this is reflected in the greater average distance maintained between the robot and nearby obstacles, as shown in Fig.\ref{fig:res_goals_d2o}. While this behavior slightly increases the total path length and mission time (see Fig.\ref{fig:res_goals_dist} and Fig.\ref{fig:res_goals_time}), it demonstrates the planner’s ability to proactively avoid risky areas.

In the case of Goal 3, which is located farther from the starting point, dynamic obstacles have more time to disperse into the environment. Consequently, the robot is more frequently forced to traverse partially occupied regions, weaving between obstacles while avoiding denser areas. This results in longer trajectories and extended mission durations. Despite the added complexity, the robot manages to reach all goals in this experiment, with a success rate of 100\%, as indicated in Fig.\ref{fig:res_goals_succ}.

In the dispersed scenario, where there are no entirely obstacle-free paths from the start, the robot is often in closer proximity to obstacles, as reflected in the consistently smaller distances to the nearest obstacle in Fig.\ref{fig:res_goals_d2o}. Here, \emph{Tr-FMM} must continuously strike a balance between following a direct route to the goal and avoiding high-risk regions. Although the average distances and times to reach the goal are similar to those observed in the dense case, the absence of safe corridors increases the possibility of encountering difficult configurations. In some instances, the robot successfully finds longer but safer paths, while in others, it fails due to some collisions—these failures are reflected in the success rate drop in Fig.\ref{fig:res_goals_succ}.

Even under the assumption of complete and continuous knowledge of the positions of all dynamic obstacles, collisions are still observed in some simulations. This outcome is explained by the combination of two factors: on the one hand, the use of DWA as the local planner, which does not always guarantee avoidance of dynamic obstacles; on the other hand, scenarios in which dynamic obstacles are highly dispersed throughout the environment, leaving no feasible path to the goal that avoids interaction with them. As a result, the robot must navigate through regions with moving obstacles, where the limited reactivity of DWA may lead to collisions, despite the accuracy of the global path planning with \emph{Tr-FMM}.

\begin{figure*}[tb!]
    \begin{center}
        \subfigure[All positions known t=5s]{\frame{\includegraphics[width=0.325\textwidth]{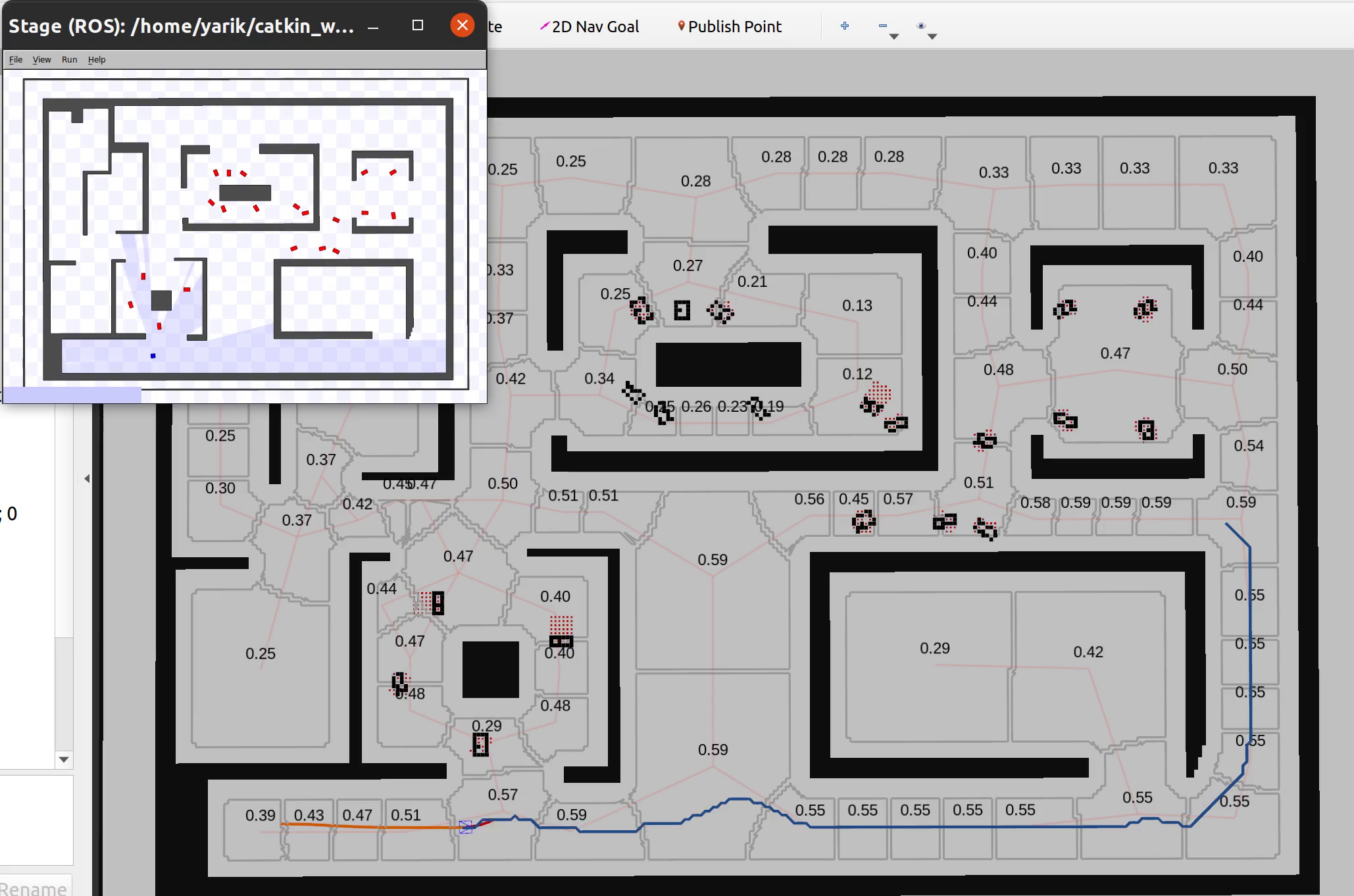}\label{fig:res_execution_a1}}} \hfill
        \subfigure[All positions known t=15s]{\frame{\includegraphics[width=0.325\textwidth]{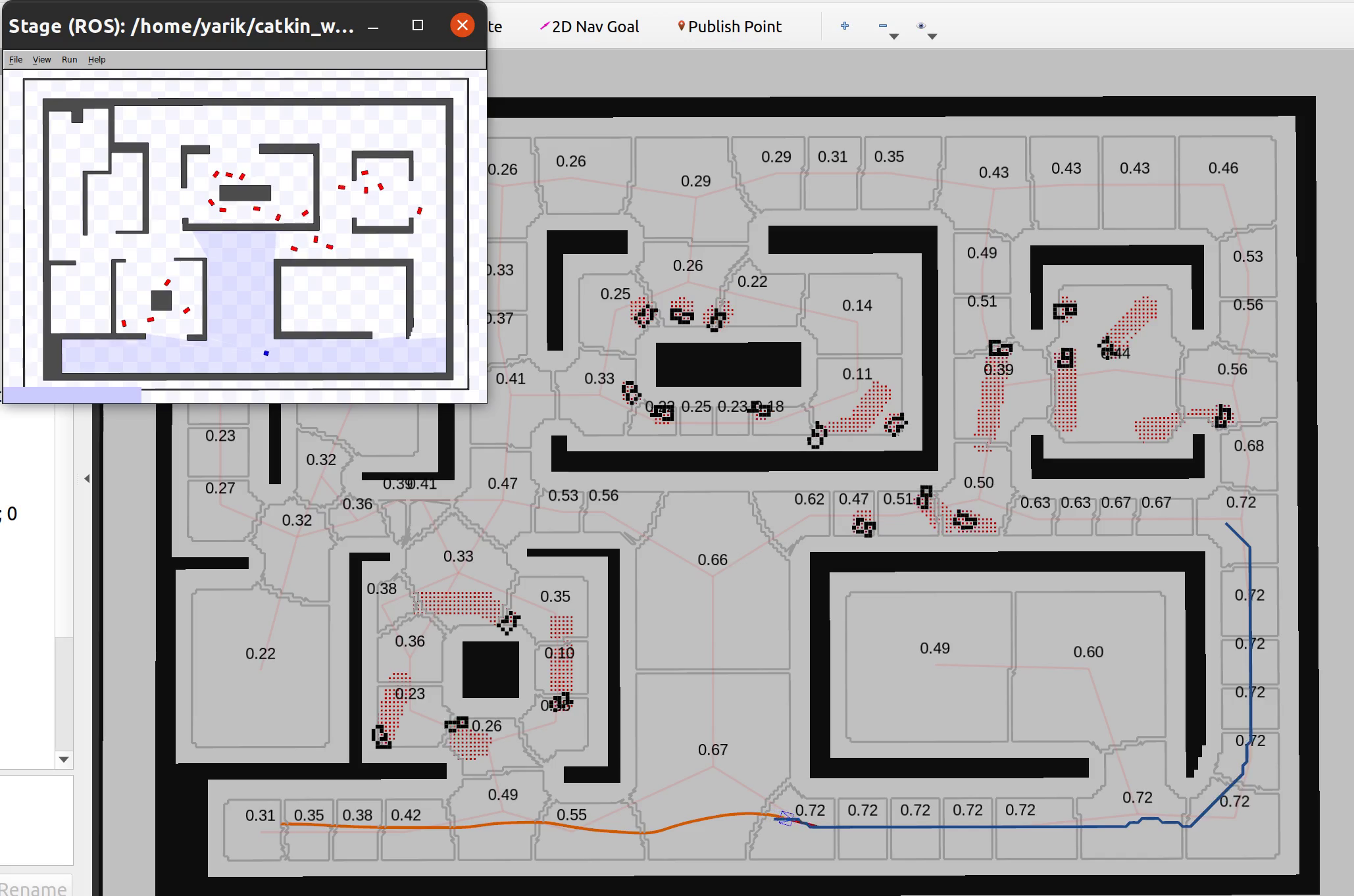}\label{fig:res_execution_a2}}} \hfill
        \subfigure[All positions known t=37s]{\frame{\includegraphics[width=0.325\textwidth]{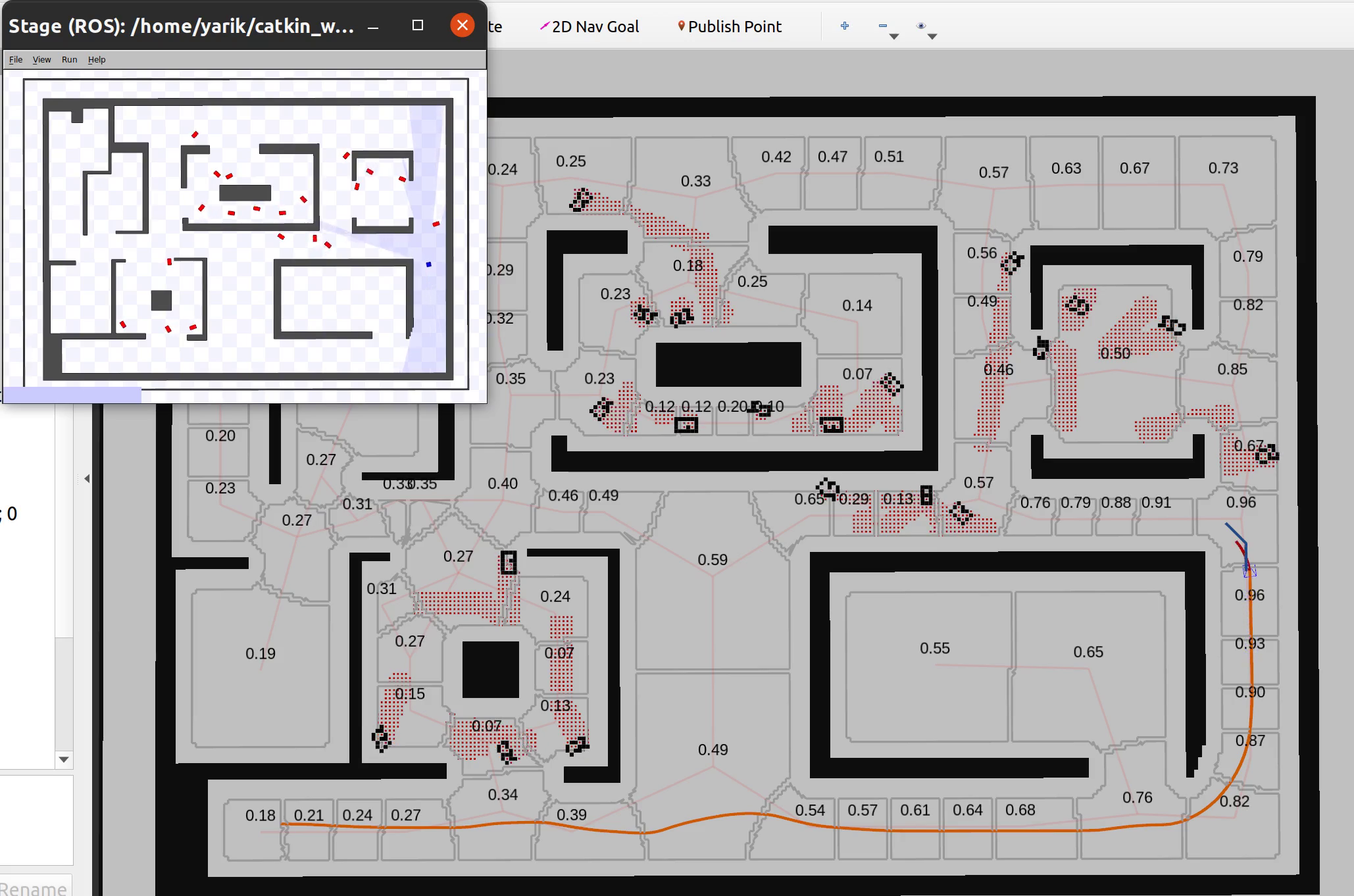}\label{fig:res_execution_a3}}} \\
        \subfigure[Line-of-sight t=5s]{\frame{\includegraphics[width=0.325\textwidth]{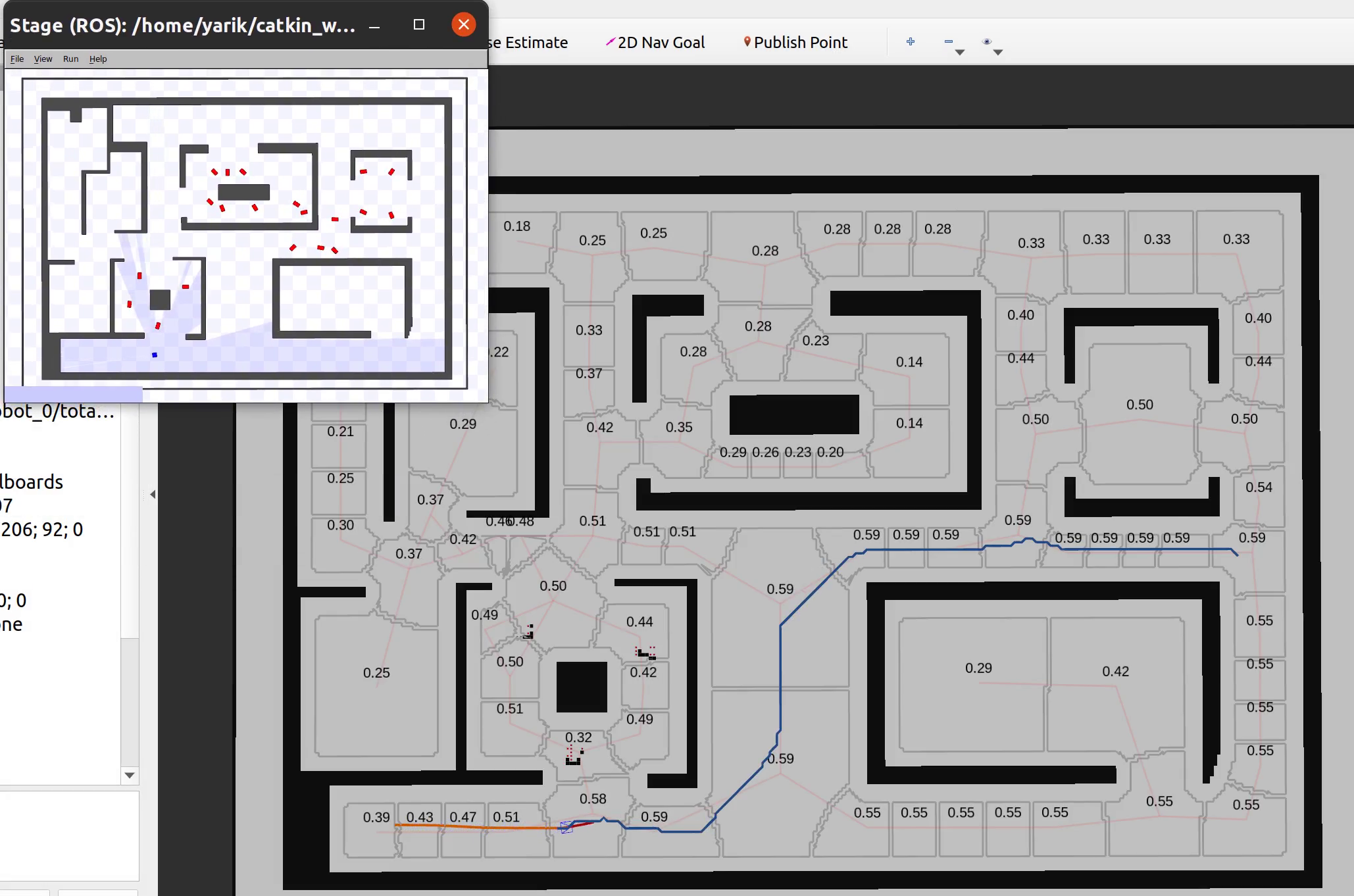}\label{fig:res_execution_l1}}} \hfill
        \subfigure[Line-of-sight t=24s]{\frame{\includegraphics[width=0.325\textwidth]{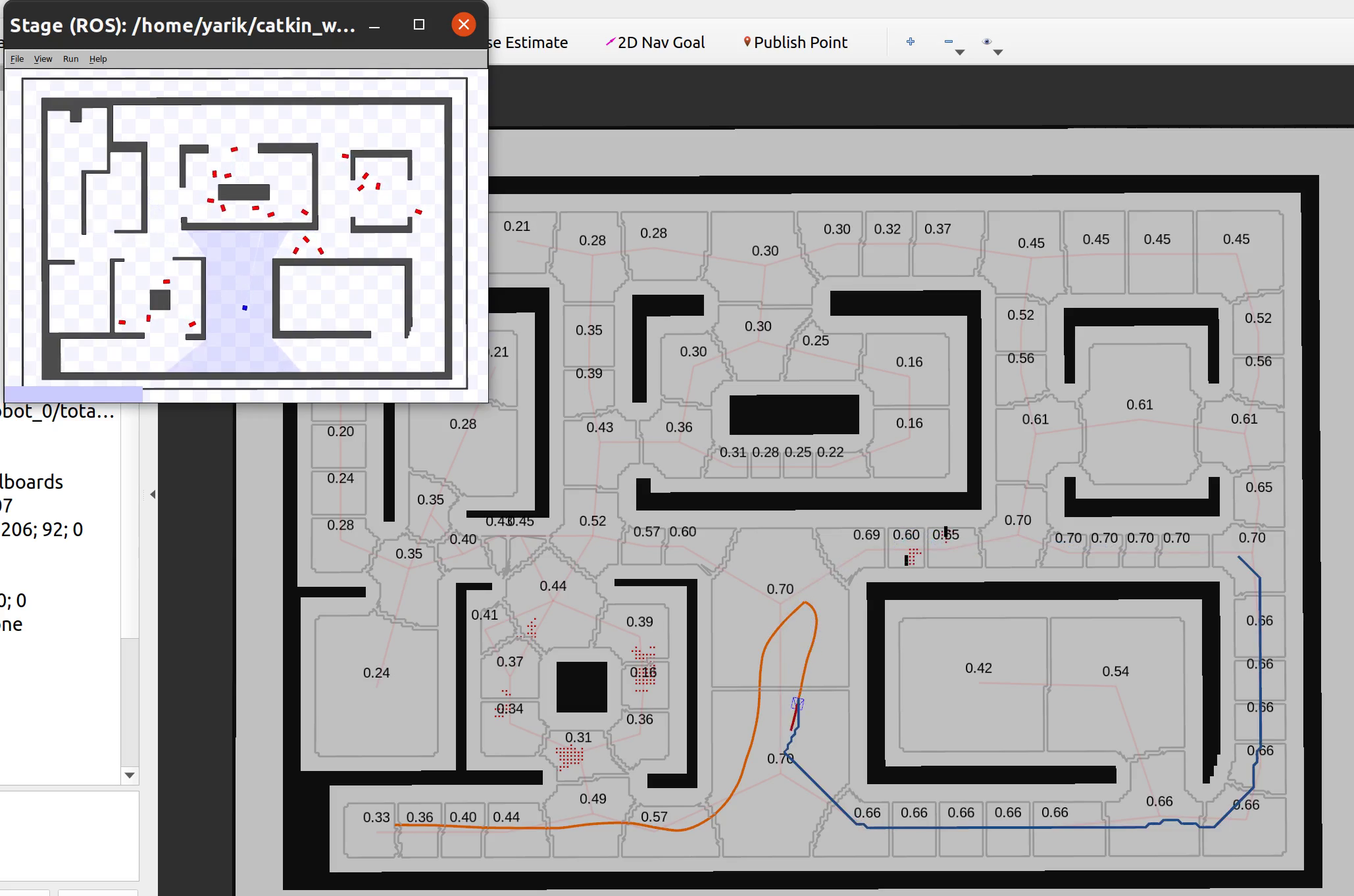}\label{fig:res_execution_l2}}} \hfill
        \subfigure[Line-of-sight t=50s]{\frame{\includegraphics[width=0.325\textwidth]{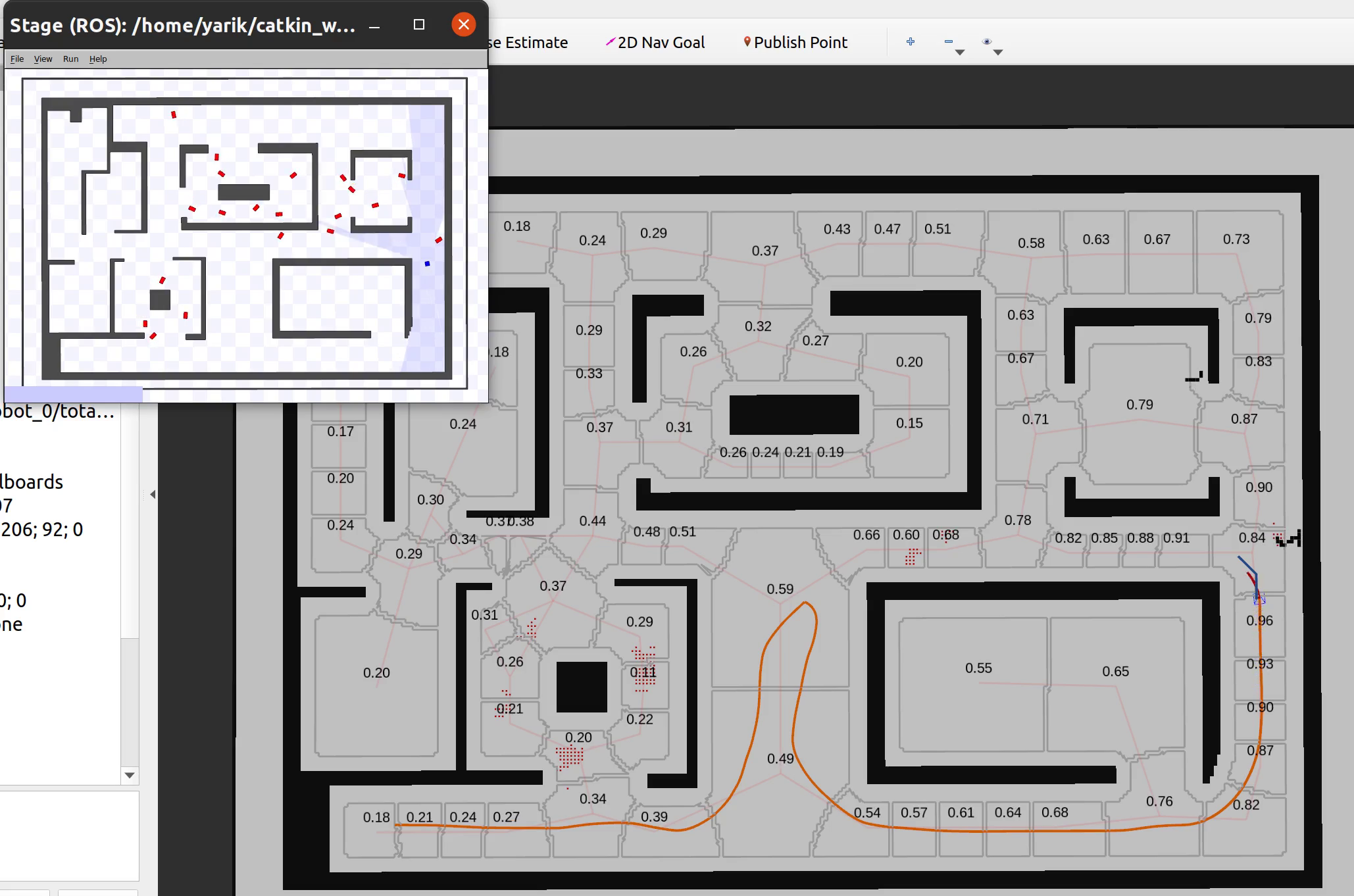}\label{fig:res_execution_l3}}}
    \end{center}
    \caption{
        Difference when the robot only knows all the positions of all the dynamic obstacles from external sensors (a)-(c) and when the robot registers the positions of obstacles from onboard sensors when they are in Line-of-Sight (d)-(f). When all positions of dynamic obstacles are known, \emph{Tr-FMM} directly obtains the path through safer regions (a), without needing to reactively changing the path (b) and reaching the goal earlier (c). With line-of-sight, the robot does not have information of the obstacles, so it plans direct path to goal (d). When it discovers the dynamic obstacles (e), it re-plans the path trough larger but safer regions (f).
    }
    \label{fig:res_execution_la}
\end{figure*}

Interestingly, failed trials often show shorter distances and times than successful ones. This is explained by the premature termination of the mission upon collision, which prevents the robot from completing the full trajectory. It is also worth mentioning that, in both scenarios, the robot occasionally faces local deadlocks—situations where progress is temporarily blocked. However, these situations are typically resolved quickly, with the time spent handling deadlocks averaging only 5\% of the total mission time, as illustrated in Fig.\ref{fig:res_goals_time}. This highlights the robustness and responsiveness of the planning strategy even under uncertainty and high-density environments.

\subsubsection{Partial observation vs. All positions of obstacles are known}

This experiment compares two modes of acquiring environmental information regarding dynamic obstacles. 
In the first mode, the one described in the previous section, the robot has full knowledge of the positions of dynamic obstacles at all times, for example, using an external camera system—feasible in indoor environments such as stores, offices, or warehouses. In the second mode, the more typical setup, the robot perceives the motion of dynamic obstacles through its onboard sensors within Line-of-Sight (LoS), thus having limited awareness of the environment.
The difference between these two modes is illustrated in the snapshots of Fig.\ref{fig:res_execution_la}. By acquiring the positions of obstacles from external sensors, \emph{Tr-FMM} directly computes the path to goal avoiding the crowded regions, Fig.\ref{fig:res_execution_a1}-\ref{fig:res_execution_a3}. It is important to note that even with complete information on obstacle positions, uncertainty in their future movements remains. By contrast, when the robot takes the readings from onboard sensor needs to replan the paths when detecting new dynamic obstacles, Fig.\ref{fig:res_execution_l1}-\ref{fig:res_execution_l3}, needing an extra time to reach the goal.

\begin{figure*}[tb!]
    \begin{center}

        \subfigure[Traveled distance (m)]{\includegraphics[width=.45\textwidth]{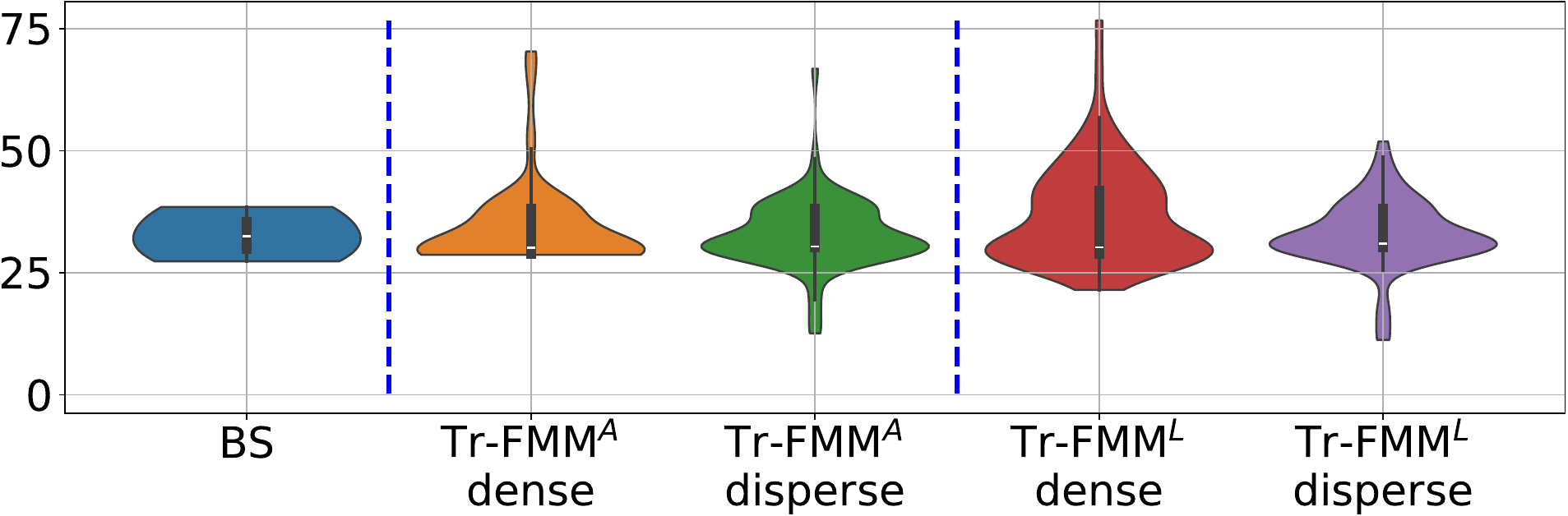}\label{fig:res_obstacles_dist}} \hspace{5mm}
        \subfigure[Times (s)]{\includegraphics[width=.45\textwidth]{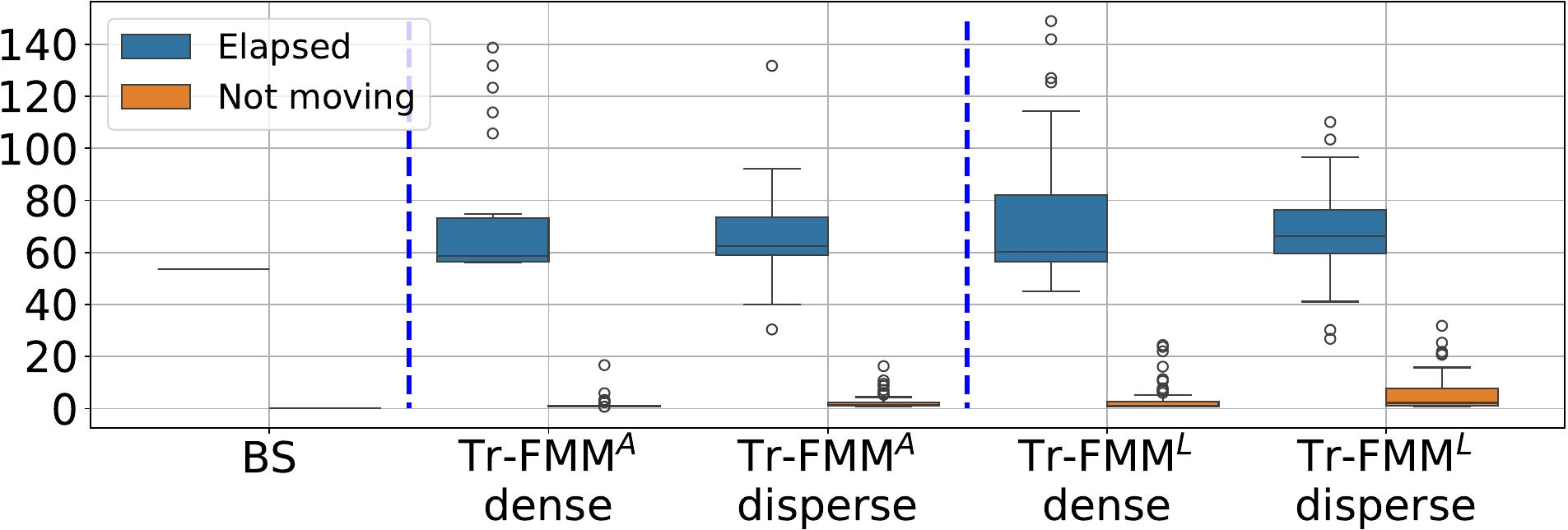}\label{fig:res_obstacles_time}} \\
        \subfigure[Distance to obstacles (m)]{\includegraphics[width=.45\textwidth]{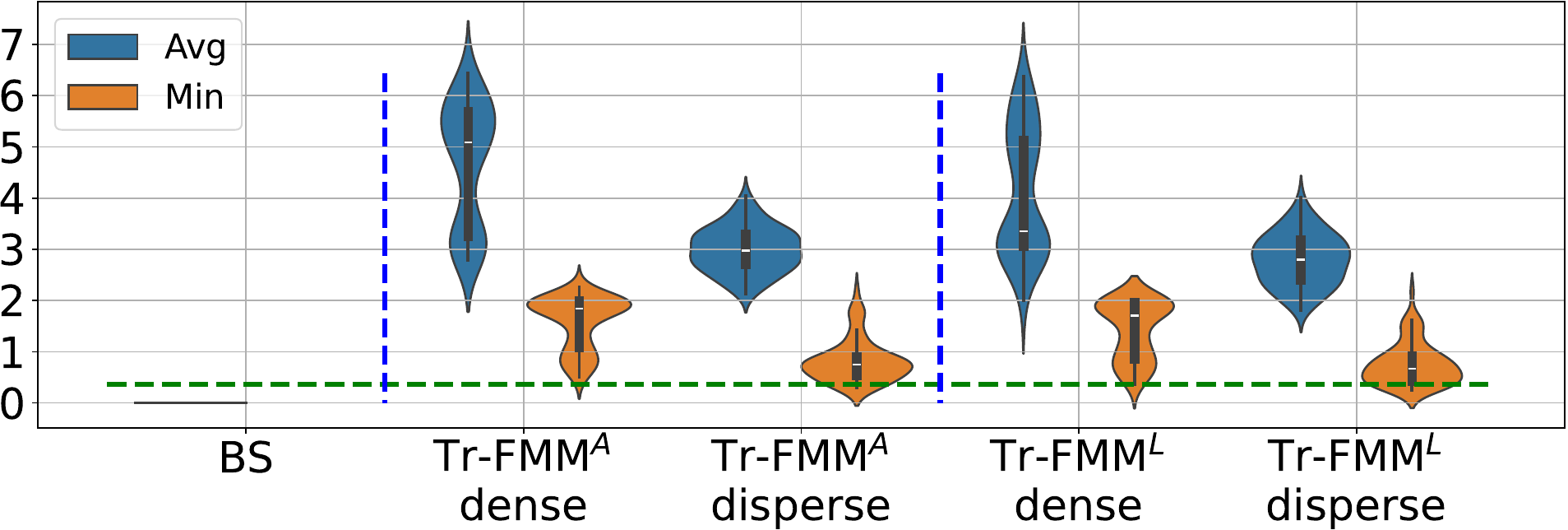}\label{fig:res_obstacles_d2o}} \hspace{5mm}
        \subfigure[Success rates]{\includegraphics[width=.45\textwidth]{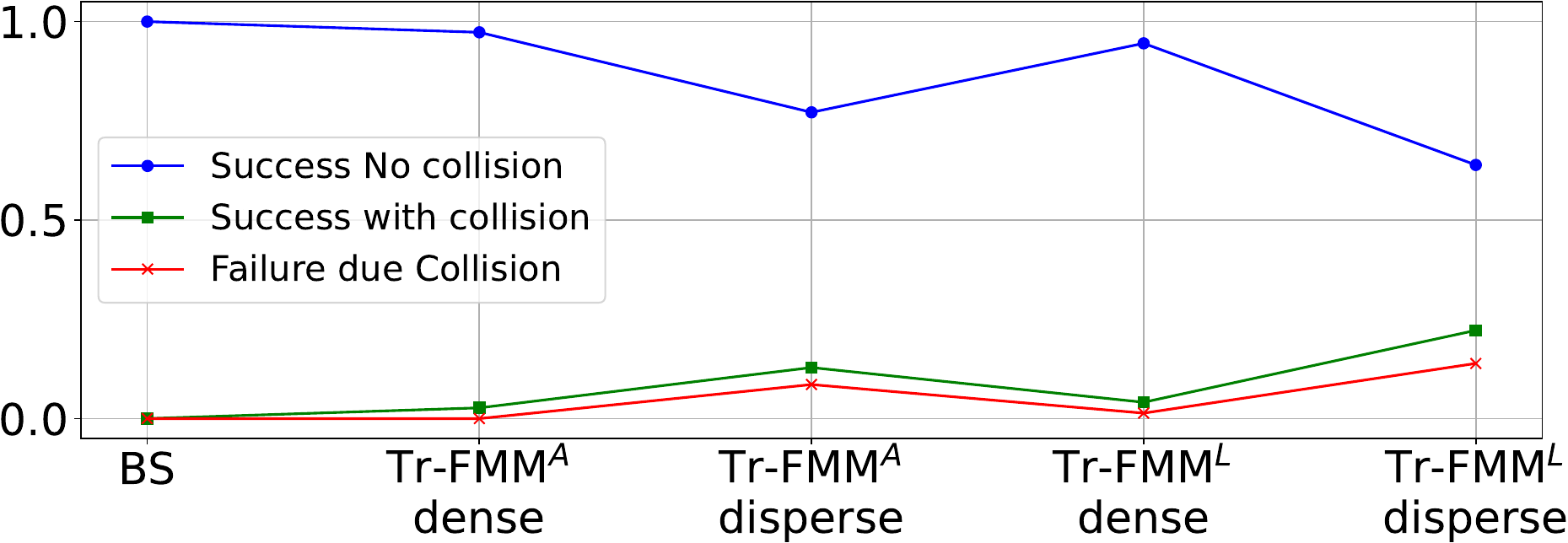}\label{fig:res_obstacles_succ}}
    \end{center}
    \caption{
        Results based on the knowledge of the obstacles positions
    }
    \label{fig:res_obstacles}
\end{figure*}

The results are categorized as $\emph{Tr-FMM}^A$ when the robot has full and continuous knowledge of obstacle all positions, and as $\emph{Tr-FMM}^L$ when only limited LoS awareness is available. Fig.\ref{fig:res_obstacles} consolidates the performance metrics for all three goal positions, showing the correspondence of $\emph{Tr-FMM}^A$ results with those presented earlier in Fig.\ref{fig:res_goals}.

Focusing first on the disperse scenario, the outcomes between both knowledge modes are largely comparable. Given the sparse distribution of obstacles, the environment offers very few truly obstacle-free paths, compelling the robot to operate in close proximity to dynamic obstacles, as reflected in the consistently small distances to nearby obstacles shown in Fig.\ref{fig:res_obstacles_d2o}. This proximity necessitates frequent adjustments to the robot’s trajectory, resulting in an increase in path replanning events and idle times, evident in Fig.\ref{fig:res_obstacles_time}. Consequently, the collision rate is higher, with 15\% of missions ending in failure due to collisions, and an additional 20\% of runs completing successfully but with collisions occurring during execution (Fig.\ref{fig:res_obstacles_succ}). These statistics highlight the challenges imposed by the limited sensing capabilities and the high density of obstacles in this scenario.

In contrast, in the dense scenario where obstacle clusters are more pronounced, the planner’s ability to identify safer, obstacle-free corridors results in longer, but more reliable paths. Distances traveled and mission durations increase as the robot deliberately detours to avoid risky regions, as shown by the upward trends in Fig.\ref{fig:res_obstacles_d2o}. This behavior stems from the consideration of obstacle dispersion within regions, as modeled by eq.\ref{eq:dispersion}, which effectively repels planned paths away from regions with high obstacle occupancy towards safer alternatives. The low collision rate observed in Fig.\ref{fig:res_obstacles_succ} confirms the effectiveness of this strategy. It is worth noting that some distances recorded for $\emph{Tr-FMM}^L$ are shorter than those for $\emph{Tr-FMM}^A$; this is primarily because about 5\% of $\emph{Tr-FMM}^L$ trials ended prematurely due to collisions, artificially lowering the average distance metric.

From these observations, an important intermediate conclusion emerges: even when operating with only line-of-sight information, $\emph{Tr-FMM}$ demonstrates robust performance by effectively balancing safety and efficiency. In scenarios characterized by limited obstacle dispersion, the planner adopts more conservative strategies, executing larger detours to maintain safe separations from dynamic obstacles under uncertain future movements. Conversely, when obstacles are widely dispersed and no fully obstacle-free paths exist, $\emph{Tr-FMM}$ flexibly prioritizes navigating through less dense, and consequently less risky, regions, optimizing path safety without excessively sacrificing efficiency.

\subsubsection{Comparison with other approaches}

\begin{figure*}[tb!]
    \begin{center}
        \subfigure[SFMM]{\includegraphics[width=.29\textwidth]{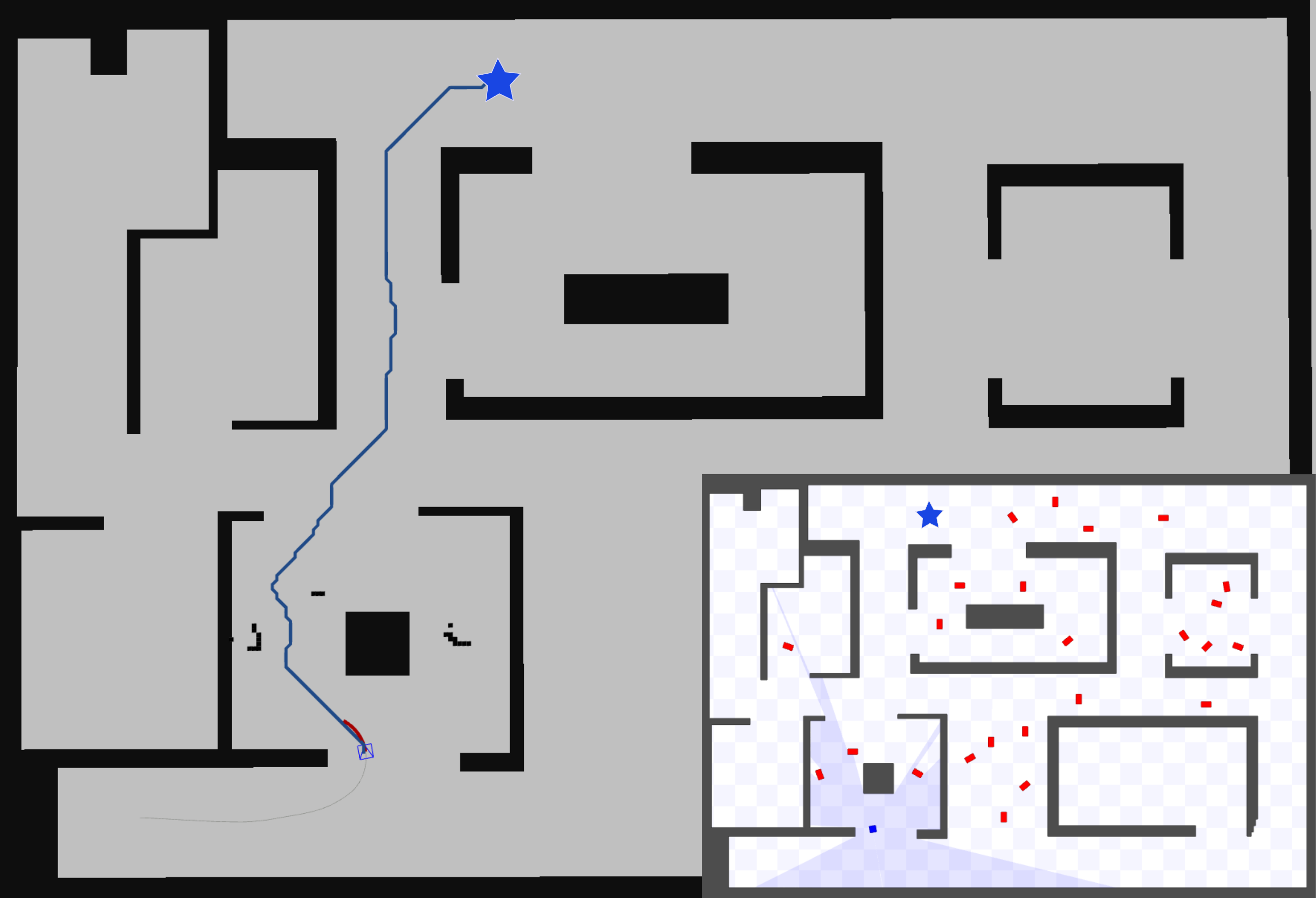}\label{fig:res_metodos_sfmm}} \hspace{5mm}
        \subfigure[CBDB]{\includegraphics[width=.29\textwidth]{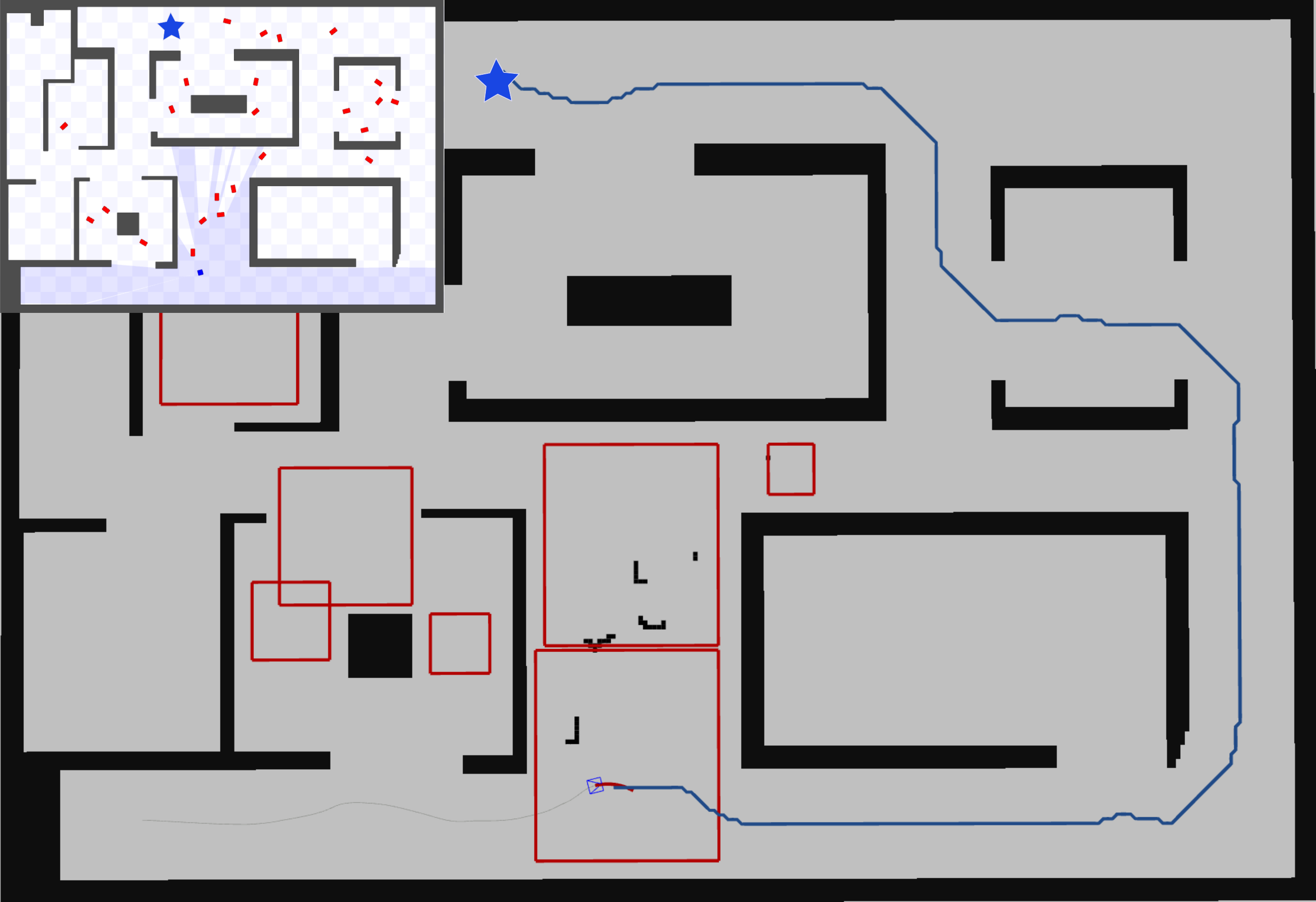}\label{fig:res_metodos_cbdb}} \hspace{5mm}
        \subfigure[\emph{Tr-FMM}]{\includegraphics[width=.29\textwidth]{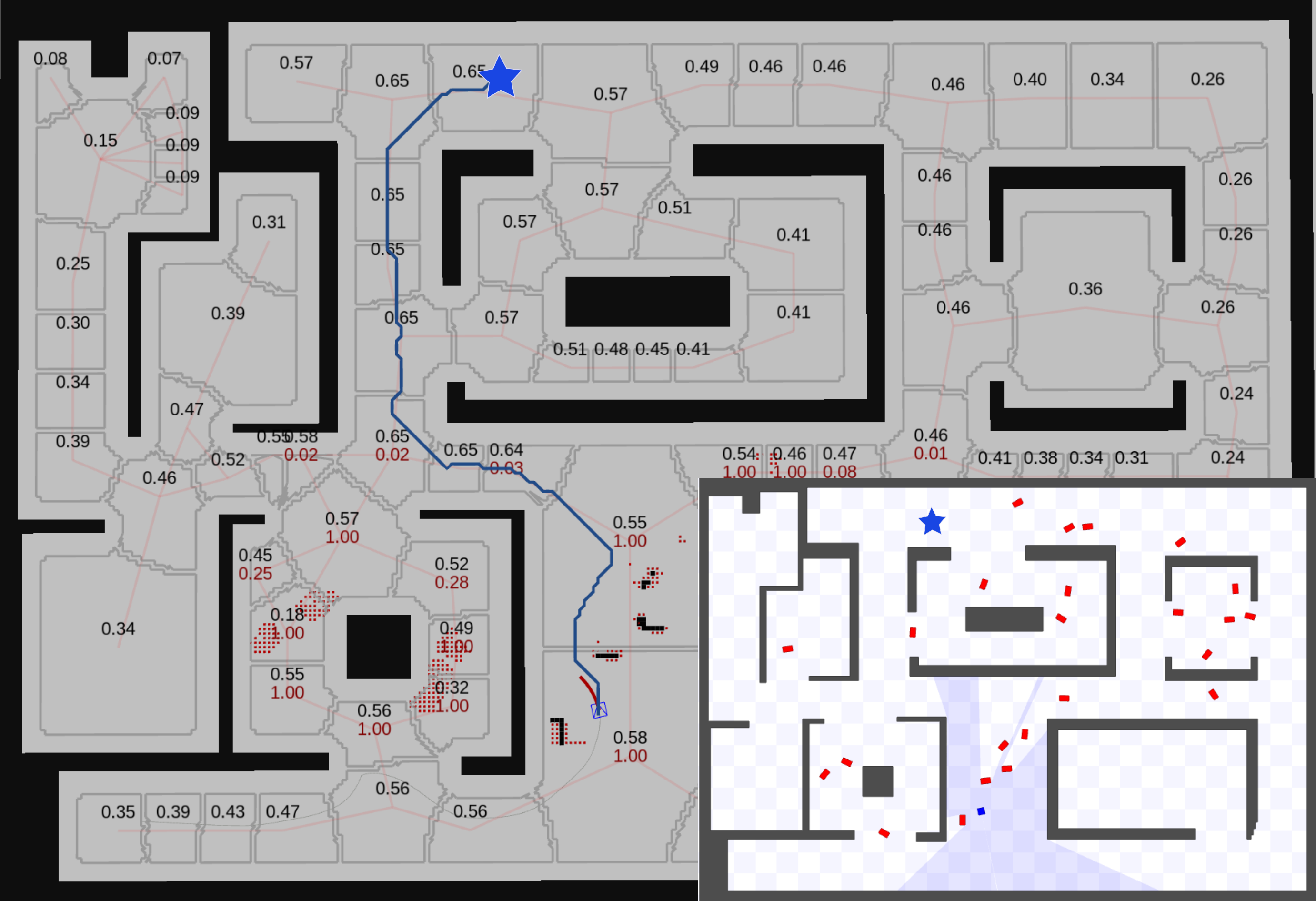}\label{fig:res_metodos_trfmm}}
    \end{center}
    \caption{
        Tested approaches. (a) With SFMM, the path is repelled from obstacles but still traverses crowded areas. (b) With CBDB, a longer path ensures obstacle-free traversal. (c) \emph{Tr-FMM} balances region occupation and deviation to compute the path.
    }
    \label{fig:res_metodos_ej}
\end{figure*}

\begin{figure*}[tb!]
    \begin{center}

        \subfigure[Traveled distance (m)]{\includegraphics[width=.45\textwidth]{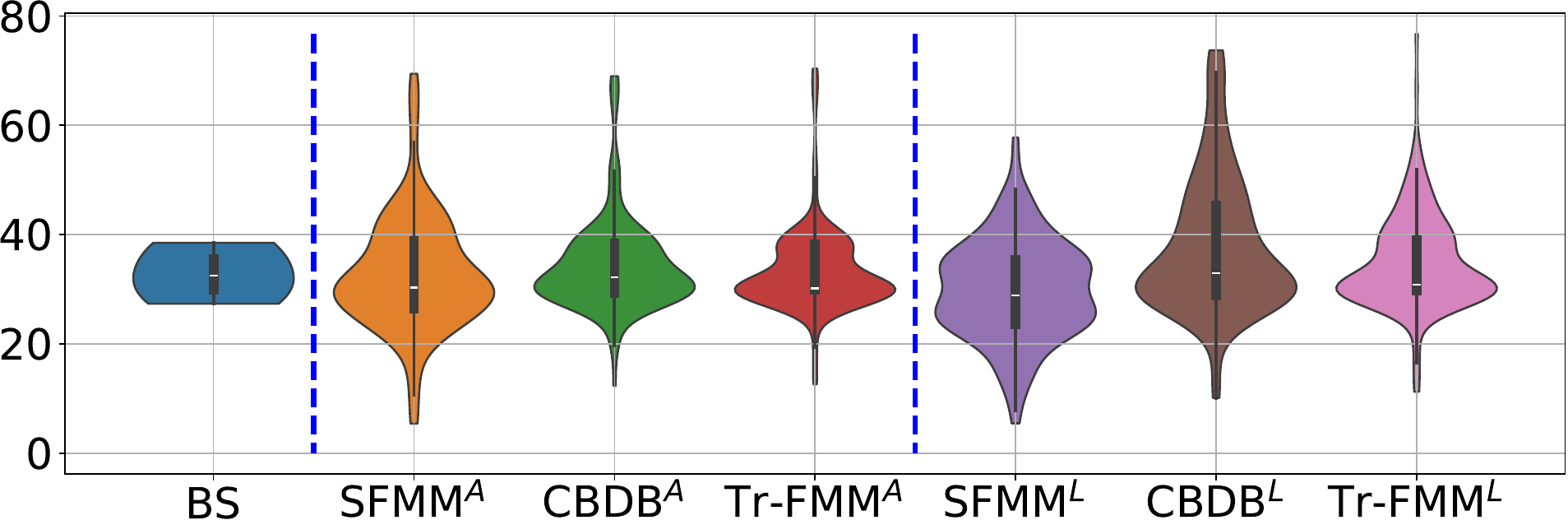}\label{fig:res_metodos_dist}} \hspace{5mm}
        \subfigure[Times (s)]{\includegraphics[width=.45\textwidth]{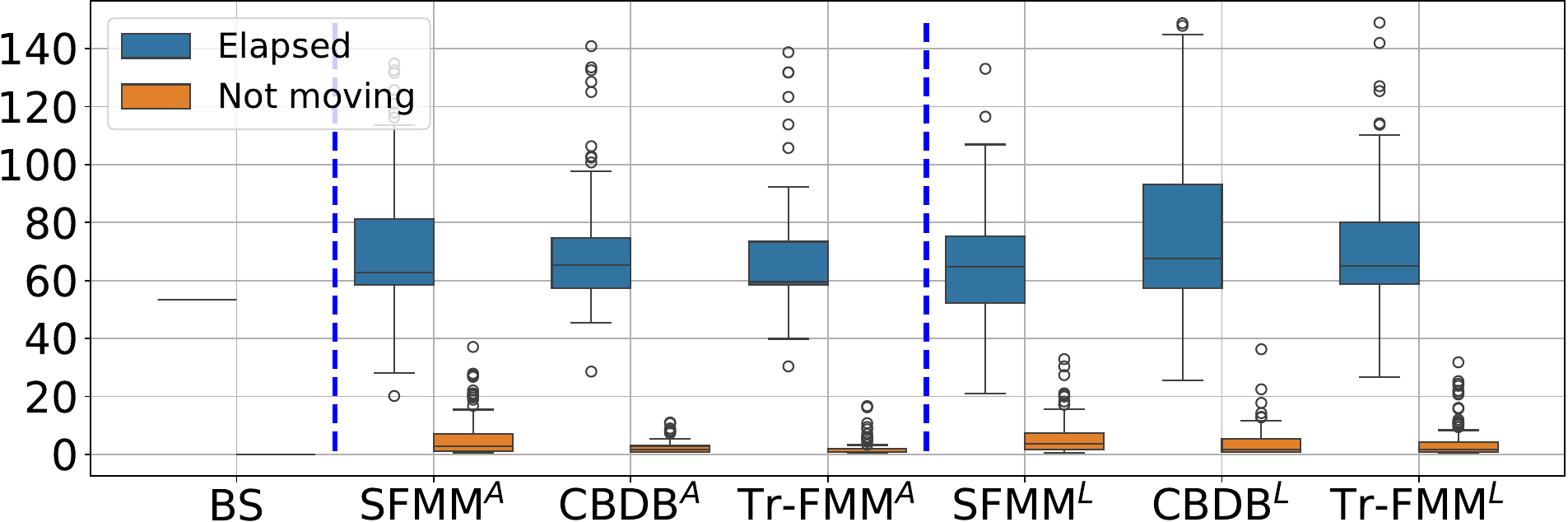}\label{fig:res_metodos_time}} \\
        \subfigure[Distance to obstacles (m)]{\includegraphics[width=.45\textwidth]{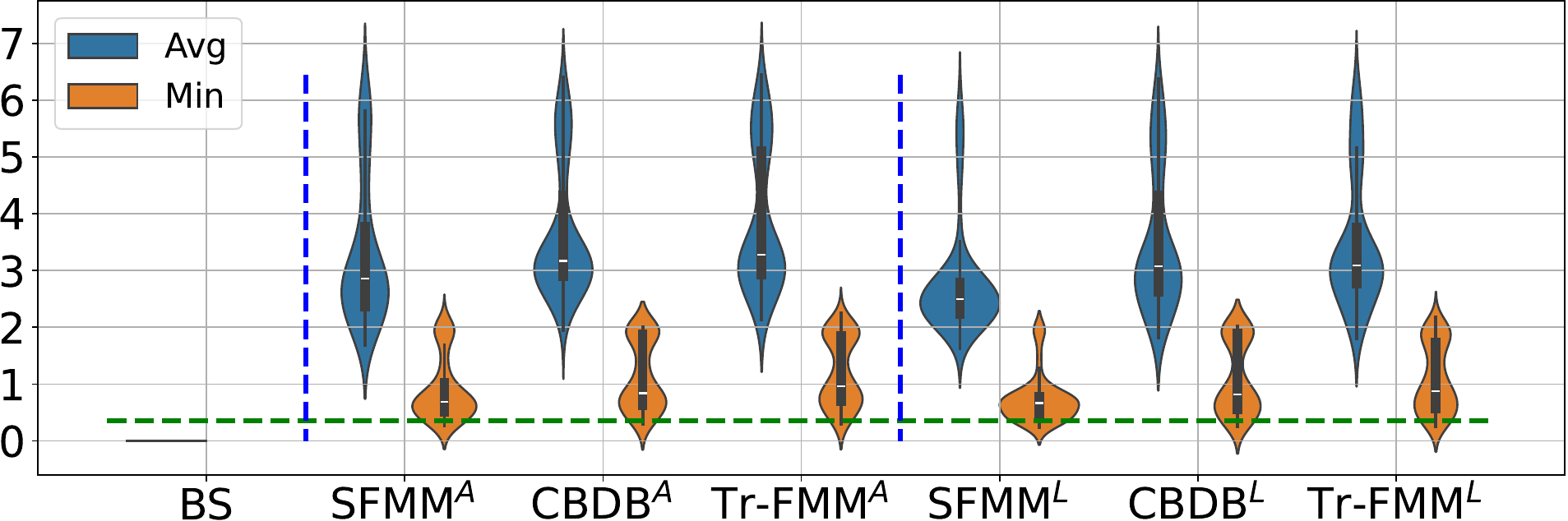}\label{fig:res_metodos_d2o}} \hspace{5mm}
        \subfigure[Success rates]{\includegraphics[width=.45\textwidth]{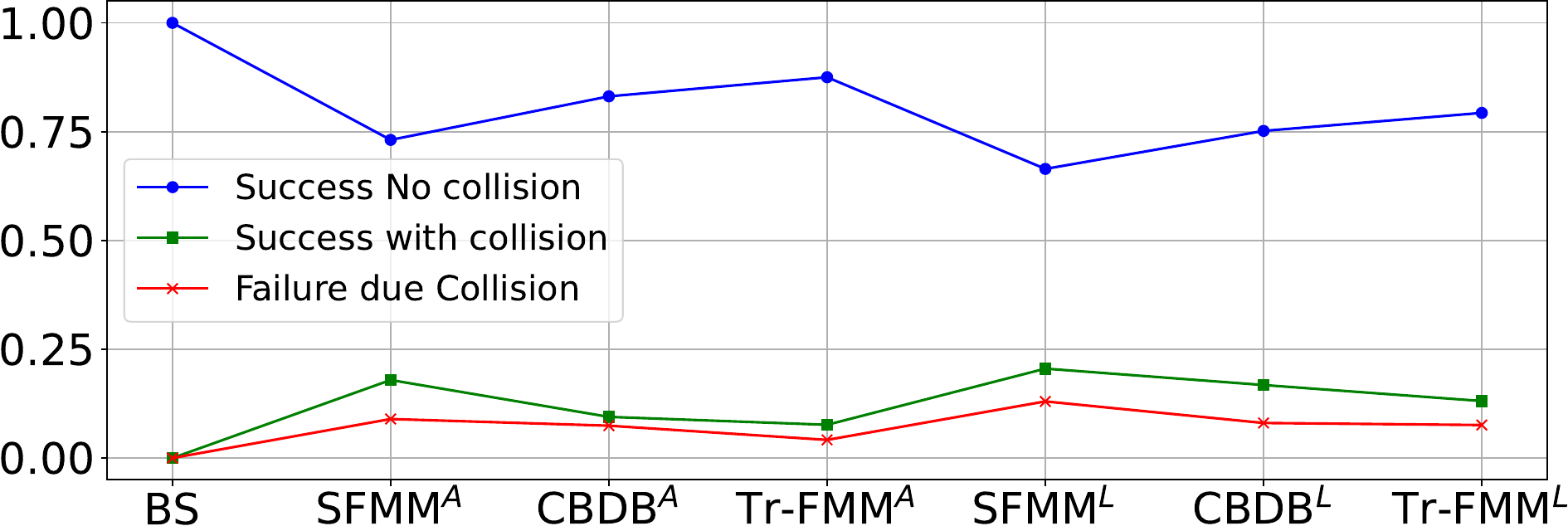}\label{fig:res_metodos_succ}}
    \end{center}
    \caption{
        Comparison with other approaches.
    }
    \label{fig:res_metodos}
\end{figure*}

Now, we compare \emph{Tr-FMM} with two other well-known approaches, illustrated in Fig.\ref{fig:res_metodos_ej}. The first is the FMM-based social path planner (SFMM) introduced in \cite{social_path_planning_fmm}, which relies on a predefined \emph{social space} designed to keep the robot’s trajectory at a safe distance from humans or moving obstacles, as shown in Fig.\ref{fig:res_metodos_sfmm}. Both SFMM and \emph{Tr-FMM} share the common foundation of Fast Marching Methods and aim to maintain a safe buffer from obstacles. However, SFMM achieves safety by reducing the velocity map around static obstacles, effectively forcing the robot to navigate through areas that are often cluttered with obstacles. In contrast, \emph{Tr-FMM} strategically selects entire regions that are safer to traverse by evaluating the risk of obstacle movements within them, thereby keeping the robot’s path as far away from obstacles as possible, as highlighted in Fig.\ref{fig:res_metodos_trfmm}. This regional-level consideration of dynamic obstacle behavior allows \emph{Tr-FMM} to better anticipate and avoid risky areas, improving the overall safety and smoothness of the path.

The second method, from \cite{cbdb}, utilizes a Crowd-Based Dynamic Blockages (CBDB) layer that explicitly identifies and penalizes crowded regions with high obstacle density, Fig.\ref{fig:res_metodos_cbdb}. CBDB evaluates candidate paths by assigning penalties to those that cross through areas with dense obstacle occupation, effectively steering the robot towards less congested routes. Unlike CBDB, \emph{Tr-FMM} not only minimizes path deviation but also incorporates predictions of obstacle dispersion into nearby obstacle-free regions, as modeled by eq.\ref{eq:dispersion}, which accounts for future potential occupation. Simultaneously, \emph{Tr-FMM} balances the detour length with safety by considering deviation from the direct goal path to reduce excessive wandering, while prioritizing less crowded regions to minimize collision risk, as demonstrated in Fig.\ref{fig:res_metodos_trfmm}. To ensure a fair and consistent comparison, we utilize the FMM method to compute paths for CBDB instead of the A* algorithm originally employed in \cite{cbdb}, guaranteeing that the robot maintains adequate distance from static obstacles.

The performance results are presented in Fig.\ref{fig:res_metodos}. With SFMM, the robot exhibits notably riskier behavior, often navigating through areas dense with obstacles. This is evidenced by the minimum distance to obstacles in most trials being very close to the robot’s radius, as illustrated in Fig.\ref{fig:res_metodos_d2o}, which directly contributes to an increased number of collisions, depicted in Fig.\ref{fig:res_metodos_succ}. Despite this elevated risk, SFMM does not achieve reductions in travel distances or mission times, which remain comparable to other methods as shown in Fig.\ref{fig:res_metodos_dist} and Fig.\ref{fig:res_metodos_time}. Moreover, SFMM tends to spend significantly more time stationary. This occurs because the robot enters densely occupied regions, which subsequently block its progress, forcing frequent replanning and path adjustments. By contrast, both CBDB and \emph{Tr-FMM} proactively avoid these crowded regions and opt directly for safer alternative routes, suggesting that evaluating and steering clear of entire regions with high dynamic obstacle presence is more effective than simply repelling the robot’s path from individual obstacles.

\begin{figure*}[tb!]
    \begin{center}

        \subfigure[Traveled distance (m)]{\includegraphics[width=.45\textwidth]{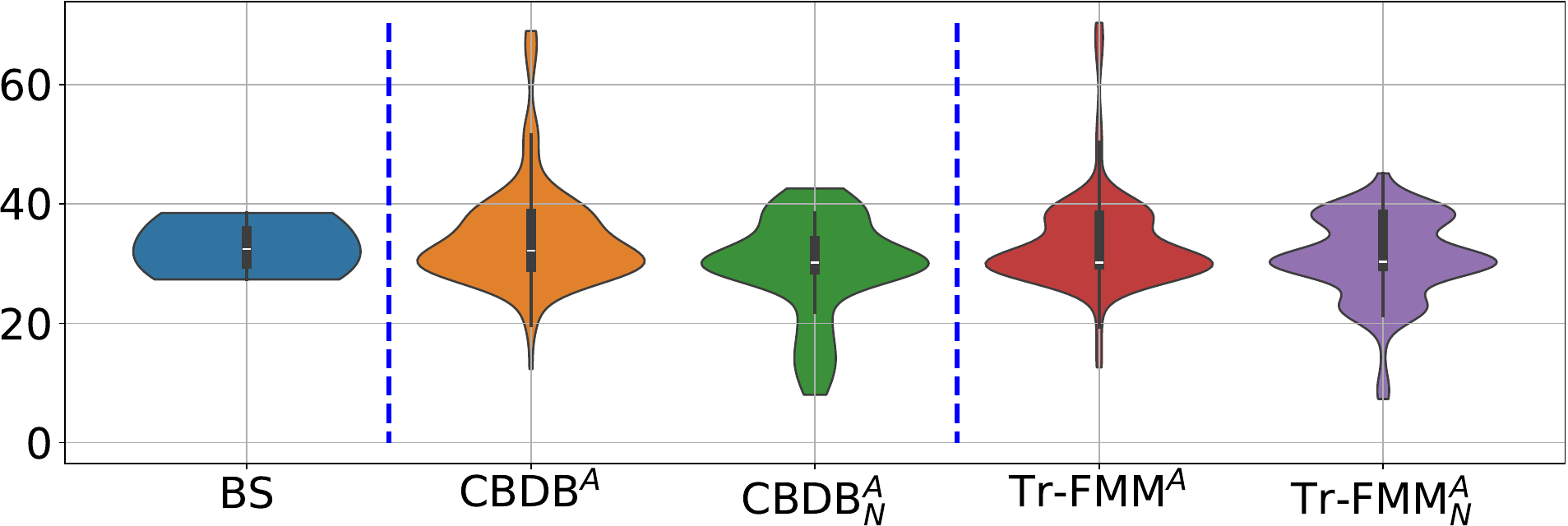}\label{fig:res_noreplan_dist}} \hspace{5mm}
        \subfigure[Times (s)]{\includegraphics[width=.45\textwidth]{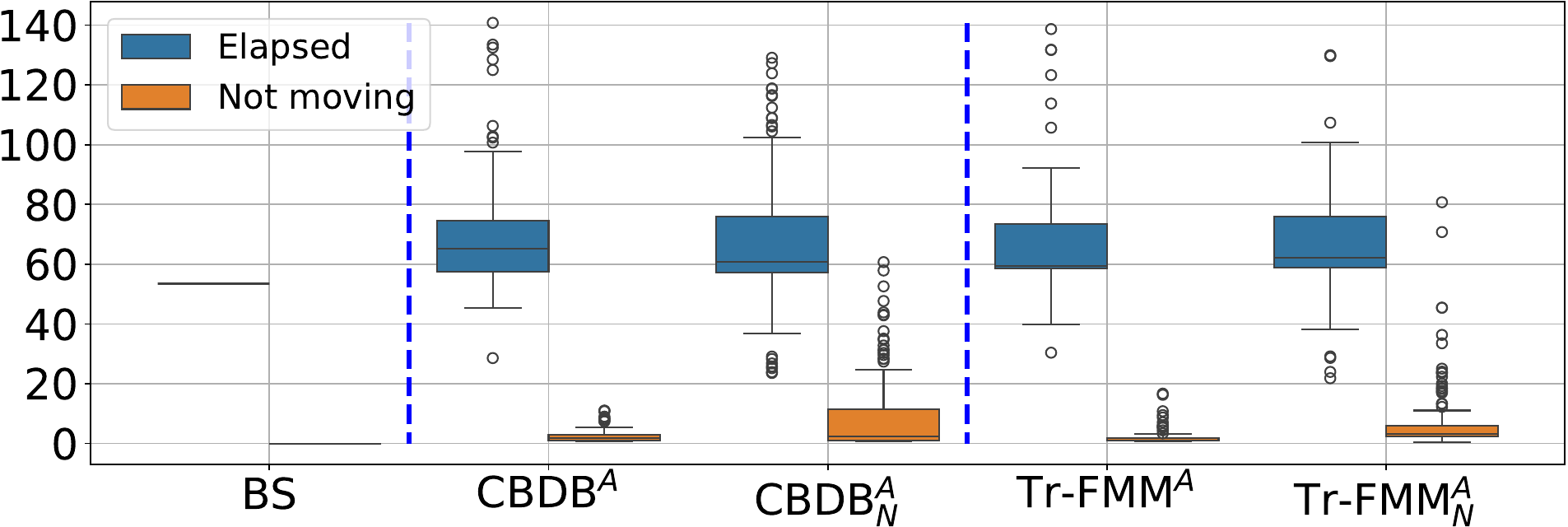}\label{fig:res_noreplan_time}} \\
        \subfigure[Distance to obstacles (m)]{\includegraphics[width=.45\textwidth]{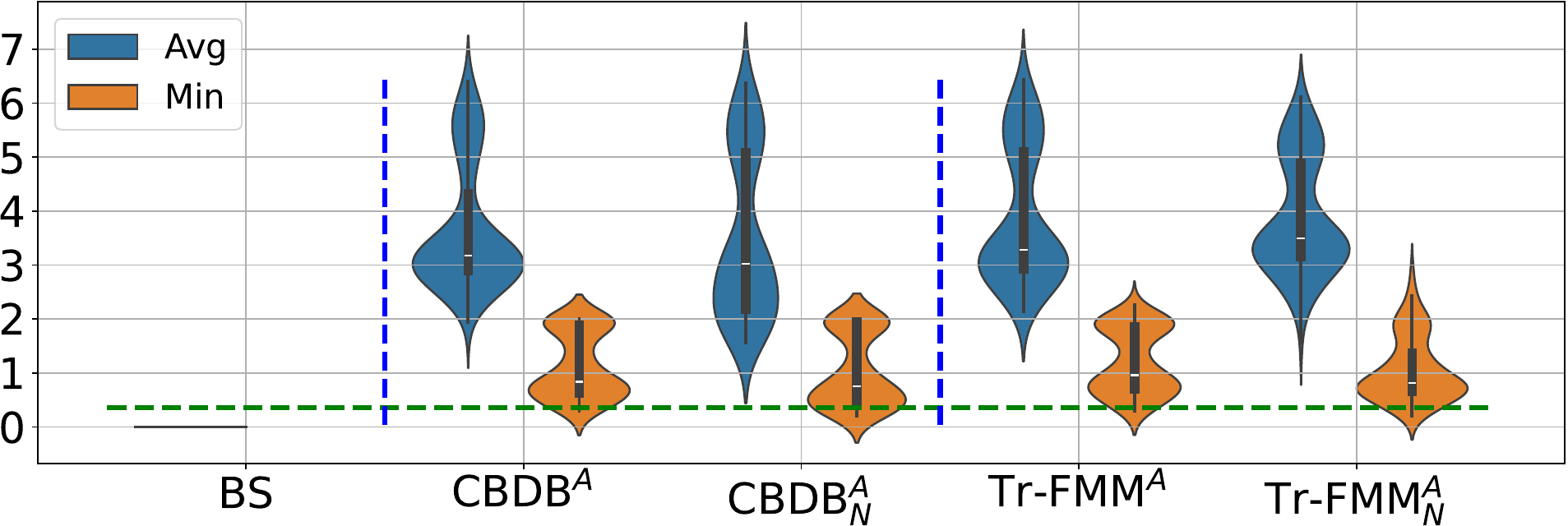}\label{fig:res_noreplan_d2o}} \hspace{5mm}
        \subfigure[Success rates]{\includegraphics[width=.45\textwidth]{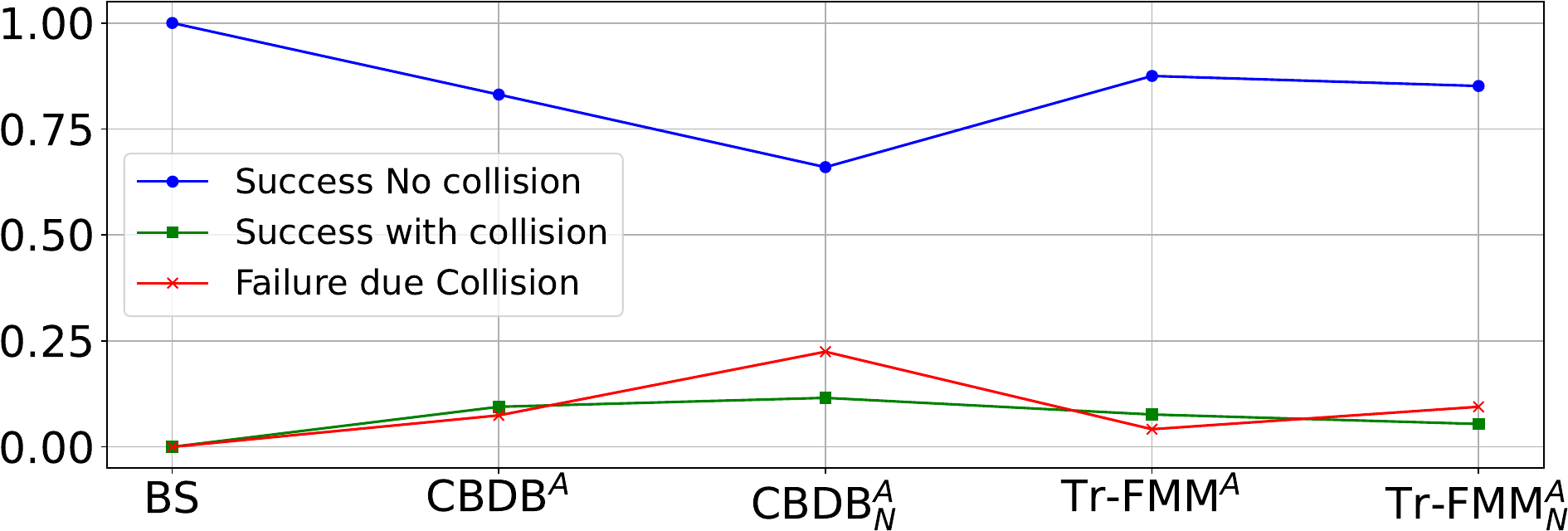}\label{fig:res_noreplan_succ}}
    \end{center}
    \caption{
        Results with and without replanning.
    }
    \label{fig:res_noreplan}
\end{figure*}

When analyzing CBDB’s performance under full obstacle position knowledge, its behavior is broadly similar to \emph{Tr-FMM}, with \emph{Tr-FMM} showing a slight but consistent advantage across all measured metrics. This gap widens when line-of-sight information only is available. The key difference is that \emph{Tr-FMM} not only counts the number of obstacles within regions, as CBDB does, but also accounts for the dynamic nature of these obstacles and their potential to interfere with movement, by leveraging eq.\ref{eq:occupation}, which encodes where these obstacles are moving inside regions. Furthermore, \emph{Tr-FMM} incorporates obstacle dispersion predictions to avoid regions likely to be occupied in the near future, resulting in fewer collisions. This underlines the critical role of eq.\ref{eq:dispersion} in forecasting occupied regions and guiding safer path planning. Consequently, \emph{Tr-FMM} typically exhibits higher travel distances and longer mission times compared to CBDB, reflecting its strategic willingness to accept greater deviations from the direct path to the goal in exchange for significantly enhanced collision avoidance.

\subsubsection{Planning once vs. replanning each timestep}

With this final experiment, we evaluate the capability of \emph{Tr-FMM} to predict future obstacle movements and its impact on navigation performance. The results are illustrated in Fig.\ref{fig:res_noreplan}. The test is conducted in a dense scenario where full knowledge of all obstacle positions is available. Unlike previous experiments involving continuous replanning, in this case, we compute the robot’s path only once at the start using both CBDB and \emph{Tr-FMM}. The robot then follows this initial path using the Dynamic Window Approach (DWA), stopping whenever obstacles physically block its way. These conditions are denoted as $CBDB^A_N$ and $\emph{Tr-FMM}^A_N$, where the subscript $N$ indicates no replanning during execution.

Importantly, because \emph{Tr-FMM} incorporates a prediction of dynamic obstacles dispersion through eq.\ref{eq:dispersion}, the obstacles are allowed to move for 20\% of the time the robot would take to reach the goal in an obstacle-free environment. This movement time of the obstacles is excluded from the final timing results to ensure fairness in comparison.

As anticipated, the performance of both methods deteriorates under these constraints without replanning, yet \emph{Tr-FMM} exhibits considerably less degradation. Although the minimum distance to obstacles decreases for \emph{Tr-FMM}, as shown in Fig.\ref{fig:res_noreplan_d2o}, its average distance to obstacles remains comparable to the scenario where continuous replanning is employed. In contrast, CBDB results in the robot remaining consistently close to obstacles throughout its trajectory. This close proximity leads to significantly more idle time during which the robot must wait for obstacles to clear its path, as captured in Fig.\ref{fig:res_noreplan_time}, and also causes an increased number of collisions, seen in Fig.\ref{fig:res_noreplan_succ}. Such collisions frequently occur because the robot either becomes trapped within crowded regions or directly encounters dynamic obstacles blocking its way.

The core advantage of \emph{Tr-FMM} in this experiment stems from its incorporation of eq.\ref{eq:dispersion}, which effectively models the potential future dispersion of obstacles based on their current dynamism. This predictive ability enables the robot to anticipate areas likely to become risky and proactively select paths that minimize the chance of encountering blockages or collisions. Therefore, even in the absence of continuous replanning, \emph{Tr-FMM} achieves safer navigation by embedding foresight into its initial path computation, leading to smoother and more reliable mission execution.

\begin{figure*}[tb!]
    \begin{center}
        \subfigure[Setup and map (73.3x50 m)]{\includegraphics[width=.32\textwidth]{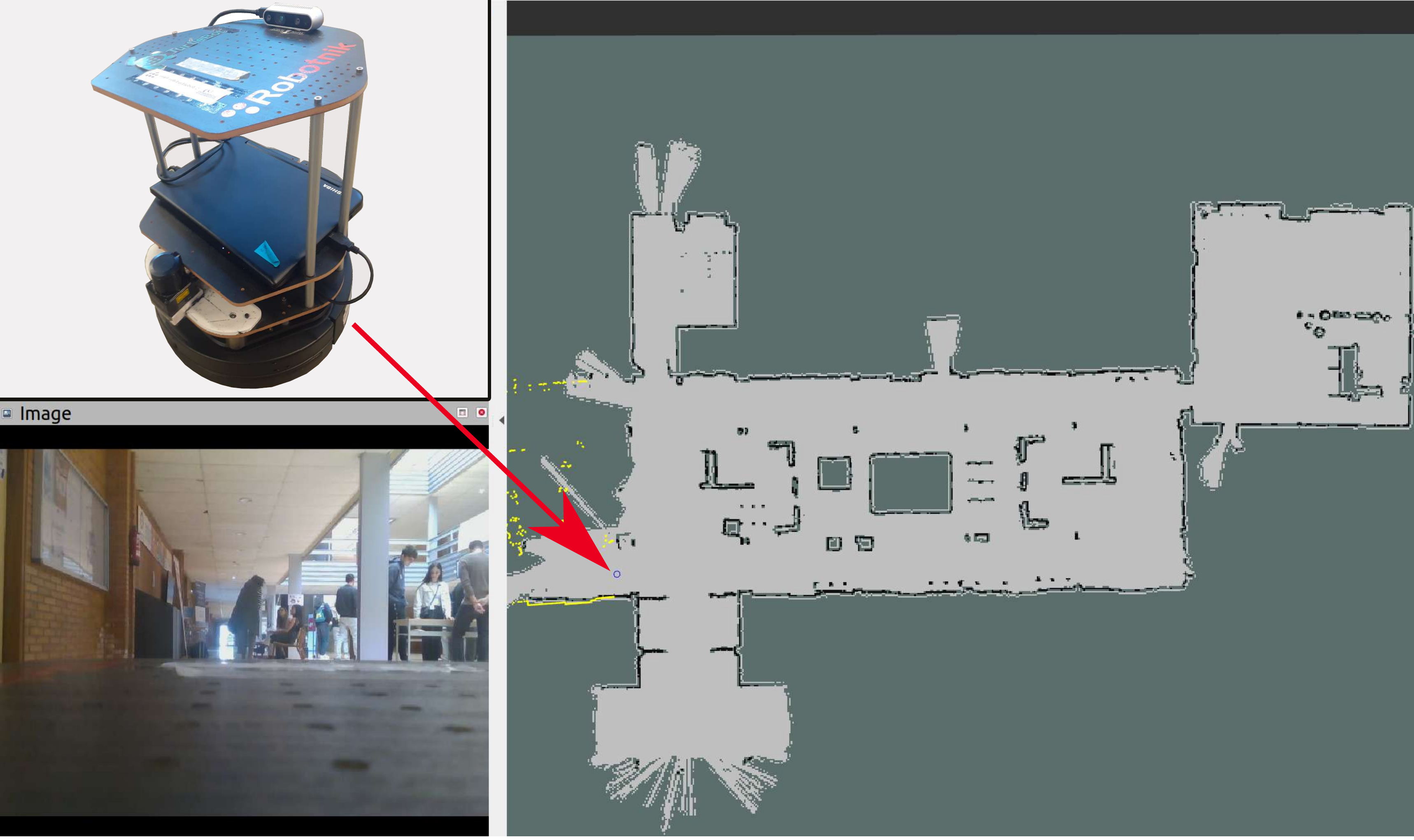}\label{fig:experiment_setup}} \hfill
        \subfigure[Experiment with few obstacles]{\includegraphics[width=.32\textwidth]{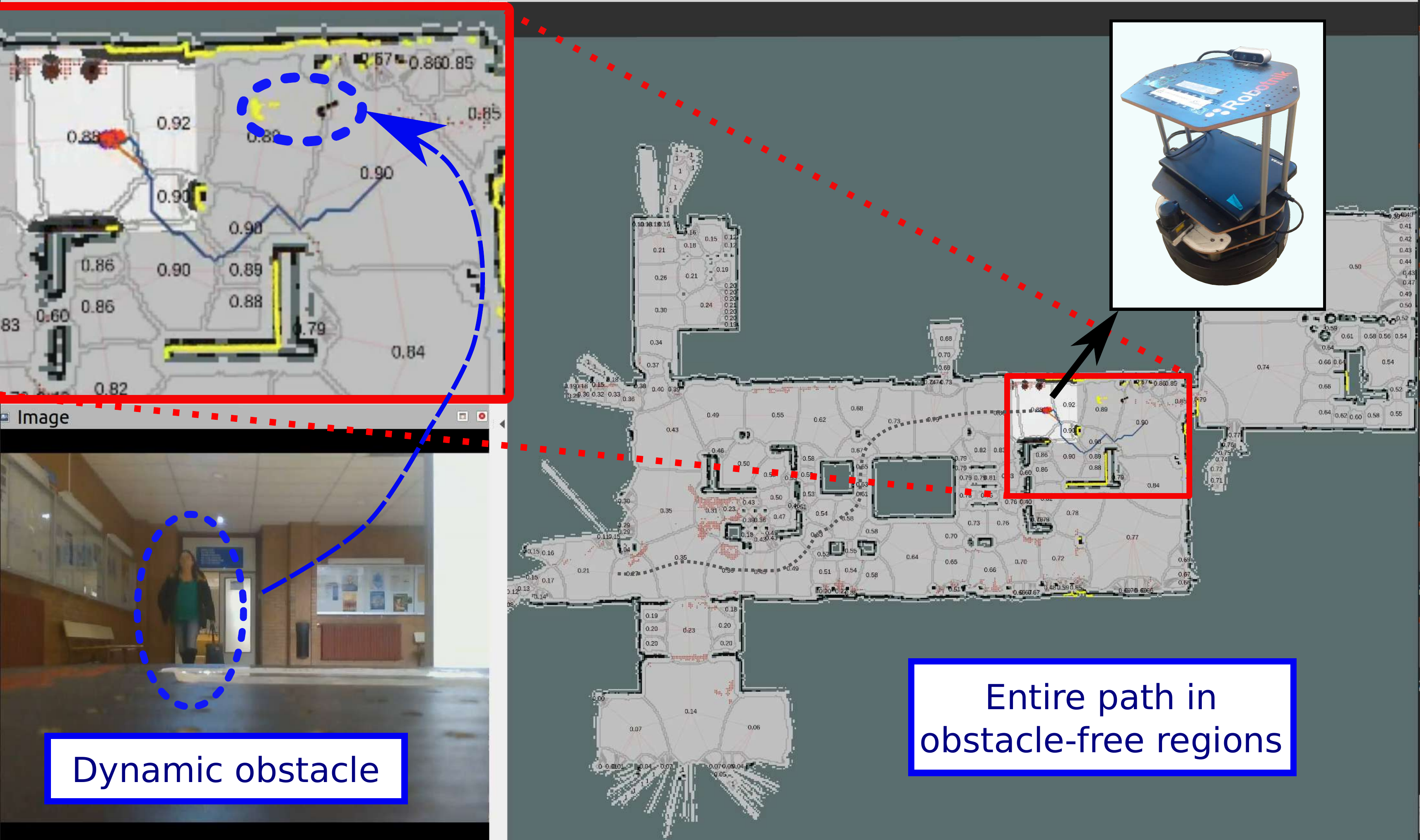}\label{fig:experiment1}} \hfill
        \subfigure[Experiment with many obstacles]{\includegraphics[width=.32\textwidth]{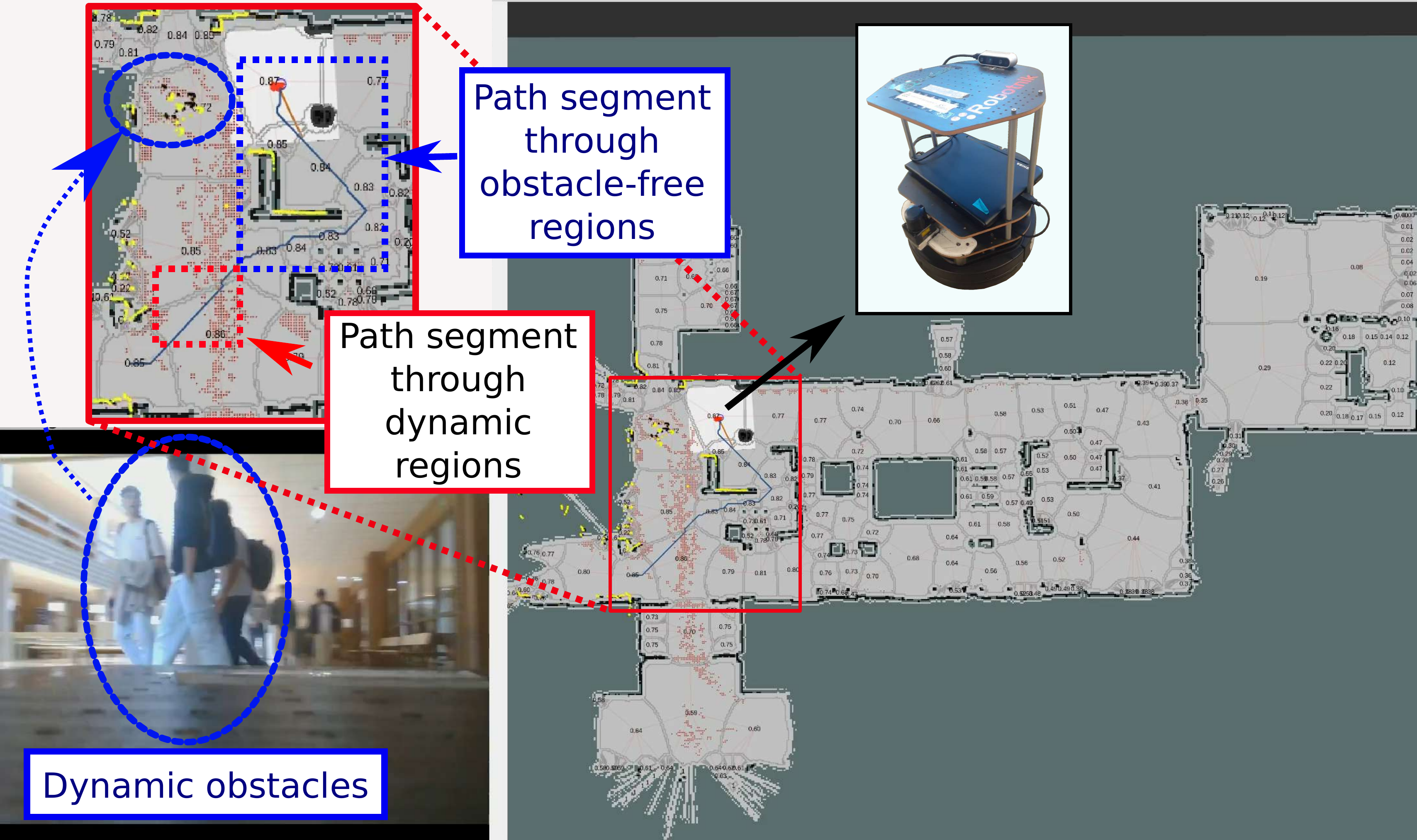}\label{fig:experiment2}}
    \end{center}
    \caption{
        Real-world experimentation. (a) shows the map of the environment, the robot setup and its initial position in the map. (b) captures the robot avoiding shared space with a moving obstacle while heading toward the goal. (c) depicts a frame where the robot plans a path to minimize its presence in regions with high human flow, where a segment of the path is within dynamic regions due to proximity to goal.
    }
    \label{fig:res_real_world}
\end{figure*}

\subsection{Real-world experimentation}

Three real-world experiments were conducted to validate the proposed \emph{Tr-FMM} method, progressively increasing in complexity—from a controlled laboratory setup to a highly dynamic public space. An illustrative example of a mission execution is available in the accompanying video\footnotemark. In all experiments, a Turtlebot platform equipped with a Hokuyo lidar sensor (20-meter range, 180° field of view) was used for obstacle detection. A camera mounted on the robot recorded the onboard perspective. Localization was provided by Adaptive Monte Carlo Localization (AMCL). For navigation, the \emph{Tr-FMM} planner was used to generate global paths, which were then smoothed and executed via DWA local planner.

\footnotetext{\url{https://bitly.cx/SYhSn}}

\textbf{Experiment 1 — Controlled Lab Setting.}  
The first experiment took place in the Robotics Lab at the University of Zaragoza (Fig.~\ref{fig:intro}), a small and irregular indoor environment characterized by the presence of static objects that are typically not included in idealized floor plans (e.g., tables, shelves, and equipment). Only one dynamic obstacle (a walking person) was present. This setup allowed us to validate the system under controlled conditions.

This experiment served to confirm several core capabilities of the proposed system:
\begin{itemize}
    \item The discretization module successfully extracted meaningful topological structure from the environment.
    \item The three components of the traversability function—goal orientation, region occupation, and obstacles dispersion—interacted coherently to guide the planner’s decisions.
    \item The planner generated a safe and efficient path that avoided the dynamic obstacle and reached the goal without incident.
\end{itemize}

A noteworthy observation emerged during the preparation phase: in the absence of central static obstacles (e.g., the boxes in Fig.\ref{fig:1}), the environment discretization process would have produced a single large region. This would reduce the topological complexity of the environment, yielding fewer alternative paths. As a result, the planner might generate a sub-optimal solution by guiding the robot closer to the dynamic obstacle due to a lack of routing options.

To address this, future work will explore automatic subdivision of large regions to increase path diversity. Instead of relying on detected local maxima (as in Fig.\ref{fig:discr_max}), a fixed number of seed points could be placed within each large region to generate more balanced subregions, as suggested in Fig.\ref{fig:discr}.

\textbf{Experiment 2 — Large, Low-Dynamic Environment.}  
The second experiment was conducted in the main hall of the university (Fig.\ref{fig:experiment_setup}), a large and similarly irregular space. This environment contained few dynamic obstacles and no prior knowledge of exact obstacle placements, making it ideal to evaluate planning performance under realistic conditions (Fig.\ref{fig:experiment1}).

This experiment demonstrated several important capabilities of \emph{Tr-FMM}:
\begin{itemize}
    \item The discretization module was able to extract a coherent topological structure even from imperfect maps.
    \item The planner operated in real time ($<$100 ms per plan), for goal located more than 40 meters away.
    \item The planner decisions were strongly driven by goal direction and low occupancy levels.
\end{itemize}

While minor inconsistencies between the static map and the actual environment (e.g., unmodeled obstacles or localization errors) were present, their impact was limited. This robustness is attributed to the occupancy estimation in eq.\ref{eq:occupation}, which uses real-time obstacle data to dynamically update traversability. Moreover, these inaccuracies typically affected only the periphery of regions, minimizing their influence on the overall path planning.

\textbf{Experiment 3 — Dynamic, Crowded Environment.}  
The third experiment reused the same hall but placed the robot at the main entrance of the building—an area characterized by a continuous, dense flow of people (Fig.\ref{fig:experiment2}). Over a 10-minute session, the user provided new goals at runtime, requiring the robot to traverse through highly dynamic zones.

Crucially, no pedestrians were informed of the robot’s task, ensuring natural and unaltered human behavior. Despite the increased complexity, \emph{Tr-FMM} maintained its preference for regions with lower pedestrian density, avoiding crowded zones whenever possible, as can be observed in the video and Fig.\ref{fig:experiment2}. In only two cases did the robot enter these high-density areas: once when no alternative path existed, and once when the detour would have resulted in excessive deviation from the goal.

Importantly, no collisions occurred during the experiment. This is likely due to the combination of the robot’s real-time trajectory adaptation and the natural tendency of people to avoid the robot. These results reinforce the importance of planning strategies that respect human movement patterns and prioritize social safety. Quantitatively, this behavior is reflected in the metrics: the robot maintained a minimum distance of 0.29 meters and an average of 2.43 meters from the nearest person—values consistent with those observed in simulation in dispersed dynamic scenarios.

In summary, these experiments confirm that \emph{Tr-FMM} generalizes well to real-world settings, handles map imperfections robustly, and provides safe, efficient, and socially aware navigation even in complex and dynamic environments.

\section{Conclusions}\label{sec:conclusions}

In this work, we introduced the Traversability-aware FMM (\emph{Tr-FMM}) for path planning in environments with moving obstacles. Our approach discretizes the environment and assigns a traversability value to each region based on observed obstacle movements. This value considers obstacle density, disturbance level, probability of dispersion, and estimated deviation from the direct path to the goal. \emph{Tr-FMM} balances risk reduction when traversing dynamic areas with the need to avoid excessive detours. Unlike purely local planners that reactively avoid individual moving obstacles, our global planning strategy identifies and proactively avoids densely populated risk zones. This enables the robot to maintain a better trade-off between path efficiency, pedestrian safety, and reduced disruption to human motion in accordance with social navigation norms.

The results show that \emph{Tr-FMM} outperforms techniques relying on social models or dynamic congestion avoidance. By considering not only the number of obstacles but also their movement patterns, it better models traversal risk. Additionally, incorporating obstacle dispersion into the traversability computation, reduces the chances of entering a region that later becomes occupied. Even when collisions occur, \emph{Tr-FMM} tends to select less hazardous regions, allowing the robot to complete its mission safely.

Future work will refine the dispersion model by incorporating obstacle movement tendencies rather than assuming equal probability in all directions. We also aim to extend \emph{Tr-FMM} to multi-robot systems and to other mission types, such as coverage and exploration. Another direction is integrating environmental context  in the decision-making process, such as factory workspaces or queue formations in stores and airports, to adapt the obstacle avoidance strategies.

\section*{Acknowledgments}

This work was partially supported by the Spanish projects
PID2022-139615OB-I00/MCIN/AEI/10.13039/501100011033/FEDER-UE and Aragon Government FSE-T45\_23R.

\section*{Declaration of generative AI and AI-assisted technologies in the writing process }

During the preparation of this work the authors used ChatGPT in order to improve text readability. After using this tool, the authors reviewed and edited the content as needed and take full responsibility for the content of the published article.

\bibliographystyle{elsarticle-num}
\balance
\bibliography{ref}

\end{document}